\documentclass[manuscript]{acmart}
\AtBeginDocument{%
  }

\setcopyright{acmlicensed}
\copyrightyear{2018}
\acmYear{2018}
\acmDOI{XXXXXXX.XXXXXXX}
\acmConference[Conference acronym 'XX]{Make sure to enter the correct
  conference title from your rights confirmation email}{June 03--05,
  2018}{Woodstock, NY}
\acmISBN{978-1-4503-XXXX-X/2018/06}

\usepackage{ragged2e}
\usepackage{graphicx}
\usepackage{amsmath}
\usepackage{amsfonts}
\usepackage{array,booktabs,longtable}
\usepackage{multirow}
\usepackage{enumitem}
\usepackage[ruled,vlined]{algorithm2e}
\usepackage[ruled,vlined]{algorithm2e}
\usepackage{subcaption}
\usepackage{tikz}
\usetikzlibrary{arrows.meta,shapes.geometric,positioning,calc}
\hypersetup{hidelinks}
\usepackage{pgfplots}
\usepackage{pgfplotstable}
\usepackage{xcolor}
\pgfplotsset{compat=1.18}
\newcommand{\TauH}{\tau_H}
\newcommand{\TauF}{\tau_F}

\newcommand{\IR}{\mathrm{IR}}
\newcommand{\GE}{\mathrm{GE}}

\newcolumntype{L}[1]{>{\raggedright\arraybackslash}p{#1}}

\begin{document}

\title{Illusion or Integrity? Geometrical Consistency Metric for AIGC Video Quality Evaluation}

\author{Yifei Xue}
\authornote{Both authors contributed equally to this research.}
\affiliation{%
  \institution{Hunan University}
  \country{China}
}

\author{Yuanchen Fei}
\authornote{Both authors contributed equally to this research.}
\affiliation{%
  \institution{Hunan University}
  \country{China}
}

\author{Hao Zhang}
\affiliation{%
  \institution{Hunan University}
  \country{China}
}

\author{Chenzhi Nie}
\affiliation{%
  \institution{Hunan University}
  \country{China}
}

\author{Tie Ji}
\affiliation{%
  \institution{Hunan University}
  \country{China}
}

\author{Yizhen Lao}
\affiliation{%
  \institution{Hunan University}
  \country{China}
}

\renewcommand{\shortauthors}{Trovato et al.}

\begin{abstract}
  Recently, AI-driven video generation has attracted considerable attention. This surge increases demand for reliable video quality assessment (\textbf{VQA}) metrics to assess AI-generated content (\textbf{AIGC}) videos and guide model optimization. Existing work examines VQA via visual harmony, video–text consistency, and domain-specific alignment, yet lacks quantitative metrics for fidelity to physical laws. To address this limitation, we present a novel benchmark that evaluates the quality of AIGC videos through their compliance with physical principles, achieved by quantitatively measuring the geometrical consistency across frames extracted from the generated sequences, thus as a proxy to gauge the extent to which the generated videos conform to real-world physics rules. Specifically, \textbf{GeoCon-Bench} gates global motion via translation estimates, fits a homography or fundamental matrix to background correspondences, and reports complementary metrics (inlier ratio, geometric error). We also release a dataset of 20 scenes across six motion categories. Experiments on state-of-the-art AIGC models demonstrate GeoCon-Bench's reliability as a video quality metric.
\end{abstract}


\begin{teaserfigure}
    \centering
    \includegraphics[width=0.8\textwidth]{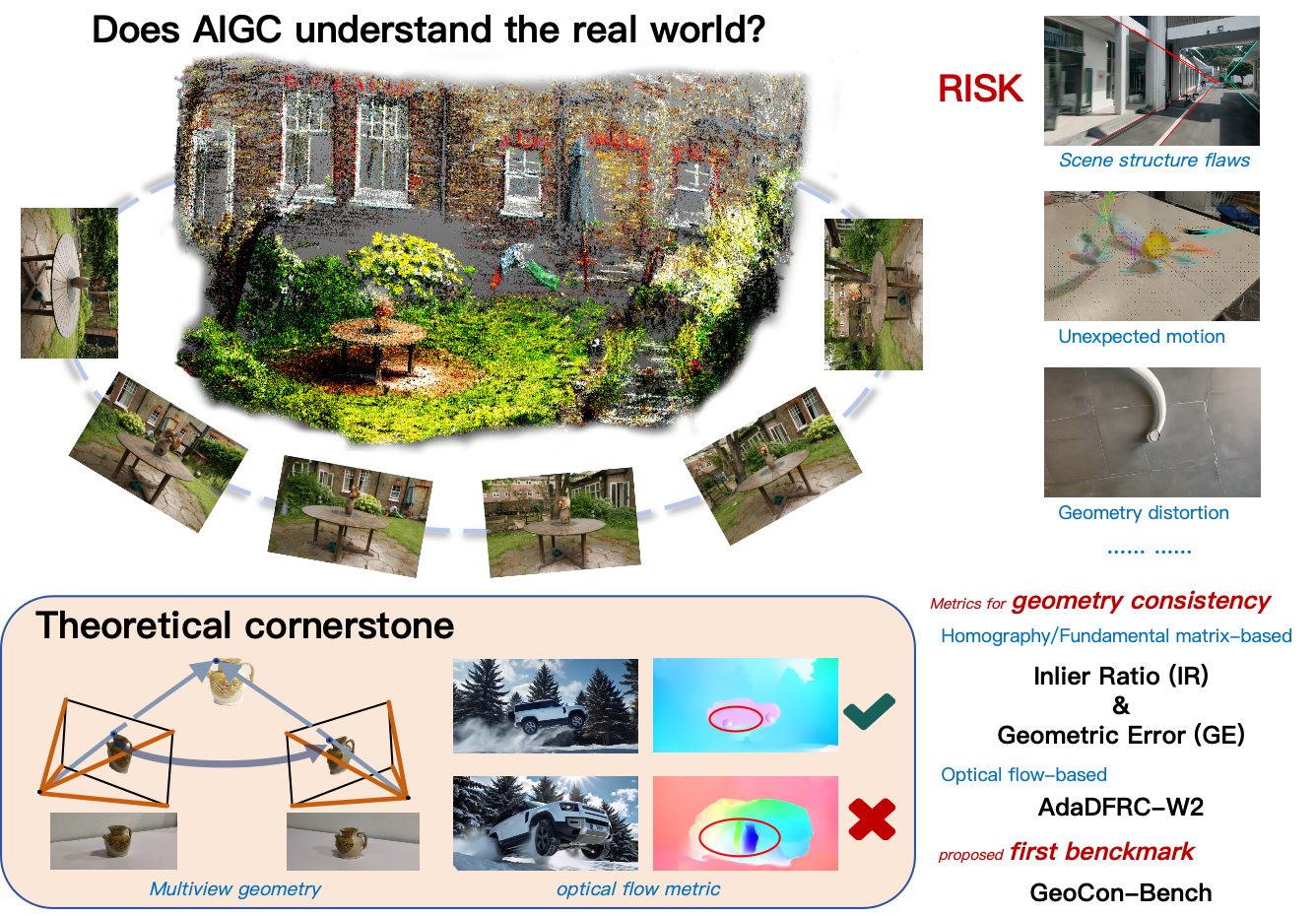}
    \caption{To quantify how well AIGC-generated videos preserve physical-world geometry, we introduce GeoCon-Bench, a model-agnostic evaluation framework grounded in multiview geometry and optical-flow analysis. The figure shows (top) a generated camera sweep; (right) representative failure modes; and (bottom) the theoretical cornerstones we use. GeoCon-Bench scores clips via homography/fundamental-matrix metrics for rigid backgrounds and flow-based tests for dynamic regions, enabling reproducible auditing of geometric consistency.}
    \Description{}
    \label{fig:teaser}
\end{teaserfigure}

\received{20 February 2007}
\received[revised]{12 March 2009}
\received[accepted]{5 June 2009}

\maketitle

\section{Introduction}
\label{sec:introduction}

Generative video offers a tantalizing pathway to \emph{lower the cost and latency} of 3D asset creation~\cite{mildenhall2021nerf,kerbl20233d}.
Rather than capturing calibrated footage and running bespoke reconstruction pipelines, practitioners could in principle \emph{feed an idealized video} to a reconstruction system and obtain geometry ready for downstream use.
Yet this promise is ultimately gated by \emph{photogrammetry and multi-view geometry}~\cite{hartley2004multiple}: usable parallax, consistency with either homography or epipolar structure, and coherent occlusion handling are \emph{prerequisites}, not niceties.
When those prerequisites fail, error cascades are immediate and practical—point clouds fragment or rubber-sheet, camera poses drift, novel views shear, and AR overlays refuse to lock—even though the sequence may still look plausible to the eye~\cite{jang2025too,kamali2025characterizing}.
Recent progress in diffusion/transformer generators~\cite{bar2024lumiere,kondratyuk2023videopoet,vaswani2017attention,dosovitskiy2020image,ho2020denoising} increases perceptual quality but does not guarantee photogrammetric validity~\cite{heusel2017gans,unterthiner2018towards,zhang2018unreasonable,peebles2023scalable,chu2020learning}.
Accordingly, the \emph{value proposition} of video-to-3D is determined not by aesthetics or prompt adherence, but by \emph{geometric reliability}~\cite{schonberger2016structure}.

\paragraph{Perceptual plausibility $\neq$ photogrammetric correctness}~\cite{wang2004image,huang2024vbench,huang2024vbench++}.
Contemporary systems can score well on perceptual or distributional metrics while \emph{systematically violating} multi-view constraints.
Typical failure modes that matter for reconstruction include: (i) background slide where planar regions move inconsistently with the camera; (ii) rubber-sheet warps that break the assumption of a single rigid background model; and (iii) inconsistent scale–depth relations across time that a bundle adjuster cannot reconcile.
These violations have been documented both qualitatively and quantitatively in generative pipelines~\cite{sarkar2024shadows}, and they are \emph{fatal} for multi-view reconstruction, pose estimation, and novel-view synthesis (NVS).

Mainstream evaluation emphasizes pixel fidelity, learned perceptual similarity, distributional distances, semantics, or composite leaderboards~\cite{huang2024vbench,huang2024vbench++,fan2023aigcbench,liu2024evalcrafter,liu2024survey,zhang2025q,qu2024exploring,duan2025worldscore}.
While valuable for breadth, such axes are \emph{insensitive to multi-view geometric validity}, especially in real-data settings where ground truth (GT) is unavailable.
In practical platform comparisons, GT is \emph{usually absent}: real camera trajectories and scene geometry are proprietary or intractable to share; generated videos often have no physical camera at all; and synthetic GT, while useful scientifically, can \emph{misalign with deployment distributions and sensors}, confounding cross-platform conclusions.
Therefore, the community lacks a \emph{GT-free, photogrammetry-aligned, auditable} protocol that can be reproduced across generators and data regimes.

Hence, we derive four design requirements for a practical geometry-consistency protocol:
\textbf{R1 Regime awareness}: background modeling must switch between homography and fundamental descriptions based on \emph{translation evidence};
\textbf{R2 Background–dynamic disentanglement}: rigid backgrounds and moving/non-rigid regions obey different assumptions and must be \emph{scored against different objectives};
\textbf{R3 GT-free feasibility}: every stage must operate without GT, so that results are reproducible across platforms and datasets;
\textbf{R4 Auditability and comparability}: decisions and scores must be \emph{traceable} (with gate rationales, inlier diagnostics, and threshold sensitivity) so platform-level comparisons are defensible.

We propose a unified, GT-free protocol that (i) extracts translation evidence from pose logs to \emph{gate} the background model, (ii) fits the model with \emph{calibrated dual thresholds} tailored to each residual family, (iii) scores dynamic regions separately via an optical-flow-guided measure, and (iv) logs diagnostics for auditability.
Concretely, a gate variable $\gamma_t\!\in\!\{\mathrm{H},\mathrm{F}\}$ selects a homography when translation is negligible and a fundamental model otherwise.
For background matches, we compute
\(\mathrm{IR}_H,\mathrm{GE}_H\) under a homography residual with threshold $\tau_H$ and
\(\mathrm{IR}_F,\mathrm{GE}_F\) under an epipolar/Sampson residual with threshold $\tau_F$,
explicitly allowing $\tau_H\!\neq\!\tau_F$ because the two residual families have \emph{different statistical scales}.
We accompany indicators with \emph{threshold–score sensitivity scans} and \emph{percentile alignment} to ensure cross-platform interpretability.
Dynamic/non-rigid pixels are evaluated with a flow-guided measure (e.g., AdaDFRC-W2): forcing them into H/F would \emph{pollute} inlier sets, destabilize robust fitting, and render IR/GE less diagnostic of background validity.
The protocol logs gate rationales, inlier maps, threshold–score curves, and pose-log vs.\ fit sanity checks to support reproducible, inspector-friendly audits.

To our knowledge, this is the first \emph{GT-free, regime-aware, photogrammetry-aligned} evaluation protocol for AIGC videos that is auditable across platforms. To summarize, our contributions are threefold: 

\begin{itemize}
    \item We introduce the first GT-free, regime-aware, and photogrammetry-aligned evaluation framework GeoCon-Bench for AIGC videos that supports auditable, cross-platform assessments. 
    \item We design a translation-gated framework that adaptively selects between homography and fundamental modeling, proposes calibrated dual-threshold indicators for background consistency, and introduces AdaDFRC-W2, a Wasserstein-2 regional flow-consistency metric for robust evaluation of dynamic content.
    \item We release auditing assets including lightweight logs, coverage statistics, score curves, and a dedicated benchmark dataset that spans diverse scenes and motion regimes, enabling reproducible and inspector-friendly audits. \textit{GeoCon-Bench will be publicly available after acceptance}.
\end{itemize}

\section{Related Works}
\label{sec:related}

Mainstream evaluation families—pixel fidelity (PSNR/SSIM)~\cite{wang2004image}, learned perceptual similarity (LPIPS)~\cite{zhang2018unreasonable}, distributional distances (FID/FVD)~\cite{unterthiner2019fvd}, semantic/prompt alignment (CLIP)~\cite{hessel2021clipscore,radford2021learning}, and composite leaderboards—provide breadth but are largely \emph{insensitive to multi-view geometric validity}, particularly under GT-free conditions~\cite{huang2024vbench,huang2024vbench++}.
Their objectives emphasize local textures, global feature statistics, or text–image agreement, not whether homography/epipolar relations hold across viewpoints.
As diffusion/transformer generators improve perceptual quality~\cite{peebles2023scalable,chu2020learning,wang2025lavie,ma2024latte}, such blind spots become consequential: videos may look convincing yet remain unusable for 3D reconstruction, degrading pose estimation, bundle adjustment, and novel-view synthesis.
Recent multi-objective video benchmarks broaden axes (aesthetics, prompt adherence, temporal smoothness), but typically \emph{do not enforce} photogrammetric constraints on real data without ground truth~\cite{huang2024vbench,huang2024vbench++,jiang2024genai}.
Our work complements this breadth with explicit, photogrammetry-aligned checks that are feasible and auditable in GT-free settings.

Photogrammetry and multi-view geometry formalize how scene structure and camera motion are recovered from image correspondences; practical pipelines rely on correspondence quality (inlier ratios), geometric residuals (reprojection or epipolar/Sampson)~\cite{hartley2004multiple,fathy2011fundamental}, robust model fitting, and bundle adjustment~\cite{fischler1981random,torr2000mlesac,chum2005matching,chum2003locally,lebeda2012fixing}.
However, thresholds inside these systems are typically tuned for \emph{optimization stability} on particular datasets rather than for \emph{cross-platform comparability} and \emph{auditability}.
Crucially, the appropriate background model depends on camera motion: negligible translation~\cite{nister2004efficient} supports a homography on planar/parallax-free backgrounds, whereas observable translation calls for epipolar geometry on backgrounds with depth variation.
Modern monocular multi-view methods (e.g., DUSt3R) expose \emph{pose logs} that can be repurposed as \emph{translation evidence} to inform such regime choices~\cite{wang2024dust3r}.
We build on these signals but reframe them as \emph{comparable indicators} by (i) tying regime selection to translation evidence and (ii) calibrating \emph{dual thresholds} $(\tau_H,\tau_F)$ for model-appropriate residual families with sensitivity analyses, so reported IR/GE are interpretable across generators and datasets.

Moving or deforming content violates rigid-background assumptions; if dynamic pixels are forced into homography/fundamental fitting, they \emph{pollute} inlier sets, destabilize robust estimation, and reduce the diagnostic power of IR/GE for the background.
Optical flow and occlusion reasoning directly probe \emph{motion plausibility}, \emph{deformation continuity}, and \emph{disocclusion behavior}~\cite{teed2020raft,sun2018pwc,xu2022gmflow}, properties that matter for structure-from-motion~\cite{ozyecsil2017survey} and view synthesis.
In the generative setting, dynamics can be perceptually plausible while geometrically inconsistent across frames (e.g., flow fields that ignore parallax or violate occlusion ordering)~\cite{sarkar2024shadows}.
We therefore \emph{separate} background rigidity from dynamic coherence: H/F-based indicators summarize whether a \emph{single rigid} model explains the background, while a flow-guided score targets non-rigid and occlusion dynamics.
This separation is methodological rather than cosmetic, preserving interpretability and avoiding mixed-objective confounds.

GT-free or weakly supervised evaluations are standard in perceptual quality assessment, but GT-free \emph{geometry-aware} protocols remain scarce for platform-level comparisons.
Two challenges recur: \emph{defensibility} of decisions (e.g., why a particular regime/threshold) and \emph{traceability} of results (e.g., whether scores are stable across thresholds/unseen scenes).
In practice, GT is usually unavailable because real camera trajectories and scene geometry are proprietary, intractable to share, or nonexistent for generated videos. Synthetic GT, while scientifically useful, often misaligns with deployment distributions and sensor data.
We address these challenges by (i) grounding regime selection in pose-log translation evidence with sanity checks against fitted geometry, (ii) releasing \emph{threshold–score sensitivity scans} and \emph{percentile-aligned} summaries to ensure comparability, and (iii) logging inlier coverage and robust clip-level aggregation to make audits reproducible across systems.

Recent studies document how generative pipelines can violate physical/geometric constraints despite high perceptual scores~\cite{sarkar2024shadows}, while multi-objective benchmarks emphasize user-perceived axes over explicit multi-view validity~\cite{huang2024vbench,huang2024vbench++}.
We bridge these trajectories: retaining GT-free feasibility and breadth, but injecting \emph{photogrammetry-aligned} checks that are auditable and reproducible.
This reframes evaluation around practitioner questions: \emph{Will this video support reconstruction? If not, where does geometry fail, and are failures systematic across content and generators?}

Prior work provides broad perceptual/semantic coverage and mature geometric estimators, but stops short of an integrated, GT-free standard that \emph{enforces} multi-view validity while preserving auditability.
We next formalize the latent construct and observable indicators in a concept layer, then instantiate the protocol and diagnostics in \S\ref{sec:method}.

\section{Concept} 
\label{sec:concept}

\subsection{Construct, scope, and contextual factors}
We treat \emph{geometry consistency} as a latent construct with two facets:
\textbf{background rigidity} and
\textbf{dynamic-region coherence}.
Perceived salience of a specific residual depends on contextual factors, including scene depth span, planarity, edge density, motion spectrum, and occlusion complexity~\cite{teed2020raft,sun2018pwc,xu2022gmflow}. Consequently, indicators must be interpretable under diverse conditions and aggregated robustly over time.
The construct is operationalized without ground truth (GT-free)~\cite{huang2024vbench,huang2024vbench++} by checking self-consistency with photogrammetric constraints and by leveraging pose logs~\cite{wang2024dust3r} as auxiliary evidence.

\subsection{Regime-aware prior}
Let $I_t,I_{t+1}$ be consecutive frames with background matches $\{(\mathbf{x}_i,\mathbf{y}_i)\}_{i=1}^{N_b}$ and a background mask $\mathcal{M}_b$.
From pose logs (e.g., DUSt3R) we obtain a translation proxy $\tau_t$~\cite{wang2024dust3r} and define a gate
\begin{equation}
    \gamma_t \in \{\mathrm{H},\mathrm{F}\},
    \qquad
    \gamma_t=\begin{cases}
    \mathrm{H} & \text{if } \tau_t < \tau_{\text{gate}},\\[2pt]
    \mathrm{F} & \text{otherwise.}
    \end{cases}
\end{equation}
When $\gamma_t=\mathrm{H}$ we fit a homography matrix $\mathbf{H}_t$ to background correspondences; when $\gamma_t=\mathrm{F}$ we fit a fundamental matrix~\cite{hartley2004multiple} $\mathbf{F}_t$.
This prior encodes standard photogrammetric practice: homographies capture planar or parallax-free backgrounds under negligible translation, whereas epipolar geometry governs backgrounds with depth variation under translation.

\subsection{Dual-threshold indicators for background rigidity}
For a candidate model $g_t\in\{\mathbf{H}_t,\mathbf{F}_t\}$, let $e_i^{(g)}$ denote a model-appropriate residual. In our research, we utilize symmetric transfer error for $\mathbf{H}_t$, and Sampson/epipolar approximation for $\mathbf{F}_t$.
Because these residual families have \emph{different statistical scales and tails}, we calibrate \emph{dual thresholds} $(\tau_H,\tau_F)$ rather than forcing a single cutoff.
Define the inlier set
\begin{equation}
    \mathcal{S}^{(g)}_{\tau}=\bigl\{\,i\in\mathcal{M}_b \;\big|\; e_i^{(g)} \le \tau\,\bigr\},  
\end{equation}
and report two indicators~\cite{chum2005matching,torr2000mlesac}:
\begin{equation}
    \mathrm{IR}_H=\frac{|\mathcal{S}^{(\mathbf{H})}_{\tau_H}|}{N_b},\quad
    \mathrm{GE}_H=\operatorname{median}\{\,e_i^{(\mathbf{H})} : i\in\mathcal{S}^{(\mathbf{H})}_{\tau_H}\,\};
    \qquad
    \mathrm{IR}_F=\frac{|\mathcal{S}^{(\mathbf{F})}_{\tau_F}|}{N_b},\quad
    \mathrm{GE}_F=\operatorname{median}\{\,e_i^{(\mathbf{F})} : i\in\mathcal{S}^{(\mathbf{F})}_{\tau_F}\,\}.
\end{equation}
IR answers whether a single rigid model explains the background at step $t$; GE answers how tightly the consistent matches align when it does.
To ensure cross-platform interpretability, we accompany each report with \textbf{threshold–score sensitivity scans} over a grid of $(\tau_H,\tau_F)$ and with \textbf{percentile-aligned} summaries, mitigating absolute-scale differences across residual families and datasets.
See \S\ref{subsec:calibration} for threshold–score sensitivity scans, stability bands, and coverage definitions.

\subsection{Why dynamics require a separate metric}
Dynamic or non-rigid pixels on a moving object always violate the rigid-scene assumptions. That's why including them in H/F fitting (i) corrupts inlier sets, (ii) destabilizes robust estimation, and (iii) degrades the diagnostic value of IR/GE for the background.
Conversely, judging dynamics by background residuals rewards trivial solutions (e.g., shrinking inlier sets) and unfairly penalizes scenes with legitimate motion.
We therefore adopt a \emph{flow-guided} dynamic score to capture motion plausibility, deformation continuity, and occlusion ordering:
\begin{equation}
    \mathrm{Dyn}_t \;=\; \operatorname{RobAgg}\Bigl( w_i \,\bigl\| \widehat{\mathbf{u}}_i - \mathbf{u}_i \bigr\|_2 \;;\; i\in\mathcal{M}_d \Bigr),
\end{equation}
where $\mathcal{M}_d$ is the dynamic-region mask, $\mathbf{u}_i$ is measured optical flow, $\widehat{\mathbf{u}}_i$ is expected motion under consistency priors, $w_i$ down-weights unreliable/occluded pixels, and $\operatorname{RobAgg}$ denotes a robust aggregator.
We instantiate this metric with an \emph{adaptive Wasserstein-2}~\cite{dukler19wasserstein} regional formulation (AdaDFRC-W2) in \S\ref{sec:method}, which aggregates local motion-pattern discrepancies with occlusion-aware weighting.
This score is \emph{complementary} to background indicators: H/F-based IR/GE certify rigid-scene adequacy, while $\mathrm{Dyn}$ certifies that moving/occluding content evolves plausibly.

\subsection{Intended use, GT-free positioning, and boundaries}

Our protocol is an \emph{engineering QA baseline} for model comparison, regression testing, and content-side risk triage under GT-free conditions.
It complements human-subject studies rather than replacing them by enforcing photogrammetric validity without GT~\cite{li2025geoat,murai2025mast3r}. It provides a \emph{reproducible floor} upon which subjective evaluation can build.

Beyond the practical unavailability of GT in platform settings, GT-free protocols allow cross-platform comparisons on real data without disclosing proprietary trajectories or geometry and avoid mismatches introduced by synthetic GT.
By aligning to photogrammetric self-consistency and releasing audit artifacts, we retain a hard, reproducible constraint that generalizes across devices, content, and generators.

Typical uses include (i) cross-generator comparisons with regime-stratified breakdowns; (ii) pipeline regression tests where gate agreement and sensitivity curves act as sentinels; and (iii) dataset curation by filtering clips with unstable regimes or insufficient background support.
Known limitations include specularities, rolling-shutter distortions, extreme motions with low-SNR correspondences, and non-pinhole optics.

\section{Methodology}
\label{sec:method}

Our methodology is designed to quantitatively answer a fundamental question: whether a generated video adheres to the geometric principles of a rigid 3D world observed by a moving camera. The framework operates without ground truth and is built on a core principle we term \textbf{mask-first, regime-aware geometry}. This principle dictates that we must first isolate independently moving objects from the static background (mask-first), and then apply the geometrically appropriate motion model to the background based on the nature of the camera's movement (regime-aware).

\subsection{Overall Framework}
\begin{figure}[htbp]
    \centering
    \includegraphics[width=0.9\linewidth]{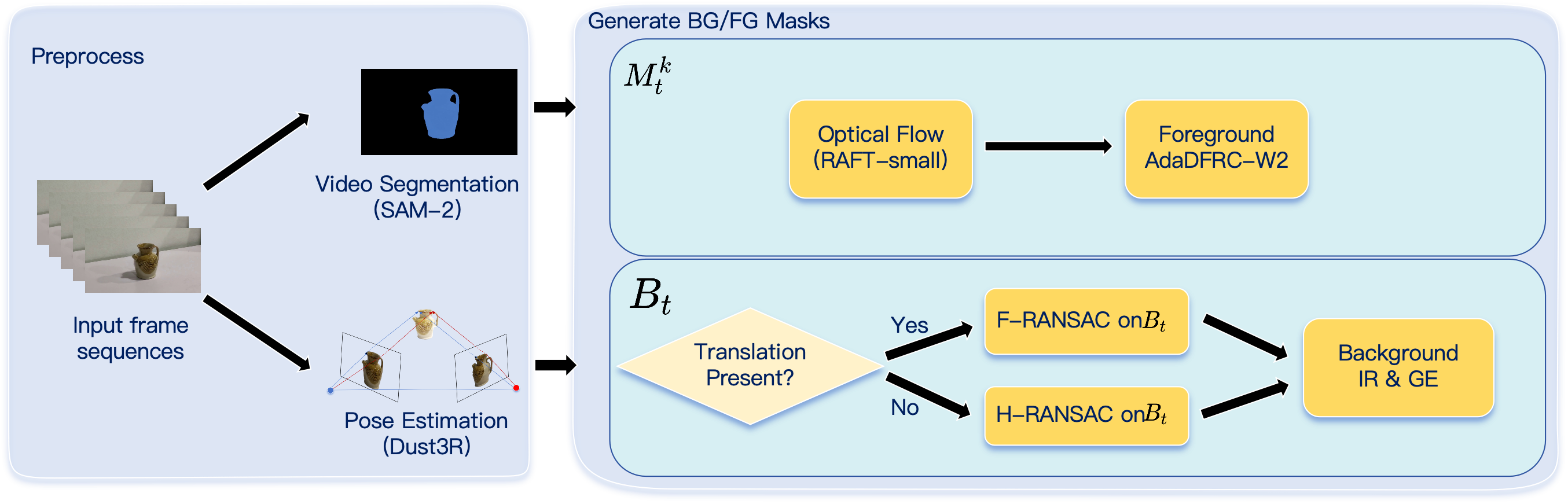}
    \caption{Our evaluation workflow. A pose estimator (DUSt3R) determines the camera motion regime to select between a Fundamental Matrix ($\mathbf{F}$) or Homography ($\mathbf{H}$) model. A segmentation model (SAM-2)~\cite{ravi2024sam2} isolates moving objects. Background consistency is then evaluated on the static regions only, while foreground objects are scored for internal rigidity using AdaDFRC-W2.}
    \label{fig:flowchart}
\end{figure}

Given a generated video sequence $\{I_t\}_{t=0}^{T}$, our evaluation proceeds in three main stages, as illustrated in Fig.~\ref{fig:flowchart}:
\begin{enumerate}
    \item \textbf{Global Motion Regime Gating:} We first determine the dominant camera motion of the input video stream. Specifically, we distinguish between motion that is purely rotational (or static) and motion that includes a translational component. This decision is critical as it dictates which geometric model is appropriate for the background.
    \item \textbf{Foreground/Background Separation:} We segment any objects that are moving independently of the camera's ego-motion. This creates a clean background mask, ensuring that our analysis of the static scene's rigidity is not contaminated by foreground motion.
    \item \textbf{Consistency Evaluation:} We apply distinct, specialized metrics to the separated background and foreground regions. The background is evaluated for rigid-body consistency using classical geometric models, while each moving foreground object is evaluated for its internal geometric integrity.
\end{enumerate}

\subsection{Pose estimation and translation-aware gating}
\label{sec:gating}
In photogrammetry, negligible translation justifies a single-plane/parallax-free background (homography), whereas observable translation requires epipolar geometry (fundamental).
We therefore route by a translation-evidence proxy $s_t$ from pose logs, with a hysteresis band $[\tau_\rho-\varepsilon,\ \tau_\rho+\varepsilon]$ to avoid flip–flop:
$\gamma_t\!=\!\mathrm{H}$ if $s_t<\tau_\rho-\varepsilon$, $\gamma_t\!=\!\mathrm{F}$ if $s_t>\tau_\rho+\varepsilon$, and otherwise we keep $\gamma_{t-1}$.

To select the correct geometric model for the background, we must first characterize the camera's ego-motion. We estimate the relative camera pose $(\mathbf{R}_t, \mathbf{t}_t) \in SE(3)$ between frames. Let $I_t,I_{t+1}$ be adjacent frames. From deep matches $\mathcal{M}_{t,t+1}=\{(x_i,x'_i)\}_{i=1}^{N}$ (pixel coordinates in homogeneous form), we estimate the relative pose $(\mathbf R,\mathbf t)\!\in\!\mathrm{SE}(3)$ using DUSt3R or a PnP solver on the same correspondences~\cite{wang2024dust3r}. Under DUSt3R’s convention, the scene is normalized to unit scale along principal axes. We declare the pair \emph{non-translation-dominant} (behaves like pure rotation) when
\begin{equation}
    \label{eq:tau-t}
    \|\mathbf t\|_\infty \;\triangleq\; \max\{|t_x|,|t_y|,|t_z|\} \;\le\; \tau_t,\qquad \tau_t=0.04,
\end{equation}
and \emph{translation-dominant} otherwise. The choice of $\tau_t$ follows a first-order pinhole analysis: for a 3D point $\mathbf P=(X,Y,Z)^T$ the translation-induced image displacement satisfies
\begin{equation}
\label{eq:parallax}
\|\delta\mathbf x\|\;\lesssim\;\frac{f}{Z}\,\|\mathbf t_\perp\|,
\end{equation}
with focal length $f$ (in pixels) and translation component $\mathbf t_\perp$ orthogonal to the view. Requiring $\|\delta\mathbf x\|\!\le\!\tau_H$ (the inlier tolerance of homography fitting, typically $\sim\!1$\,px at our working resolution) yields $\|\mathbf t_\perp\|\!\le\!(\tau_H/f)\,Z$. With median $Z\!\approx\!1$ in DUSt3R units and calibrated $f$, the practical bound $\|\mathbf t\|_\infty\!\le\!0.04$ keeps $\|\delta\mathbf x\|$ at the noise floor, so a single homography provides an adequate approximation. To stabilize decisions near the boundary, we adopt a narrow hysteresis in implementation (\(\le 0.03\): homography branch; \(\ge 0.05\): fundamental branch); in the gray zone, both models are evaluated, and the one with higher support is selected downstream. Section~\S\ref{subsec:calibration} empirically calibrates this gate by a threshold sweep.

\subsection{Foreground/Background Separation}
A core tenet of our method is that the geometric model for camera ego-motion applies only to the static parts of the scene. Therefore, we must first identify and mask out any independently moving objects. Using a video object segmentation model, we generate a set of masks $\{M_t^k\}$ for each moving object $k$ at frame $t$. The static background mask $B_t$ is then defined as the complement of the union of all foreground masks: $B_t = 1 - \bigcup_k M_t^k$. All subsequent analysis of the background's geometric consistency is performed \textit{exclusively} on feature correspondences found within the regions defined by $B_t$. This "mask-first" approach is crucial as it prevents moving objects from corrupting the estimation of the global camera motion model, which would otherwise lead to incorrect evaluations of background stability.

\subsection{Background Consistency: IR and GE}
We assess the behavior of the \emph{static background} as a single rigid body across consecutive frames by fitting a global geometric model. Specifically, we utilize either a homography $\mathbf{H}$ for the planar/rotation regime or a fundamental matrix $\mathbf{F}$ for the translation/parallax regime.
The analysis is conducted on \emph{background-only} correspondences, employing a robust estimator.
Models are estimated in normalized coordinates for numerical stability and \emph{denormalized to pixels} for residual evaluation. All residuals are computed on the \emph{content mask} (letterboxed padding ignored). Unless otherwise specified, frames are $1024{\times}720$ with letterboxing.

\begin{figure}[htbp]
    \centering
    \includegraphics[width=0.6\linewidth]{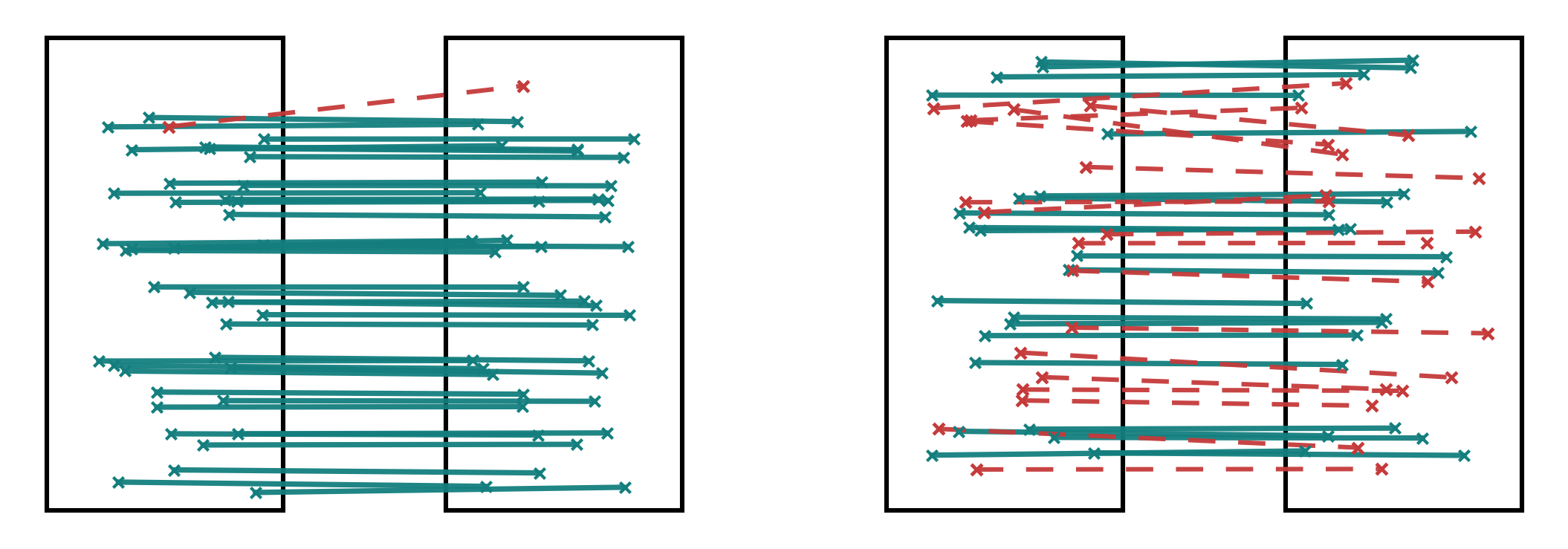}
    \caption{Each panel shows two adjacent frames with matched feature correspondences. Left: high IR—most matches are inliers, indicating a rigid background. Right: low IR—many outliers, revealing background inconsistency.}
    \label{fig:ir_schematic}
\end{figure}

Inlier Ratio (IR) is the fraction of background correspondences whose geometric residual falls below the inlier test used by the robust estimator. A high IR indicates that a single rigid transformation explains the majority of the background motion, as expected for static scenes (cf. Fig.~\ref{fig:ir_schematic}).
Geometric Error (GE) summarizes the \emph{accuracy} of alignment among the inliers only, reported as the median residual. Thus IR answers “does a single global model hold?”, while GE answers “how tightly do the consistent matches align?”. In our implementation, the \emph{same inlier test used during robust estimation} is reused to define the inlier set for reporting.

Homography and epipolar residuals differ in scale and tail behavior; we therefore calibrate \emph{dual thresholds} $(\tau_H,\tau_F)$ rather than forcing a single cutoff.
To make platform-level rankings defensible, we accompany per-clip IR/GE with (i) \textbf{threshold–score sensitivity scans} over $(\tau_H,\tau_F)$ and (ii) \textbf{percentile-aligned} summaries (e.g., reporting $\mathrm{GE}_H$ at matched $\mathrm{IR}_H$ percentiles), which mitigate absolute-scale differences across datasets and generators.

\paragraph{Homography case (non-translation regime).}
In the non-translation regime, we model background motion with a homography \(\mathbf{H}_t\). The error for each correspondence pair $(\mathbf{x}_1, \mathbf{x}_2)$ is evaluated using the \emph{symmetric transfer error}, which averages the forward and backward projection errors in the pixel domain:
\begin{equation}
\label{eq:bg-h-sym}
\hat{\mathbf{x}}_2=\pi(\mathbf{H}_t\,\mathbf{x}_1),\quad
\hat{\mathbf{x}}_1=\pi(\mathbf{H}_t^{-1}\,\mathbf{x}_2),\quad
r_H^{\mathrm{sym}}=\tfrac12\big\|\mathbf{x}_2-\hat{\mathbf{x}}_2\big\|_2+\tfrac12\big\|\mathbf{x}_1-\hat{\mathbf{x}}_1\big\|_2,
\end{equation}
where $\pi([a,b,c]^\top)=[a/c,\,b/c]^\top$ converts from homogeneous to Cartesian coordinates. Using this error metric, the two key consistency scores can be derived. We robustly estimate $\mathbf{H}_t$ (e.g., via PROSAC sampling) by finding the largest set of correspondences that satisfy the inlier condition
\begin{equation}
    r_H^{\mathrm{sym}} \le \tau_H, \quad \text{with } \tau_H = 2.0 ,\text{px}.
\end{equation}
The \textbf{IR} is the fraction of correspondences classified as inliers, and the \textbf{GE} is the median symmetric transfer error among these inliers.

\paragraph{Fundamental case (translation/parallax regime).}
The estimation of the fundamental matrix $\mathbf{F}_t$ is highly sensitive to the scale and origin of pixel coordinates. To ensure numerical stability, we follow the standard practice of coordinate normalization. For a given set of $N$ points $\{\mathbf{x}_i=[u_i, v_i, 1]^\top\}$, we first compute their centroid $\mathbf{c} = [\bar{u}, \bar{v}]^\top$ and a scaling factor $s = \sqrt{2} / \bar{d}$, where $\bar{d} = \frac{1}{N}\sum_i \| [u_i-\bar{u}, v_i-\bar{v}]^\top \|_2$ is the average distance from the centroid. The normalization matrix $\mathbf{T}$ is then constructed as:
\begin{equation}
    \label{eq:t-norm}
    \mathbf{T} = \begin{bmatrix}
    s & 0 & -s\bar{u} \\
    0 & s & -s\bar{v} \\
    0 & 0 & 1
    \end{bmatrix}.
\end{equation}
We compute $\mathbf{T}_1$ and $\mathbf{T}_2$ for the point sets in each image respectively, yielding normalized coordinates $\hat{\mathbf{x}}_1 = \mathbf{T}_1 \mathbf{x}_1$ and $\hat{\mathbf{x}}_2 = \mathbf{T}_2 \mathbf{x}_2$.

In this normalized space, we fit $\hat{\mathbf{F}}_t$ using a robust estimator (again, PROSAC can be used) exclusively in this normalized space. It finds an optimal normalized fundamental matrix $\hat{\mathbf{F}}_t$ by classifying a correspondence as an inlier if its Sampson distance, computed on normalized data, is below the unitless threshold $\tau_F=0.1$. The \textbf{Inlier Ratio (IR)} is the fraction of correspondences in this final inlier set~\cite{hartley2004multiple,fathy2011fundamental}.

Once the inlier set is identified, $\hat{\mathbf{F}}_t$ is transformed back to the original pixel domain via denormalization:
\begin{equation}
    \label{eq:f-denorm}
    \mathbf{F}_t = \mathbf{T}_2^\top \hat{\mathbf{F}}_t \mathbf{T}_1.
\end{equation}
Finally, the \textbf{Geometric Error (GE)} is defined as the \emph{median} of the Sampson distances computed for all inlier pairs. This computation uses their original pixel coordinates $(\mathbf{x}_1, \mathbf{x}_2)$ and the denormalized matrix $\mathbf{F}_t$:
\begin{equation}
    \label{eq:bg-f-sampson}
    \text{GE} = \underset{(\mathbf{x}_1, \mathbf{x}_2) \in \text{Inliers}}{\text{median}} \left\{ \frac{|\mathbf{x}_2^{\!\top}\mathbf{F}_t\mathbf{x}_1|}{\sqrt{(\mathbf{a})_1^2+(\mathbf{a})_2^2+(\mathbf{b})_1^2+(\mathbf{b})_2^2}} \right\},
\end{equation}
where $\mathbf{a}=\mathbf{F}_t\mathbf{x}_1$ and $\mathbf{b}=\mathbf{F}_t^{\!\top}\mathbf{x}_2$. This two-stage process ensures that the inlier selection is robust and independent of image resolution, while the final reported GE remains interpretable in the context of pixel-level deviations.


For ${\bf H}$ we use symmetric transfer error in \emph{pixels} with threshold $\TauH$; for ${\bf F}$ we use the \emph{normalized} Sampson error with threshold $\TauF$ (unitless). The dual-threshold design reflects fundamentally different residual scales and avoids ill-posed single-threshold comparisons.
$\IR$ is the fraction of correspondences within the relevant threshold; $\GE$ is the median residual over inliers. Both metrics use the \emph{same per-family threshold} to avoid moving standards.
For brevity, in \eqref{eq:bg-ir-ge} we write a generic cutoff $\delta$ which denotes $\TauH$ when $\gamma_t=\mathrm{H}$ and $\TauF$ when $\gamma_t=\mathrm{F}$.

Let $\mathcal{C}^{\mathrm{bg}}_t$ be the background correspondences for pair $(I_t,I_{t+1})$. We dispatch to $\mathbf{F}$ (translation) or $\mathbf{H}$ (non-translation) using a translation surrogate $\tilde{\rho}_t$ with threshold $\tau_\rho$ (\S\ref{sec:gating}). Denote the residual of a match by $r_i$ (either $r_H^{\mathrm{sym}}$ or $r_F$). With a \emph{single global} inlier threshold $\delta$ (pixels), the inlier set and the per-pair metrics are
\begin{equation}
\label{eq:bg-ir-ge}
    \mathcal{I}_t=\{\,i\in\mathcal{C}^{\mathrm{bg}}_t\mid r_i\le \delta\,\},\qquad
    \mathrm{IR}_{\mathrm{BG}}(t)=\frac{|\mathcal{I}_t|}{|\mathcal{C}^{\mathrm{bg}}_t|},\qquad
    \mathrm{GE}_{\mathrm{BG}}(t)=\operatorname{median}\{\,r_i\mid i\in\mathcal{I}_t\,\}.
\end{equation}
Pairs with too few inliers (e.g., $|\mathcal{I}_t|<30$ due to extreme blur/low texture) are marked invalid and excluded from aggregation.

For a clip, we summarize background consistency by robust medians over valid pairs:
\begin{equation}
\label{eq:bg-clip-agg}
    \mathrm{IR}_{\mathrm{BG}}^{\text{clip}}=\operatorname{median}_t\big(\mathrm{IR}_{\mathrm{BG}}(t)\big),\qquad
    \mathrm{GE}_{\mathrm{BG}}^{\text{clip}}=\operatorname{median}_t\big(\mathrm{GE}_{\mathrm{BG}}(t)\big).
\end{equation}
Scene-balanced medians and 95\% bootstrap CIs are reported across seeds within each category.

Within each category ($C1$–$C6$), we aggregate per-seed clip scores by a \emph{scene-balanced} median, i.e., each seed (scene) contributes one score with equal weight: $\tilde m_{Ck}=\operatorname{median}_{s}(m_{s})$. To quantify uncertainty around the median, we report \emph{95\% nonparametric bootstrap confidence intervals} by resampling seeds with replacement ($B{=}10{,}000$ replicates) and taking the 2.5/97.5 percentiles of the bootstrap distribution of $\tilde m_{Ck}$. This avoids distributional assumptions and prevents categories or seeds with more regenerations from dominating the estimate.
We use a family-specific inlier gate during robust fitting. Specifically, we use symmetric transfer error with $\tau_H{=}2$\,px for a homography situation, and normalized Sampson with $\tau_F{=}0.1$ at the fundamental case. At report time, we apply a pixel-domain gate on GE fixed at $\delta{=}\mathbf{3.5}$\,px.

A geometrically consistent background exhibits \emph{high} $\mathrm{IR}_{\mathrm{BG}}$ and \emph{low} $\mathrm{GE}_{\mathrm{BG}}$. Drops in IR indicate that a single global model cannot explain the pair (global inconsistency), while increases in GE at stable IR indicate imprecise alignment among otherwise consistent matches.

\subsection{Foreground Consistency: AdaDFRC-W2}
\label{sec:adadfrc}
Dynamic/non-rigid pixels violate rigid-background assumptions.
Including them in H/F fitting (i) pollutes inlier sets, (ii) destabilizes robust estimation, and (iii) degrades the diagnostic value of IR/GE for the background.
Conversely, judging dynamics by background residuals rewards trivial solutions (shrinking inlier sets) and unfairly penalizes legitimate motion.
A flow-guided score such as AdaDFRC-W2 directly targets motion plausibility, deformation continuity, and occlusion ordering, \emph{complementing} IR/GE rather than conflating objectives.

To assess the geometric integrity of an individual moving object, which standard metrics fail to capture, we introduce the \textbf{Adaptive Dense Flow Regional Consistency (AdaDFRC-W2)} metric. This metric quantifies how well the internal motion of an object adheres to a rigid or near-rigid transformation, a key property of real-world objects.

\paragraph{Theoretical Foundation}
The core principle of AdaDFRC-W2 is derived from the observation that the motion of a rigid object, when projected onto a 2D image plane under perspective projection, can be locally approximated by a 2D affine transformation. This approximation holds particularly well for objects that are relatively small in the field of view or distant from the camera. Consequently, for a geometrically consistent, rigid moving object, we expect its internal motion field—as captured by dense optical flow—to be highly structured and conform well to a single, dominant affine motion model. Any significant deviation from this model indicates non-rigid deformation, which in the context of AIGC, often manifests as generative artifacts like shearing, tearing, or texture "boiling." AdaDFRC-W2 is designed to measure the magnitude of this deviation precisely.

\paragraph{Derivation and Methodology}
For each independently moving object $k$, identified by its mask $M_t^k$ at time $t$, we compute its AdaDFRC-W2 score through a systematic, multi-step process detailed below. This process is designed to be robust and invariant to the object's on-screen size and position.

\begin{enumerate}
    \item \textbf{Adaptive Grid Normalization:} To standardize the analysis regardless of the object's scale, we first define a bounding box around the mask $M_t^k$. This bounding box is then partitioned into a uniform $N \times N$ grid of cells. This adaptive grid ensures that our analysis has a consistent spatial resolution relative to the object itself.

    \item \textbf{Robust Local Motion Summarization:} We compute a dense optical flow field from frame $I_t$ to $I_{t+1}$ within the object's bounding box. For each grid cell $(i,j)$, we identify all the object pixels (as defined by $M_t^k$) that fall within it. A single, robust motion vector $\mathbf{v}_{i,j} \in \mathbb{R}^2$ is then computed for this cell by taking the median of the flow vectors of all its constituent pixels. The median is chosen for its robustness to outliers, effectively filtering out minor noise in the flow estimation. If a cell contains no object pixels, it is excluded from further analysis.

    \item \textbf{Optimal Affine Model Fitting via Weighted Least-Squares:} The central step is to find the single affine transformation that best describes the collection of local motion vectors $\{\mathbf{v}_{i,j}\}$. An affine transformation models the motion of a point $\mathbf{p}$ as $\mathbf{p}' = \mathbf{A}\mathbf{p} + \mathbf{b}$, where $\mathbf{A}$ is a $2 \times 2$ matrix representing rotation, scaling, and shear, and $\mathbf{b}$ is a $2 \times 1$ translation vector. The motion vector is thus predicted as $\mathbf{v}_{\text{pred}}(\mathbf{p}) = \mathbf{p}' - \mathbf{p} = (\mathbf{A}-\mathbf{I})\mathbf{p} + \mathbf{b}$.

    We formulate this as a weighted least-squares problem. We seek the optimal affine parameters $(\mathbf{A}^*, \mathbf{b}^*)$ that minimize the sum of squared differences between the predicted motion and the observed local motion vectors $\{\mathbf{v}_{i,j}\}$. To give more importance to denser parts of the object, each cell $(i,j)$ is assigned a weight $w_{i,j}$ equal to the number of object pixels it contains. The optimization problem is thus:
    \begin{equation}
        (\mathbf{A}^*, \mathbf{b}^*) = \arg\min_{\mathbf{A}, \mathbf{b}} \sum_{i,j} w_{i,j} \|\, (\mathbf{A} \mathbf{p}_{i,j} + \mathbf{b}) - \mathbf{v}_{i,j} \,\|_2^2
    \end{equation}
    where $\mathbf{p}_{i,j}$ is the geometric centroid of the grid cell $(i,j)$. This is a standard linear least-squares problem that can be solved efficiently in closed form.

    \item \textbf{Score Calculation as Weighted Mean Squared Error (WMSE):} The AdaDFRC-W2 score for the object $k$ at time $t$, denoted $\mathcal{D}_t^{(k)}$, is defined as the weighted mean squared error (WMSE) of this optimal fit. This value represents the residual, non-affine component of the motion field—the part that cannot be explained by a single rigid transformation.
    \begin{equation}
        \label{eq:dfrc}
        \mathcal{D}_t^{(k)} = \frac{\sum_{i,j} w_{i,j} \|\, (\mathbf{A}^* \mathbf{p}_{i,j} + \mathbf{b}^*) - \mathbf{v}_{i,j} \,\|_2^2}{\sum_{i,j} w_{i,j}}
    \end{equation}
\end{enumerate}
A low $\mathcal{D}_t^{(k)}$ score signifies high internal consistency, indicating that the object's motion is well-described by a single affine model, which is characteristic of a rigid body. Conversely, a high score points to significant non-rigid deformation or other structural inconsistencies symptomatic of generative artifacts. The final consistency score for an object instance, $\mathcal{S}^{(k)}$, is computed by averaging $\mathcal{D}_t^{(k)}$ over its entire visible trajectory.


\section{GeoCon-Bench Dataset: A Sophisticated Collection for Geometric Consistency}
\label{sec:dataset}
To transition geometric consistency from a qualitative afterthought to a first-class, quantitative evaluation target, we construct \textbf{GeoCon-Bench}, a standard, reusable, and expandable benchmark explicitly tailored to the challenges of geometry consistency among various prompt situations in AIGC videos. It is the first benchmark of its kind to systematically factorize geometric stressors, providing a controlled environment to probe model capabilities beyond semantic and aesthetic appeal. The dataset is designed to (i) rigorously test whether a single rigid motion model can explain frame-to-frame changes, (ii) isolate independently moving objects for granular, per-object consistency analysis, and (iii) directly align with and validate our translation-aware gating mechanism.

\subsection{Design Principles}
Our benchmark is built on five core principles to ensure fairness, rigor, and utility:
\begin{itemize}
    \item \textbf{Geometry-first, not aesthetics:} Prompts and scenes are constructed to emphasize camera motion, scene structure, and object rigidity over complex textures or artistic styles. This prevents models from hiding geometric flaws behind stylistic appeal and focuses the evaluation on structural integrity.
    \item \textbf{Orthogonal stressors:} Scenarios are carefully factored along controllable geometric axes (e.g., planarity of the background, presence of parallax, number of independent motion fields). This allows for targeted analysis of specific model weaknesses, such as a failure to model epipolar geometry while succeeding at projective transforms.
    \item \textbf{Model-agnostic generation:} To ensure a fair comparison, identical prompts and random seeds are used across all tested models (Sora~\cite{liu2024sora}, Runway, Wan~\cite{wan2025}). Technical parameters such as resolution, frames per second (fps), and clip duration are also held constant.
    \item \textbf{Reproducibility:} Every generated sequence is accompanied by a detailed metadata record (prompt, seed, model version, resolution, fps, duration) and the scripts used for generation. This commitment to transparency allows for full verification and extension of our results.
    \item \textbf{Replaceable components:} The tools used in our evaluation pipeline (segmentation, feature matching, optical flow) are treated as default choices, not fixed requirements. The benchmark is structured to allow for drop-in alternatives (e.g., swapping RAFT for a future optical flow model), ensuring its long-term relevance as underlying technologies evolve.
\end{itemize}

\subsection{Scenario Taxonomy}
The dataset is partitioned into six distinct scenarios. We first include the \textbf{Static Scene} as a negative-control baseline. Its purpose is not to test complex 3D understanding but to evaluate temporal stability and the fidelity of executing a zero-motion command. It reveals texture flicker, unintended camera drift, and minor object instability. For a perfect static video, camera motion should be zero and appearance-consistency between frames should approach zero; any deviation directly quantifies inherent instability. Failing this simplest case indicates that errors in more complex scenarios likely stem from a lack of temporal coherence rather than motion understanding.

\begin{enumerate}[label=\arabic*)]
    \item \textbf{Static Scene (Temporal Stability Baseline):} see above.
    \item \textbf{Pure Camera Rotation (Homography-dominant):} A static 3D scene where the camera rotates around its optical center. This is a foundational test of projective consistency.
    \textit{Expected Behavior:} The motion should be classified as \emph{non-translation-dominant}. A single homography ($H$) should explain nearly all background feature matches, resulting in an Inlier Ratio (IR) approaching 100\% and a very low Geometric Error (GE).
    
    \item \textbf{Pure Camera Translation (Epipolar-dominant):} A static scene with significant depth variations (e.g., near and far objects) where the camera translates laterally or forward. This is a critical test of a model's ability to synthesize motion parallax.
    \textit{Expected Behavior:} The motion must be classified as \emph{translation-dominant}. A single fundamental matrix ($F$) should model the epipolar geometry, yielding a high IR and low GE. Failure often manifests as scene "breathing" or warping, where distant objects do not remain stable relative to near ones.

    \item \textbf{Static Background + Single Moving Object:} A static camera observes a scene where a single foreground object moves independently. This scenario tests the model's ability to disentangle different motion fields.
    \textit{Expected Behavior:} The background should remain perfectly rigid, passing the $H/F$ test with high scores. The foreground object is evaluated for internal rigidity using AdaDFRC-W2 (Eq.~\eqref{eq:dfrc}); a low score indicates a rigid object, while a high score reveals non-rigid deformations like stretching or shearing.

    \item \textbf{Complex Camera Motion (Rotation+Translation):} Realistic, combined camera movements such as a dolly-zoom or arc shot. This probes the model's ability to maintain a coherent 3D representation under more challenging, free-form motion.
    \textit{Expected Behavior:} The motion should be classified as \emph{translation-dominant}. The evaluation checks if a single fundamental matrix ($F$) can still robustly model the background motion, despite its complexity, and if foreground objects (if any) can be stably decomposed and tracked.

    \item \textbf{Geometric Stress Tests:} Adversarial but physically-plausible setups designed to push models to their limits. This includes scenes with thin, detailed structures (prone to disappearing), reflective or transparent surfaces (challenging for feature matching), and occlusions.
    \textit{Expected Behavior:} We anticipate a significant drop in performance across all metrics. These scenarios are diagnostic, designed to reveal failure modes like structural "breathing," object wobble, or non-rigid drift that might not be apparent in simpler scenes.
\end{enumerate}

\subsection{Prompt Templates and Lexicon}
To ensure precise control over the generated geometry and minimize semantic ambiguity, we adopt a structured, modular prompt system. This forces the models to contend with the geometric request directly, rather than relying on semantic shortcuts. Each prompt uses four slots: \texttt{Motion Command} + \texttt{Scene Composition} + \texttt{Object Specification} + \texttt{Qualitative Style}. A comprehensive prompt lexicon, detailing the allowed vocabulary for each slot, is provided in an appendix (cf. Appendix~\ref{apx:lexicon}) to standardize wording.

\paragraph{Examples (abbrev.).}
\begin{itemize}
    \item \textbf{Pure Rotation:} \emph{“Single static shot. The camera rotates in place around the building facade. No translation. Background absolutely fixed.”}
    \item \textbf{Pure Translation:} \emph{“Single static shot. Camera dollies right past trees and a building, revealing parallax (near vs far).”}
    \item \textbf{Static BG + Object:} \emph{“Static camera. A red car drives a circular path in front of a fixed building. Background must stay rigid.”}
\end{itemize}

\subsection{Generation Protocol and Dataset Statistics}
All clips in GeoCon-Bench are generated under a strict protocol. In addition to the released seed-image scenes, we include standardized videos from three representative AIGC models, including \emph{Sora} (commercial), \emph{Runway} (industry-leading), and \emph{Wan} (open-source). Each clip is generated at a fixed resolution and 24 fps, with a duration of 5 seconds (120 frames). For each of the 20 unique prompts, we generate 3 clips per model using different seeds. We advocate for \emph{scene-balanced} reporting, where metrics are first averaged within each scenario before being averaged across scenarios. This prevents a model's high performance on an easy category from masking its failures on more challenging ones.

\subsection{Metadata and Release Format}
To maximize the benchmark's utility and promote reproducible science, each clip is accompanied by a JSON metadata file containing: \texttt{\{prompt, model, model\_version, seed, resolution, fps, duration, clip\_id, masks\_path\}}. We will publicly release (i) all prompts, (ii) all generated videos, (iii) all extracted foreground masks, (iv) evaluation configurations, and (v) scripts to reproduce all tables and figures in this paper.

\subsection{Grounded Validation}
To anchor our metrics and calibrate key thresholds, the benchmark includes a small subset of "grounded" scenarios where the intended geometry is unambiguous. These include textured planar surfaces for rotation checks (which should yield a perfect homography with IR=100\%) and scenes with clearly separated near/far layers for translation checks. These anchors provide a sanity check for the entire evaluation pipeline and are used to calibrate the gating threshold $\tau_\rho$ by observing the IR saturation across these known scenarios.

\section{Experimental Validation}
\label{sec:experiments}
We conducted a comprehensive set of experiments to validate our proposed framework and evaluate the geometric consistency of state-of-the-art video generation models using a custom benchmark, GeoCon-Bench. The goal is not merely to rank models but to establish a standardized, interpretable procedure for judging geometric fidelity.

We investigate:
\textbf{RQ1 (Gate validity)}: Does translation-evidence gating agree with photogrammetric fits across scenes and platforms?
\textbf{RQ2 (Background rigidity)}: Are $\mathrm{IR}_{H/F}$ and $\mathrm{GE}_{H/F}$ stable under dual-threshold scanning and informative for cross-platform comparison?
\textbf{RQ3 (Dynamic coherence)}: Does the flow-guided dynamic score (AdaDFRC-W2) capture complementary failures that background indicators cannot?
\textbf{RQ4 (Comparability \& auditing)}: Do percentile-aligned summaries and audit cards yield stable platform rankings and reproducible diagnostics?

\subsection{Implementation Details}
This section details the specific choices of algorithms and parameters used to implement our framework, clearly separating the practical implementation (cf. pseudocode Algorithm~\ref{alg:flowchart}) from the theoretical methodology of Section~\ref{sec:method}.

The abstract components of our framework were realized with the following state-of-the-art models:
\begin{itemize}
    \item \textbf{Pose Estimation:} We used the official implementation of \textbf{DUSt3R} for pairwise relative pose estimation. Its ability to handle uncalibrated images makes it robust for AIGC evaluation.
    \item \textbf{Video Segmentation:} Foreground object masks were generated using \textbf{SAM-2}, leveraging its streaming architecture for efficient video processing.
    \item \textbf{Optical Flow:} Dense flow fields for the AdaDFRC-W2 metric were computed with Recurrent All-Pairs Field Transforms (\textbf{RAFT-small}), chosen for its balance of accuracy and efficiency.
\end{itemize}

\begin{algorithm}[t]
\small
\DontPrintSemicolon
\SetKwInOut{Input}{Input}
\SetKwInOut{Output}{Output}
\Input{
Seed $I_0$; generated frames $I_{1:T}$; scales $S=\{1/4,1/2,1\}$;\\
DUSt3R translation gate $\tau_t$ (default $\mathbf{0.04}$); fixed background gates $\TauH{=}2$\,px, $\TauF{=}0.1$.}
\Output{$\mathrm{IR}_{\mathrm{BG}},\mathrm{GE}_{\mathrm{BG}}$; per-object $\mathcal{S}^{(O)}$; detected category.}

\textbf{A. DUSt3R translation decision (first--last)}\;
estimate pose $\mathbf{T}=(\mathbf{R},\mathbf{t})$ between $(I_0,I_T)$;\;
\lIf{pose available}{ set $t_\infty\!\leftarrow\!\|\mathbf{t}\|_\infty$, \ \texttt{translation\_flag}$\leftarrow(t_\infty>\tau_t)$ }
\lElse{ \texttt{translation\_flag}$\leftarrow$\texttt{F} \ (present-only policy) }

\textbf{B. SAM-2 segmentation (mask-first)}\;
run SAM-2 $\rightarrow \{M_t^k\}$; define content mask (remove letterbox) and background $B_t=\text{content}(I_t)\setminus\cup_k M_t^k$;\;
build background-only correspondences $\mathcal{C}^{\mathrm{bg}}_{t,s}$ on $B_t\cap B_{t+1}$ for all $t$ and $s\in S$.\;

\textbf{C. Background geometry on masks-outside}\;
\For{$t=0$ \KwTo $T-1$}{
  \For{$s\in S$}{
    \eIf{\texttt{translation\_flag}}{
       fit $\mathbf{F}_{t,s}$ on $\mathcal{C}^{\mathrm{bg}}_{t,s}$ (robust, gate $\TauF$); compute IR$_F$ and median Sampson residual GE$_F$ (pixels);\;
    }{
       fit $\mathbf{H}_{t,s}$ on $\mathcal{C}^{\mathrm{bg}}_{t,s}$ (robust, gate $\TauH$); compute IR$_H$ and median symmetric reprojection GE$_H$ (pixels);\;
    }
  }
}
aggregate over time/scales by robust medians to obtain $\mathrm{IR}_{\mathrm{BG}},\mathrm{GE}_{\mathrm{BG}}$;\;
\textit{(optional)} also record best-of(H/F) per pair for analysis, without affecting the main scores.\;

\textbf{D. Movers: AdaDFRC-W2}\;
\For{each object $O$ and $s\in S$}{
  estimate dense flow and occlusion inside $O$; grid the bbox, take per-cell median flow $v_{i,j}$;\;
  fit affine $(\mathbf{A},\mathbf{b})$ and compute weighted MSE per frame $\mathcal{D}_t^{(O)}$; average over the track $\rightarrow \mathcal{S}^{(O)}$.\;
}

\textbf{E. Post-hoc categorization}\;
combine \texttt{translation\_flag} with $(\mathrm{IR}_{\mathrm{BG}},\mathrm{GE}_{\mathrm{BG}})$ under fixed report gates (IR$\ge 0.97$, GE$\le 3.5$\,px) to assign one of \\
\{\texttt{Static}, \texttt{Rotation-dominant}, \texttt{Translation-dominant}, \texttt{StaticCam+MovingObjs}, \texttt{Complex}, \texttt{Stress}\}.\;

\Return $\mathrm{IR}_{\mathrm{BG}},\mathrm{GE}_{\mathrm{BG}},\{\mathcal{S}^{(O)}\}$.\;

\caption{Mask-first, translation-aware evaluation. Residuals are evaluated in \emph{pixels} on the content mask; the background model (H or F) is chosen once from the DUSt3R decision on $(I_0,I_T)$.}
\label{alg:flowchart}
\end{algorithm}

\subsection{DUSt3R Translation-Gate Calibration}
\label{subsec:calibration}

Given the DUSt3R relative pose between the first/last frames, we declare a pair \emph{translation-dominant} (F) if
$\|\mathbf{t}\|_\infty > \tau_t$ and rotation/planar-like (H) otherwise, where
$t_\infty\!\triangleq\!\|\mathbf{t}\|_\infty\!=\!\max(|t_x|,|t_y|,|t_z|)$.
Throughout the paper, background inlier gates are kept fixed to empirically chosen constants and are \emph{not} ablated:
$\TauH{=}2$\,px (homography reprojection) and $\TauF{=}0.1$ (Sampson residual); when background consistency is required we use
$\mathrm{IR}\!\ge\!0.97$ and $\mathrm{GE}\!\le\!3.5$\,px on the content mask (pixels, after denormalization).

For the calibration of $\tau_t$ we use \emph{present-only} pairs (i.e., DUSt3R poses available; ``fail'' pairs have undefined $t_\infty$), across all three platforms and all six prompt families.
The sampling unit is a \textit{(seed, platform, family)} triplet, yielding \textbf{$N{=}300$} observations with valid $t_\infty$
(\textbf{146} F-expected: C3/C5/C6; \textbf{154} H-expected: C1/C2/C4).
This analysis is purely geometric and independent of semantic scores.

\begin{figure*}[t]
  \centering
  \begin{subfigure}[t]{.48\linewidth}
    \centering
    \includegraphics[width=\linewidth]{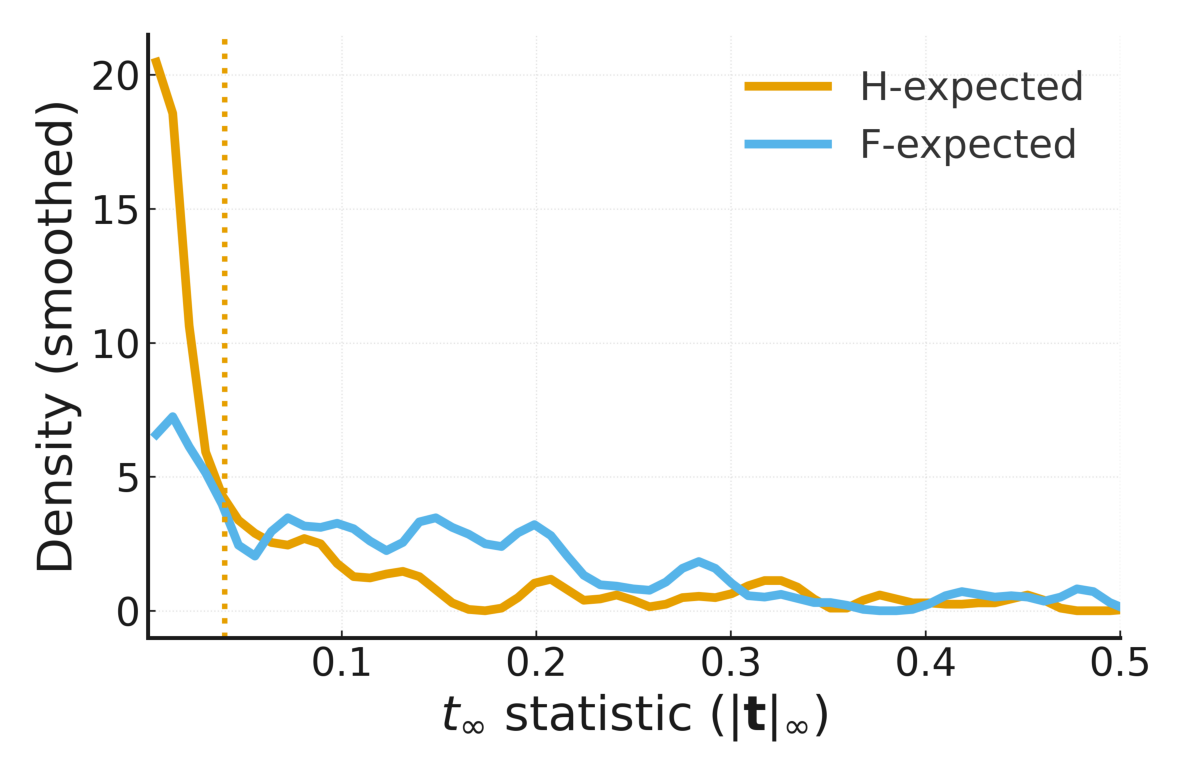}
    \caption{$t_\infty$ density for H-expected (C1/2/4) vs.\ F-expected (C3/5/6).
    A valley appears near $t_\infty\!\approx\!0.04$, indicating a natural split for $\tau_t$.}
  \end{subfigure}\hfill
  \begin{subfigure}[t]{.48\linewidth}
    \centering
    \includegraphics[width=\linewidth]{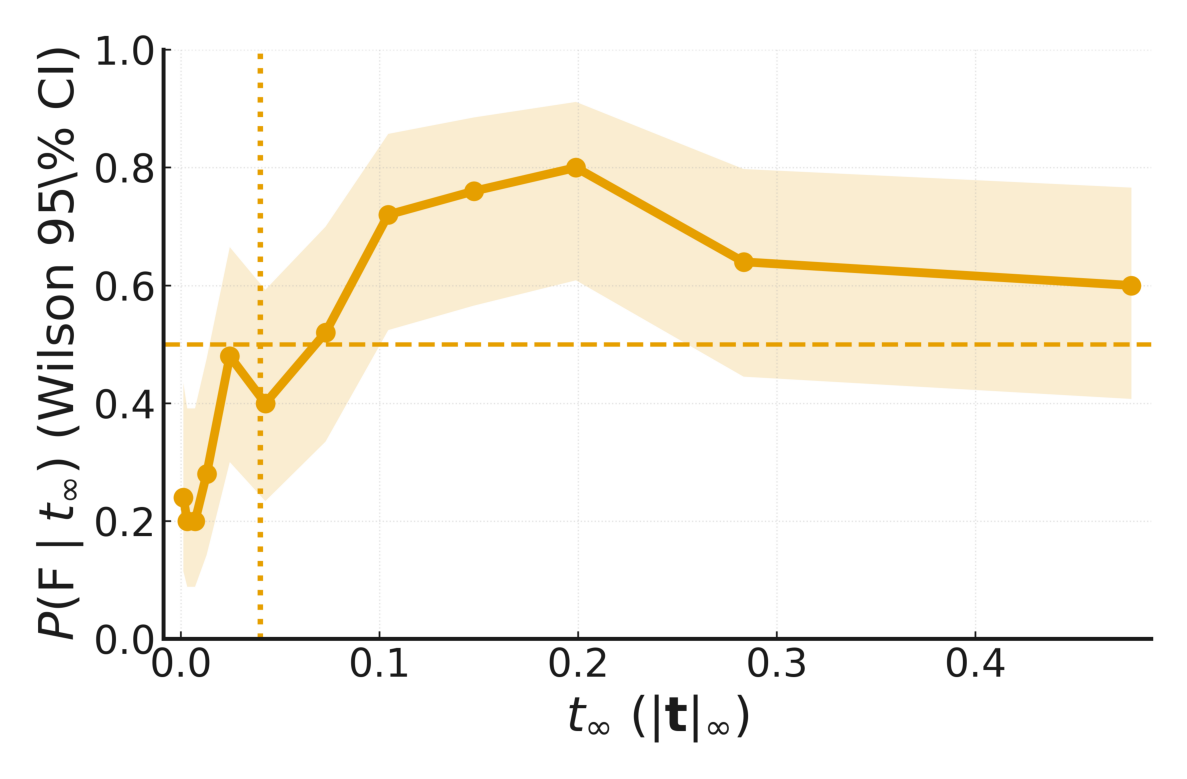}
    \caption{Empirical calibration $P(\mathrm{F}\mid t_\infty)$ with Wilson 95\% CIs (equal-count bins; $\approx$25 samples/bin).
    The 0.5 crossing occurs within $[0.04,0.07]$; $\tau_t{=}0.04$ lies inside this transition band.}
  \end{subfigure}
  \caption{Translation-gate calibration for pose estimation.}
  \label{fig:tau_t_pick}
\end{figure*}

Fig.~\ref{fig:tau_t_pick}(a) shows smoothed densities of $t_\infty$ for H-expected (C1/2/4) and F-expected (C3/5/6) families.
Densities are reported as equal-width histograms over the central 99\% range with a mild Gaussian smoothing (for visualization only; no decision is made on smoothed values).
A clear valley emerges around $t_\infty\!\approx\!0.04$, indicating a natural separation between the two regimes; choosing $\tau_t$ in that valley minimizes overlap of the class-conditional distributions.

To quantify the separation without assuming parametric forms, we estimate the empirical calibration curve
$P(\mathrm{F}\mid t_\infty)$ using equal-count binning and Wilson 95\% confidence intervals (CIs):
for each bin $b$ with $m_b$ samples and $\hat p_b$ the observed F fraction, we report
\begin{equation}
    \bigl[\tfrac{\hat p_b+\tfrac{z^2}{2m_b}}{1+\tfrac{z^2}{m_b}} \pm
\tfrac{z}{1+\tfrac{z^2}{m_b}}\sqrt{\tfrac{\hat p_b(1-\hat p_b)}{m_b}+ \tfrac{z^2}{4m_b^2}}\bigr],\quad z{=}1.96.
\end{equation}
Fig.~\ref{fig:tau_t_pick}(b) exhibits a smooth transition from H-like to F-like regimes.
With $K{=}12$ equal-count bins (about 25 samples per bin), the 0.5 crossing lies in the interval
$t_\infty\!\in\![\mathbf{0.04},\,\mathbf{0.07}]$; notably, the bin centered at $t_\infty\!\approx\!0.043$ has
$\hat p{=}0.40$ with a Wilson CI that \emph{contains} 0.5, while the next bin centered at $\approx\!0.073$ has
$\hat p{=}0.52$ (CI also overlapping 0.5). Hence, $\tau_t{=}\mathbf{0.04}$ lies \emph{inside} the statistically
identified transition band, on the conservative side of the equiprobable boundary.

We therefore adopt $\tau_t{=}\mathbf{0.04}$ as the default translation gate:
(i) it falls in the density valley separating the H/F families, thereby reducing class overlap; and
(ii) it lies within the empirical calibration band where $P(\mathrm{F}\mid t_\infty)\!\approx\!0.5$,
providing a principled knee-point that balances recall on F-expected cases (C3/5/6) against false positives on H-expected ones (C1/2/4).
Complementary analyses in the Appendix~\ref{app:tau_t_more} show that
$\tau_t\!\in\![0.04,0.06]$ forms a broad performance plateau, and that $\tau_t{=}0.04$ is either at or statistically indistinguishable from the maxima while being slightly more conservative on spurious F.

\subsection{GeoCon Benchmark}
\label{subsec:geocon}

We evaluate whether camera \emph{translation} is dominant (``F'') or not (``H'') across three video-generation platforms (Runway, Sora, Wan) and six prompt categories.
For each clip, we feed the \emph{first} and \emph{last} frames to \texttt{DUSt3R} and extract the $4{\times}4$ relative pose; we compute Eq.~\ref{eq:tau-t} from the translation column and declare translation-dominant (''F'') iff $\|t\|_{\infty}>\tau$ with $\tau{=}0.04$; otherwise we label ''H''. Unless otherwise stated, all descriptive statistics below are computed over the \emph{available} clip–pairs (i.e., no imputation), while some visualizations additionally show a conservative variant that treats missing pairs as F.
For reporting raw coverage, we count a pair as “present” only if DUSt3R returns a valid pose. Thus, the Sora split exhibits fewer “present” pairs not because videos were absent, but because DUSt3R could not reconstruct a larger fraction of first/last-frame pairs. In all downstream analyses that depend on semantic consistency, these non-reconstructed pairs are treated as F by construction.
Unless stated, statistics are computed over \emph{present-only} pairs (DUSt3R pose available). For robustness audits, we also report a conservative variant that treats missing pairs as \textbf{F}, reflecting that large endpoint displacement often prevents pose recovery.

We evaluate three platforms (Runway, Sora, Wan) over six categories and multiple seeds. 
DUSt3R returns valid poses for $108$ pairs on Runway, $92$ on Sora, and $100$ on Wan. 
The number of distinct seeds with at least one valid pair is $20$ (Runway), $19$ (Sora), and $20$ (Wan), so the maximum possible pairs per platform are $120/114/120$, leaving $12/22/20$ missing (failure) pairs, respectively. 
Per-category coverage (present pairs divided by \#seeds) is:
Runway $\{100,\,85,\,95,\,85,\,95,\,80\}\%$ for C1–C6; 
Sora $\{84.2,\,84.2,\,78.9,\,78.9,\,78.9,\,78.9\}\%$; 
Wan $\{90,\,80,\,80,\,95,\,85,\,70\}\%$.

On present pairs, the distributions are: 
Runway (count $108$): mean $0.093$, median $0.042$, max $0.580$; 
Sora (count $92$): mean $0.129$, median $0.050$, max $0.691$; 
Wan (count $100$): mean $0.125$, median $0.076$, max $0.548$. 
Overall, $21/300=7.0\%$ of decisions lie within $\pm0.01$ of the threshold $\tau{=}0.04$ (Runway $8.3\%$, Sora $5.4\%$, Wan $7.0\%$), indicating a non-negligible band of near-threshold cases.
DUSt3R failures indicate excessive translation and are treated as translation-present; such pairs are excluded from present-only aggregation but counted in all-pairs coverage.

As expected, semantics intended to contain translation (C3, C5, C6) yield higher F-rates, while C1/C2/C4 lean toward H. 
Per-platform F-rates (fraction of F among present pairs) are:
\begin{itemize}
\item \textbf{Runway}: C1 $30.0\%$, C2 $58.8\%$, C3 $84.2\%$, C4 $11.8\%$, C5 $68.4\%$, C6 $56.2\%$; overall $51.9\%$.
\item \textbf{Sora}: C1 $25.0\%$, C2 $50.0\%$, C3 $46.7\%$, C4 $46.7\%$, C5 $66.7\%$, C6 $80.0\%$; overall $52.2\%$.
\item \textbf{Wan}: C1 $33.3\%$, C2 $62.5\%$, C3 $87.5\%$, C4 $26.3\%$, C5 $94.1\%$, C6 $64.3\%$; overall $60.0\%$.
\end{itemize}
Notably, Sora’s C3 is closer to the decision boundary (only $46.7\%$ F), whereas Wan is extremely translation-heavy on C5 ($94.1\%$ F).

\emph{By design}, categories C1 (Static), C2 (Rotation-dominant), and C4 (Static camera + moving object) are expected to be \textbf{H}-majority (translation-free at the first/last endpoints), whereas C3/C5/C6 are expected to be \textbf{F}-majority. 
Empirically, we observe \textbf{C1 and C4 behave as expected (H-majority).} F-rates for C1 are $30.0\%$/\,$25.0\%$/\,$33.3\%$ (Runway/Sora/Wan); for C4 they are $11.8\%$/\,$46.7\%$/\,$26.3\%$. 
Thus, C1 is consistently H-majority; C4 is H-majority on Runway/Wan and borderline on Sora.
\textbf{C2 deviates from expectation.} Despite the rotation-only intent, C2 shows elevated F: $58.8\%$/\,$50.0\%$/\,$62.5\%$ (Runway/Sora/Wan), i.e., not H-majority on Runway/Wan and exactly balanced on Sora. 
This suggests either residual translation between endpoints, moving content that induces effective parallax, or reconstruction biases on low-parallax/rotation-heavy pairs.
\textbf{C3 and C5 are strongly F-majority,} as intended: C3 has $84.2\%$/\,$46.7\%$/\,$87.5\%$ F; C5 has $68.4\%$/\,$66.7\%$/\,$94.1\%$ F (Runway/Sora/Wan).
\textbf{C6 (stress) is also F-leaning:} $56.2\%$/\,$80.0\%$/\,$64.3\%$ F.

Using the expected rule (C1/C2/C4 $\rightarrow$ H; C3/C5/C6 $\rightarrow$ F), an \emph{anomaly} occurs when the DUSt3R gate disagrees with the expectation. 
Present-only anomaly rates are:
\begin{itemize}
\item \textbf{Runway}: C1 $30.0\%$, C2 $58.8\%$, C3 $15.8\%$, C4 $11.8\%$, C5 $31.6\%$, C6 $43.8\%$; overall $31.5\%$.
\item \textbf{Sora}: C1 $25.0\%$, C2 $50.0\%$, C3 $53.3\%$, C4 $46.7\%$, C5 $33.3\%$, C6 $20.0\%$; overall $38.0\%$.
\item \textbf{Wan}: C1 $33.3\%$, C2 $62.5\%$, C3 $12.5\%$, C4 $26.3\%$, C5 $5.9\%$, C6 $35.7\%$; overall $29.0\%$.
\end{itemize}
Thus, C2 is systematically difficult (anomaly $\geq 50\%$ on Runway/Wan), while C1/C4 largely conform to expectations (lower anomalies, aside from Sora’s borderline C4).

When reconstruction failures are set to F prior to the check (our default protocol), overall anomaly rates become $33.3\%$ (Runway), $39.5\%$ (Sora), and $30.0\%$ (Wan). 
By category:
Runway $\{30.0,\,65.0,\,15.0,\,25.0,\,30.0,\,35.0\}\%$;
Sora $\{36.8,\,57.9,\,42.1,\,57.9,\,26.3,\,15.8\}\%$;
Wan $\{40.0,\,70.0,\,10.0,\,30.0,\,5.0,\,25.0\}\%$ (C1–C6). 
These increases concentrate in H-expected categories (C1/C2/C4), reflecting that failures are interpreted as large-displacement F.

\begin{figure}[h]
  \centering
  \includegraphics[width=0.9\linewidth]{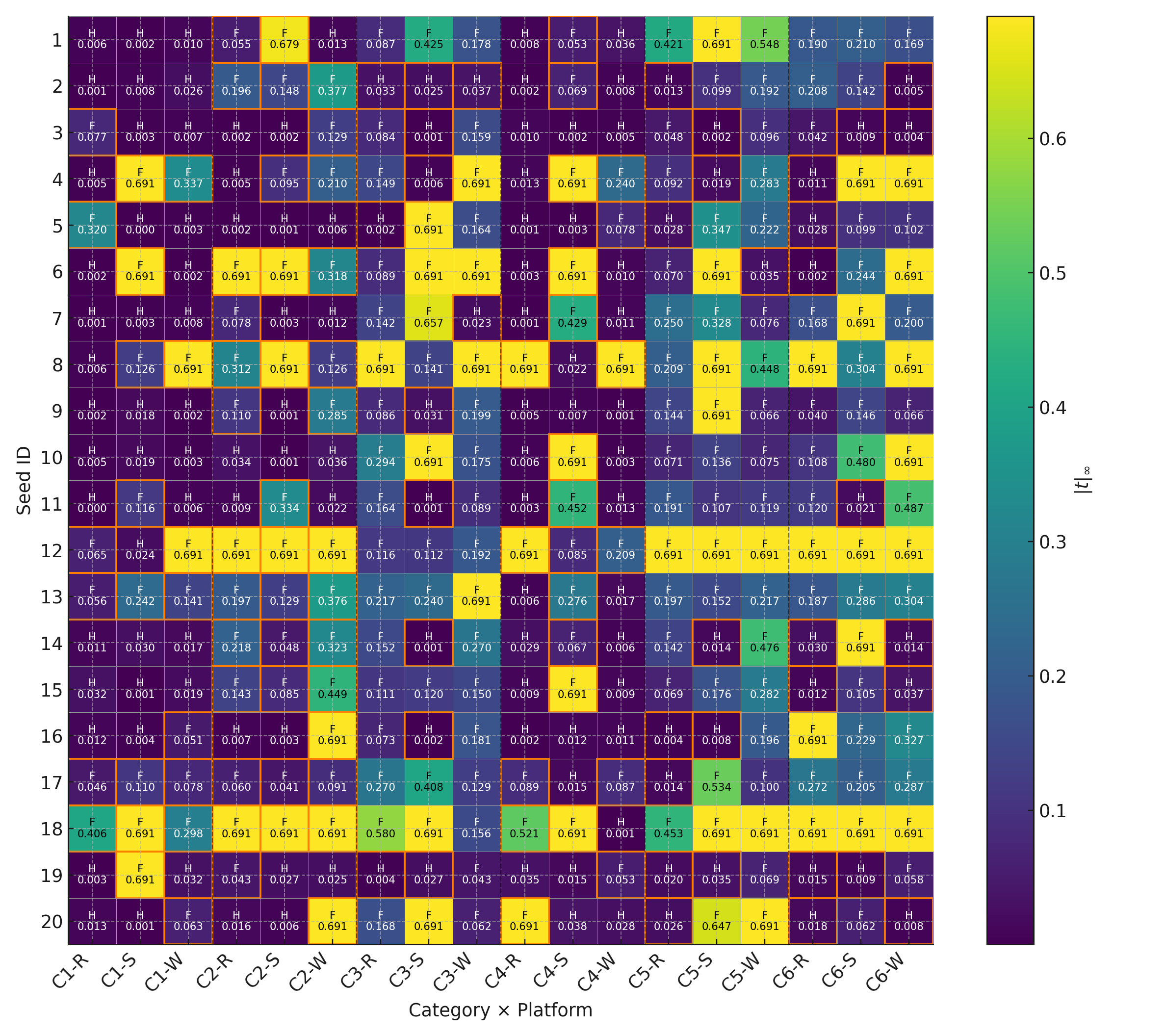}
  \caption{Combined cross-platform alignment and semantic consistency. 
  Color encodes $\lVert t\rVert_{\infty}$ from the DUSt3R first--last pose; overlaid labels show the H/F decision. 
  Orange borders mark anomalies under the expected rule (C1/C2/C4 $\rightarrow$ H; C3/C5/C6 $\rightarrow$ F), gray borders indicate matches. 
  Missing pairs are treated as F and colored with the global maximum to keep a common scale. 
  Rows follow the seed order $1$–$20$; columns are category (C1–C6) $\times$ platform (R/S/W).}
  \label{fig:geocon_combined}
\end{figure}

Fig.~\ref{fig:geocon_combined} merges cross-platform alignment and semantic consistency into a single panel. 
Each cell is color-coded by the DUSt3R-derived $\lVert t\rVert_{\infty}$ and overlaid with the H/F gate; orange borders denote anomalies under the expected rule (C1/C2/C4 $\rightarrow$ H; C3/C5/C6 $\rightarrow$ F), while gray borders indicate agreement. 
Rows list seeds in the fixed order $1$–$20$; columns enumerate category (C1–C6) $\times$ platform (R/S/W for Runway/Sora/Wan). 
Missing pairs follow our default convention (treated as F and colored with the global maximum $\lVert t\rVert_{\infty}$ to preserve a common scale). 
Under this convention, $46.7\%$ of seed–category triplets are unanimous across platforms; of those, $76.8\%$ are all-F, reflecting translation-dominant or failure-inferred cases. 
The combined view highlights three trends: (i) C3 and C5 align with translation dominance across platforms; (ii) C1 and C4 are mostly H with scattered mismatches; (iii) C2 departs from its H expectation, showing noticeable F decisions and borderline behavior near the threshold.

\subsection{Sanity checks and gate–support agreement}
\label{subsec:gate_agreement}
For each clip, we read the DUSt3R first--last pose
$\mathbf{T}=\begin{bmatrix}\mathbf{R}&\mathbf{t}\\\mathbf{0}&1\end{bmatrix}$,
compute \(t_{\infty}=\max(|t_x|,|t_y|,|t_z|)\), and apply a hysteresis gate:
\(t_{\infty}\le 0.03 \Rightarrow \mathbf{H}\),
\(t_{\infty}\ge 0.05 \Rightarrow \mathbf{F}\),
and \(0.03< t_{\infty}<0.05\) is a tolerance band (''U'').
Unless stated, Sankey plots drop U to avoid three-way clutter, whereas diagnostics report it explicitly.

    \label{fig:sankey_fh_metric}

\begin{figure}[t]
  \centering
  \begin{subfigure}[b]{0.485\linewidth}
    \centering
    \includegraphics[width=\linewidth]{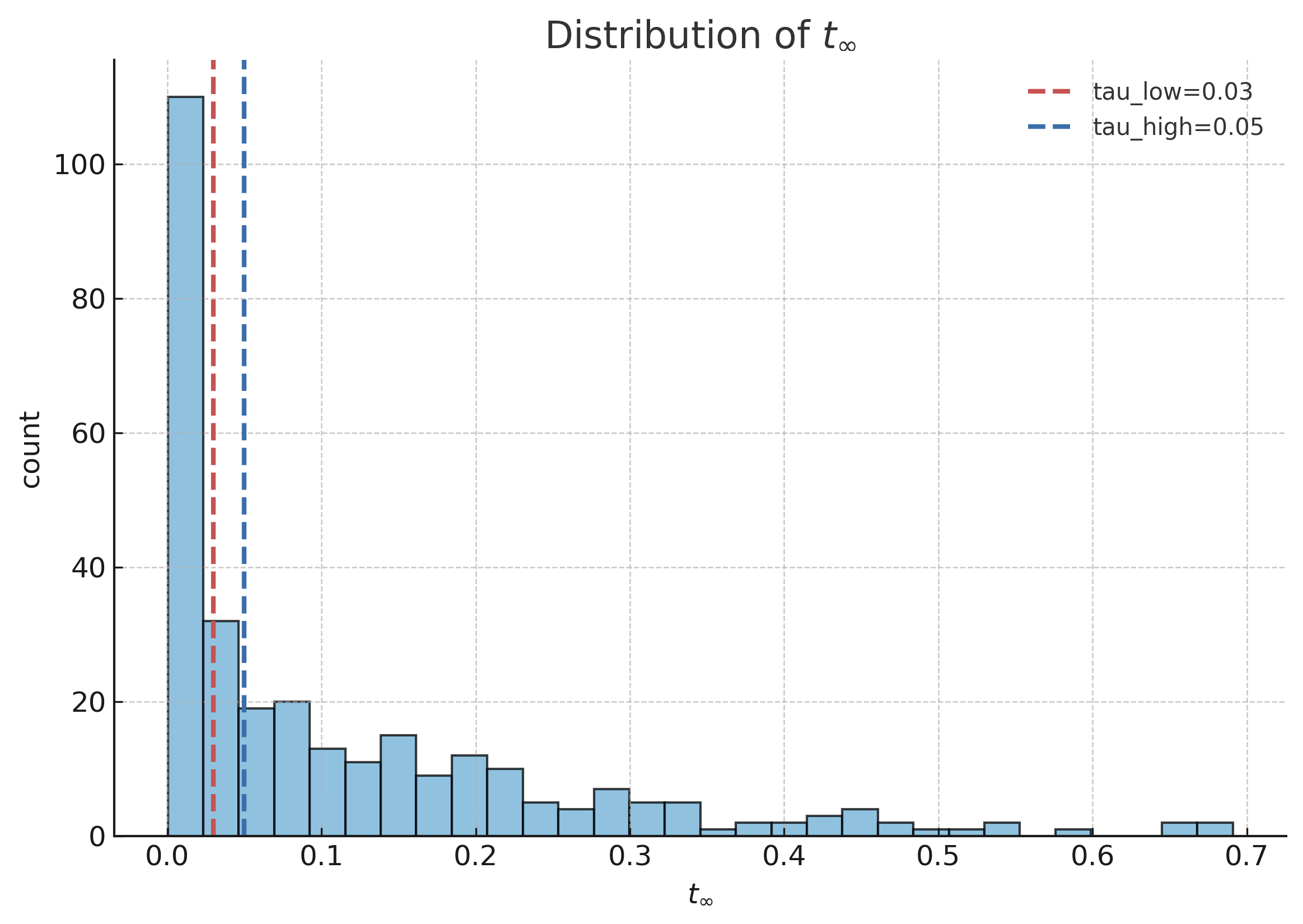}
    \caption{$t_{\infty}$ histogram.}
    \label{fig:tinf_hist}
  \end{subfigure}\hfill
  \begin{subfigure}[b]{0.485\linewidth}
    \centering
    \includegraphics[width=\linewidth]{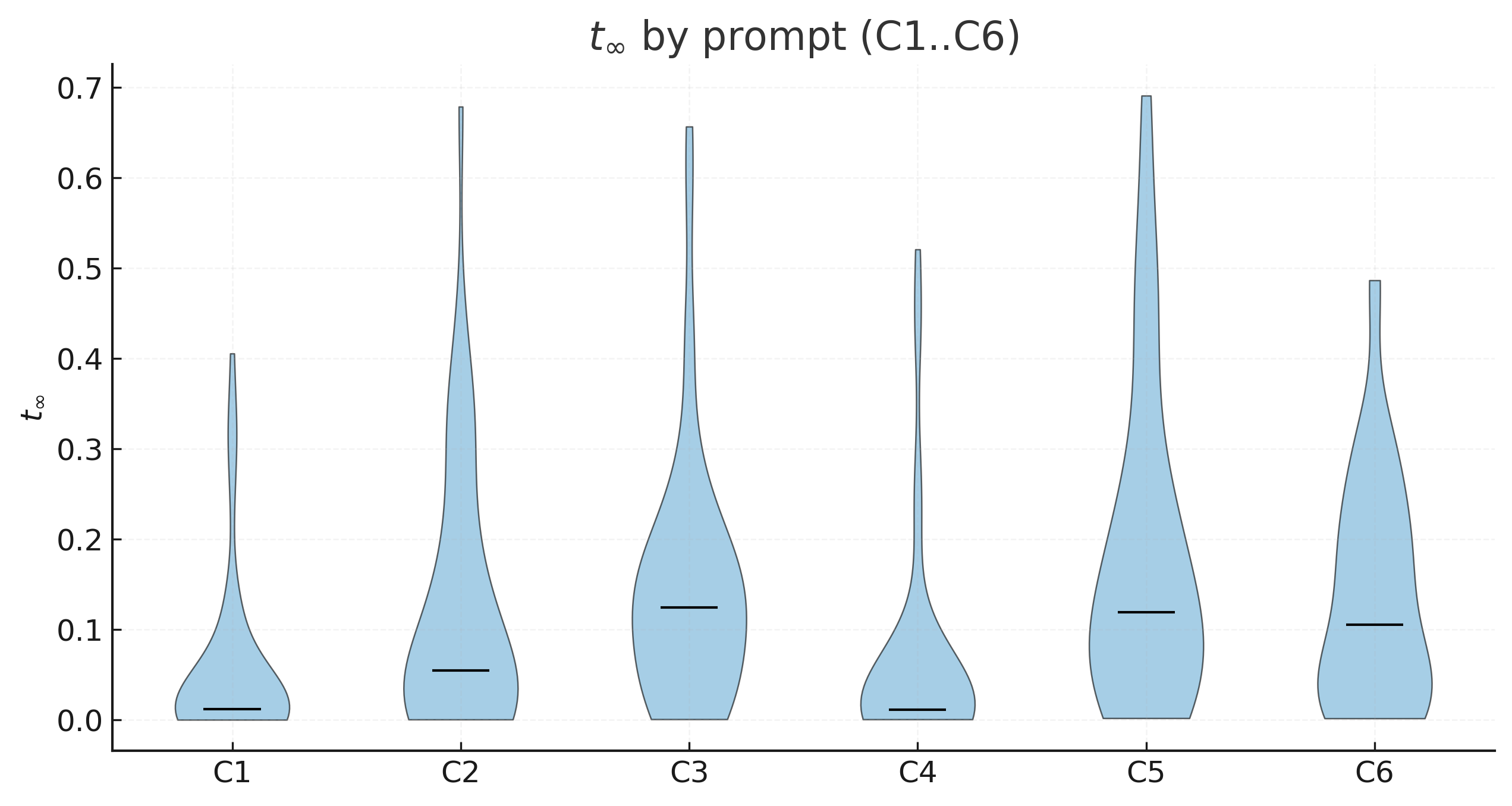}
    \caption{$t_{\infty}$ distribution by prompt (C1--C6). Violins show the full distribution with medians.}
    \label{fig:prompt_props}
  \end{subfigure}

  \vspace{0.6em}

  \begin{subfigure}[b]{0.485\linewidth}
    \centering
    \includegraphics[width=\linewidth]{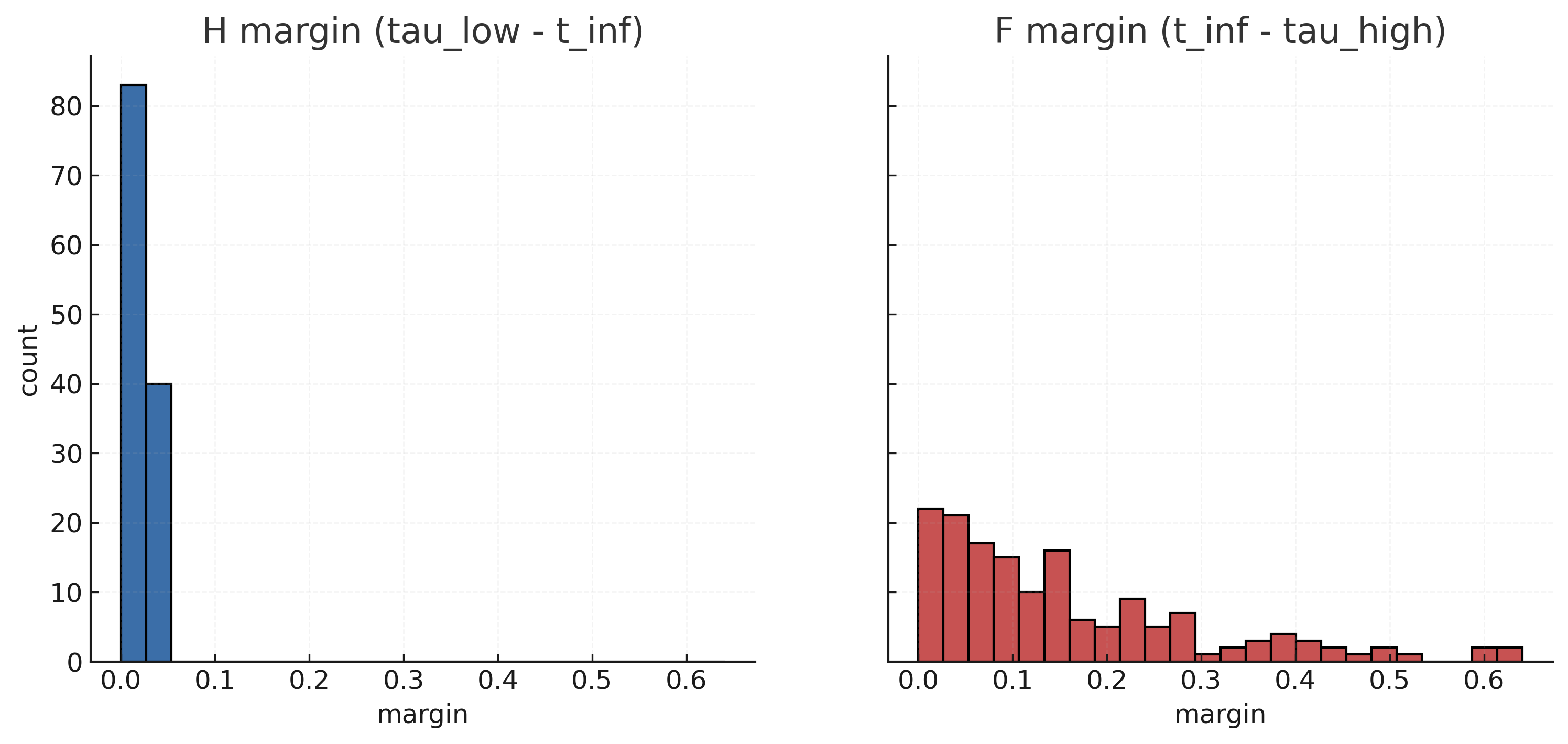}
    \caption{Gate margins (H/F panels).}
    \label{fig:gate_margin}
  \end{subfigure}\hfill
  \begin{subfigure}[b]{0.485\linewidth}
    \centering
    \includegraphics[width=\linewidth]{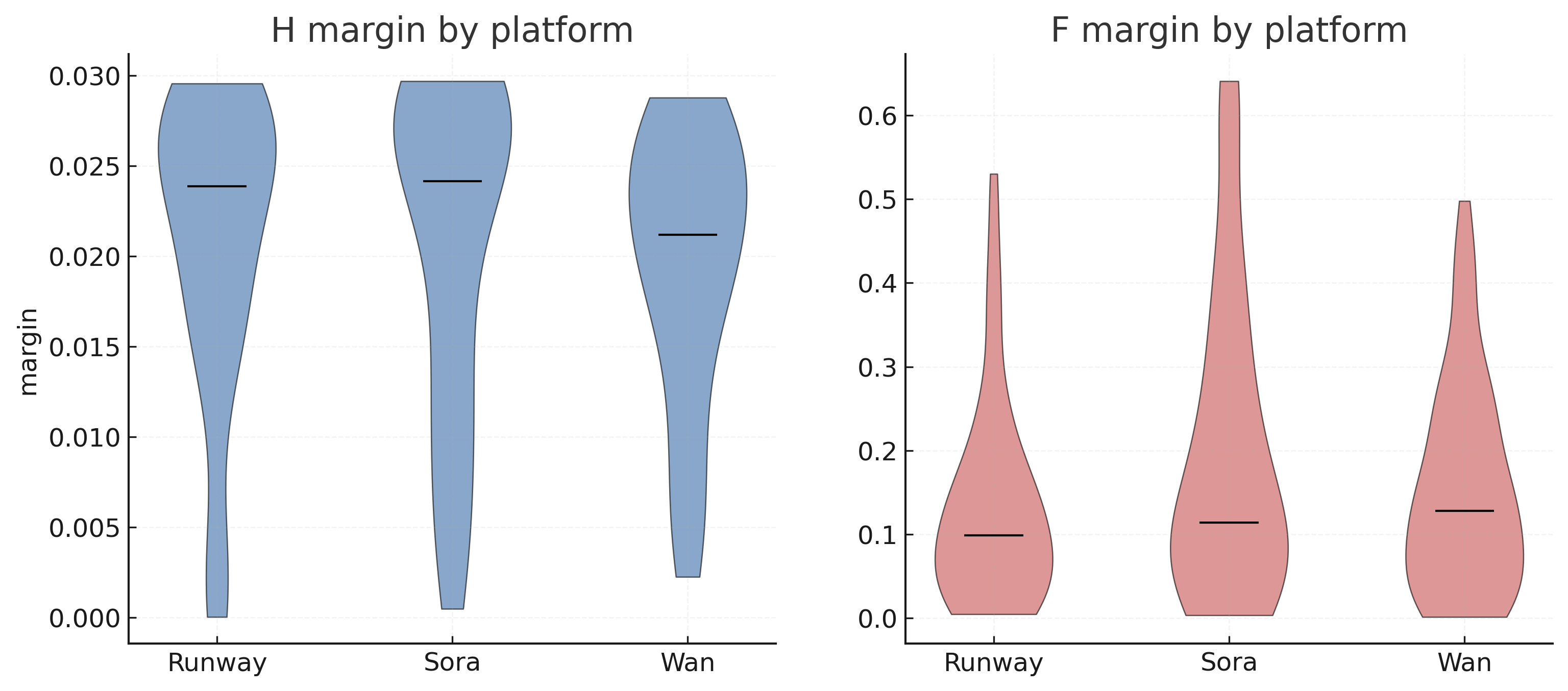}
    \caption{Gate margins by platform. Left: $m_H=0.03-t_{\infty}$ for H-gated tuples; Right: $m_F=t_{\infty}-0.05$ for F-gated tuples.}
    \label{fig:violin}
  \end{subfigure}
  \caption{Gate–support diagnostics with the hysteresis rule (\(t_{\infty}\le 0.03 \Rightarrow \mathbf{H}\), \(t_{\infty}\ge 0.05 \Rightarrow \mathbf{F}\)). 
(a) \(t_{\infty}\) histogram with thresholds, showing low occupancy of the tolerance band. 
(b) Prompt-wise \(t_{\infty}\) distributions (C1–C6; violins with medians): C1/C4 concentrate below 0.03, C3/C5 shift above 0.05, and C2 spans both regimes. 
(c) Gate margins \(m_H=0.03-t_{\infty}\) and \(m_F=t_{\infty}-0.05\) (two panels) are largely positive, indicating stable decisions away from thresholds. 
(d) Platform-conditioned margin violins (left: \(m_H\), right: \(m_F\)) reveal cross-platform differences while remaining comfortably above zero.}
  \label{fig:gate_agreement_diags}
\end{figure}

Across all available tuples (platform \(\times\) seed \(\times\) prompt) we obtain
$\mathbf{F}=156$ (52\%), $\mathbf{H}=123$ (41\%), and $\mathbf{U}=21$ (7\%).
Fig.~\ref{fig:tinf_hist} shows the global histogram of \(t_{\infty}\) with the two thresholds overlaid.
The tolerance band $[0.03,0.05]$ is sparsely occupied, supporting the defensibility of the gate.

We also visualize the full distributions with a violin plot in Fig.~\ref{fig:prompt_props}.
C1 (static scene) and C4 (static camera + moving objects) concentrate below \(0.03\) (H regime),
C3 (translation-dominant) and C5 (complex camera motion) shift mass above \(0.05\) (F regime),
while C2 (rotation-dominant) spans both sides, indicating that even small translations or scale drift can flip the decision.
\emph{C6 (geometric stress tests)} is intentionally unconstrained and shows an F-skewed but mixed distribution: $\,28/45$ F, $\,14/45$ H, $\,3/45$ U($62.2\%$ F, $31.1\%$ H, $6.7\%$ U), consistent with stress conditions that elicit both regimes.

We define gate margins as
$m_H = 0.03 - t_{\infty}$ for H-gated tuples and $m_F = t_{\infty} - 0.05$ for F-gated tuples.
Fig.~\ref{fig:gate_margin} shows the margin histograms for H and F, respectively: both concentrate away from zero, indicating decisions are not brittle.
Fig.~\ref{fig:violin} expose platform effects(left: \(m_H\), right: \(m_F\)).
Margins remain comfortably positive overall, but distributions differ across platforms, reflecting platform-specific motion priors under identical prompts.

On (seed, prompt) tuples where at least two platforms are non-U, 55/103 are unanimous (consensus) and 48/103 are split (F vs.\ H), yielding a 53.4\% consensus rate.
These differences are visible directly in the platform-conditioned margin violins in Fig.~\ref{fig:violin}.
For H-gated tuples, the distributions of $m_H=0.03-t_{\infty}$ differ by platform, and for F-gated tuples, the distributions of $m_F=t_{\infty}-0.05$ likewise separate across platforms.
Margins remain largely positive, indicating decisions are not brittle, yet their platform-specific shifts corroborate the presence of distinct motion/rendering biases under identical prompts.

\subsection{Background rigidity: distributional behavior, cross-model gaps, and effect sizes}
\label{subsec:bg_rigidity}

We evaluate background rigidity using two complementary scalars computed per tuple: 
$\mathrm{IR}$ (image-space rigidity consistency; larger is better) and $\mathrm{GE}$ (background geometry error; smaller is better).
We report both the \textbf{H} family (thresholded at $\tau_H=2$px) and the \textbf{F} family (gate near $\tau_F\approx0.10$), and we summarize per-prompt distributions (C1--C6) for each platform (Runway, Sora, Wan).
Distributions are visualized with ridgelines (per prompt) and cross-family gaps are quantified with Gardner–Altman style effect-size panels (median differences with bootstrap \mbox{95\%} CIs).

\begin{figure*}[t]
  \centering
  \begin{subfigure}[b]{0.32\linewidth}
    \centering
    \includegraphics[width=\linewidth]{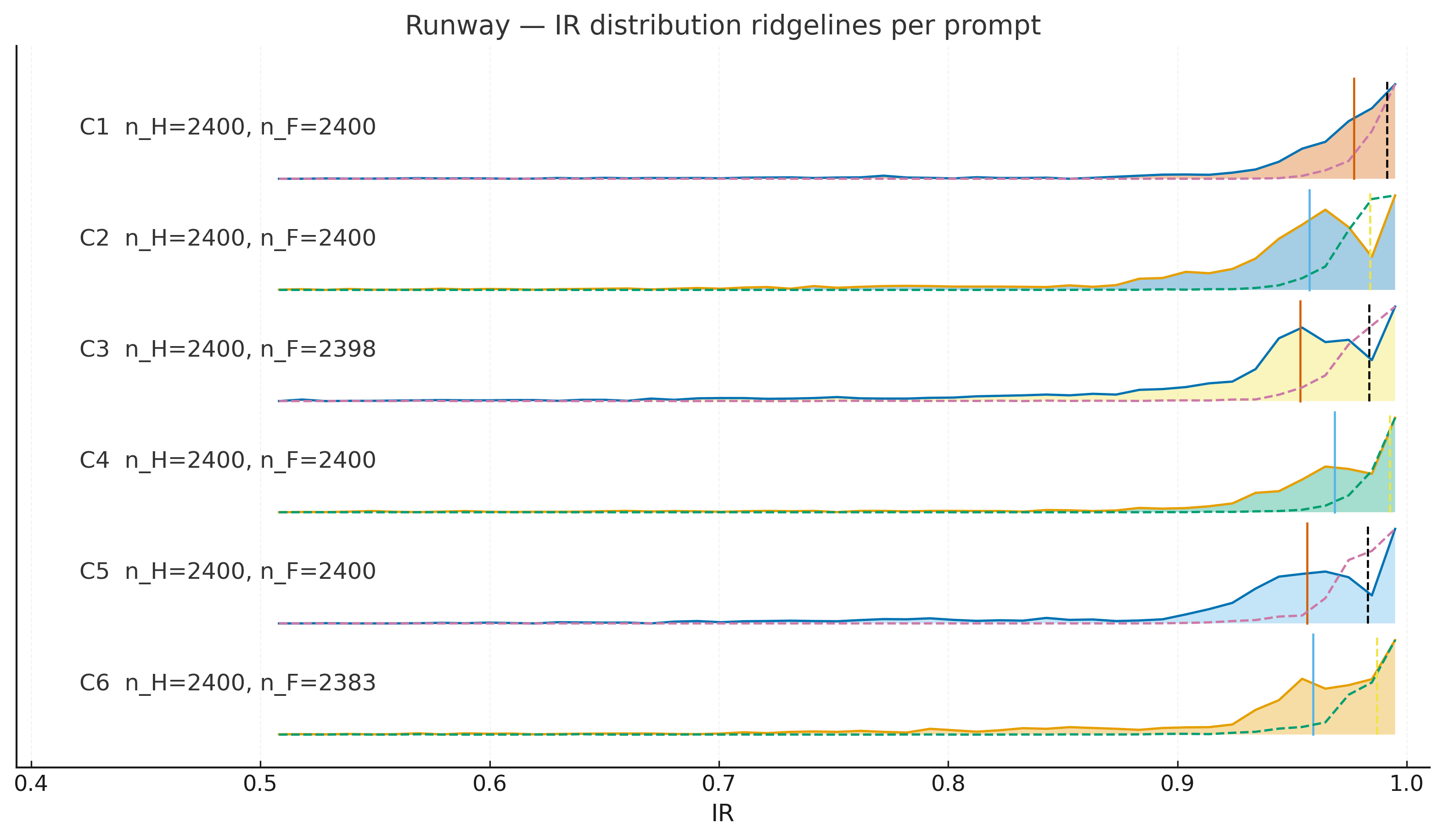}
    \caption{Runway.}
  \end{subfigure}\hfill
  \begin{subfigure}[b]{0.32\linewidth}
    \centering
    \includegraphics[width=\linewidth]{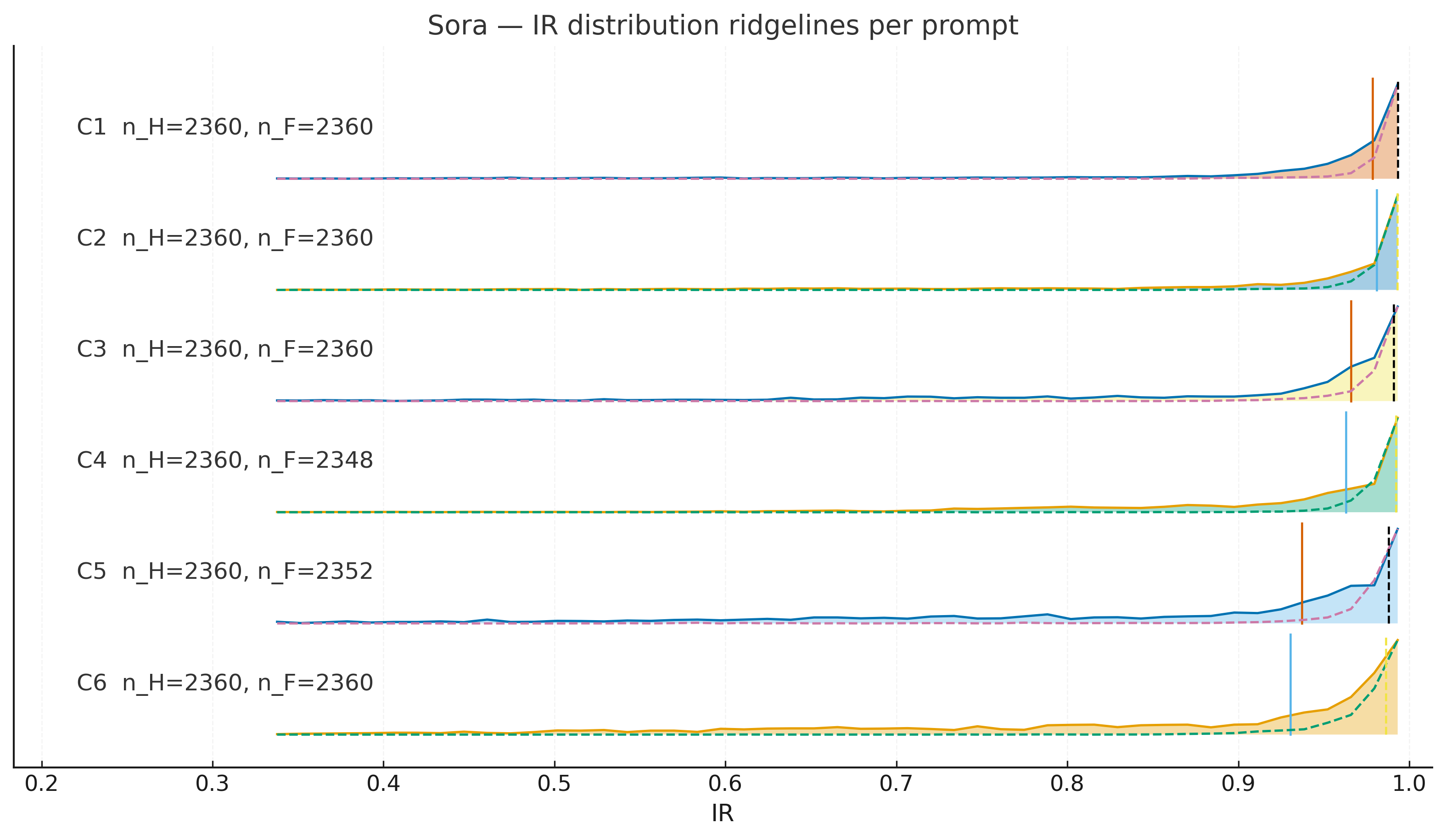}
    \caption{Sora.}
  \end{subfigure}\hfill
  \begin{subfigure}[b]{0.32\linewidth}
    \centering
    \includegraphics[width=\linewidth]{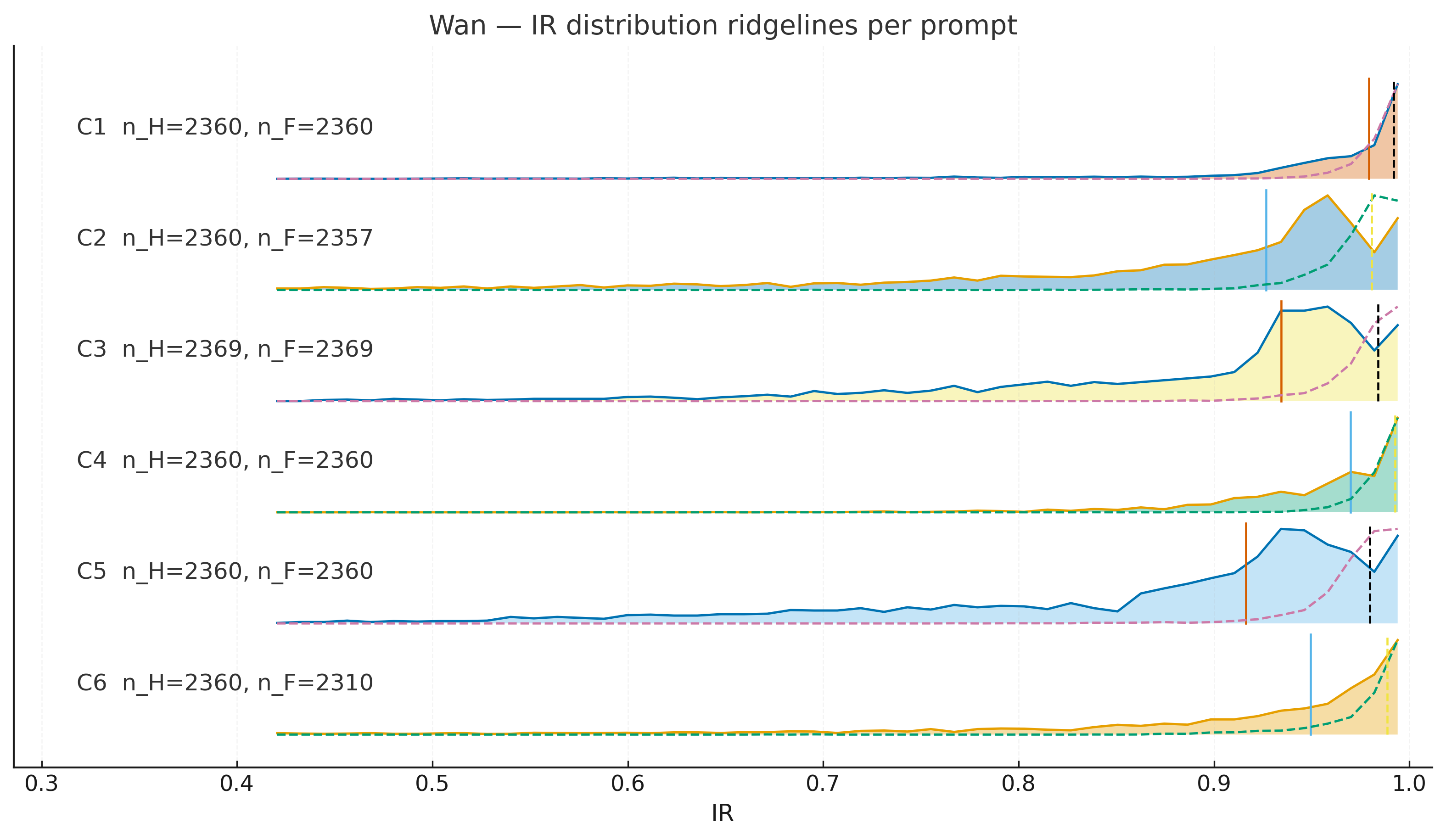}
    \caption{Wan.}
  \end{subfigure}
  \caption{IR distribution ridgelines per prompt (C1--C6).
  Each panel overlays H (solid, filled) vs.\ F (dashed) for a given platform.
  Stress prompts C4/C6 widen the upper tails across all platforms.}
  \label{fig:ridge_ir}
\end{figure*}

\begin{figure*}[t]
  \centering
  \begin{subfigure}[b]{0.32\linewidth}
    \centering
    \includegraphics[width=\linewidth]{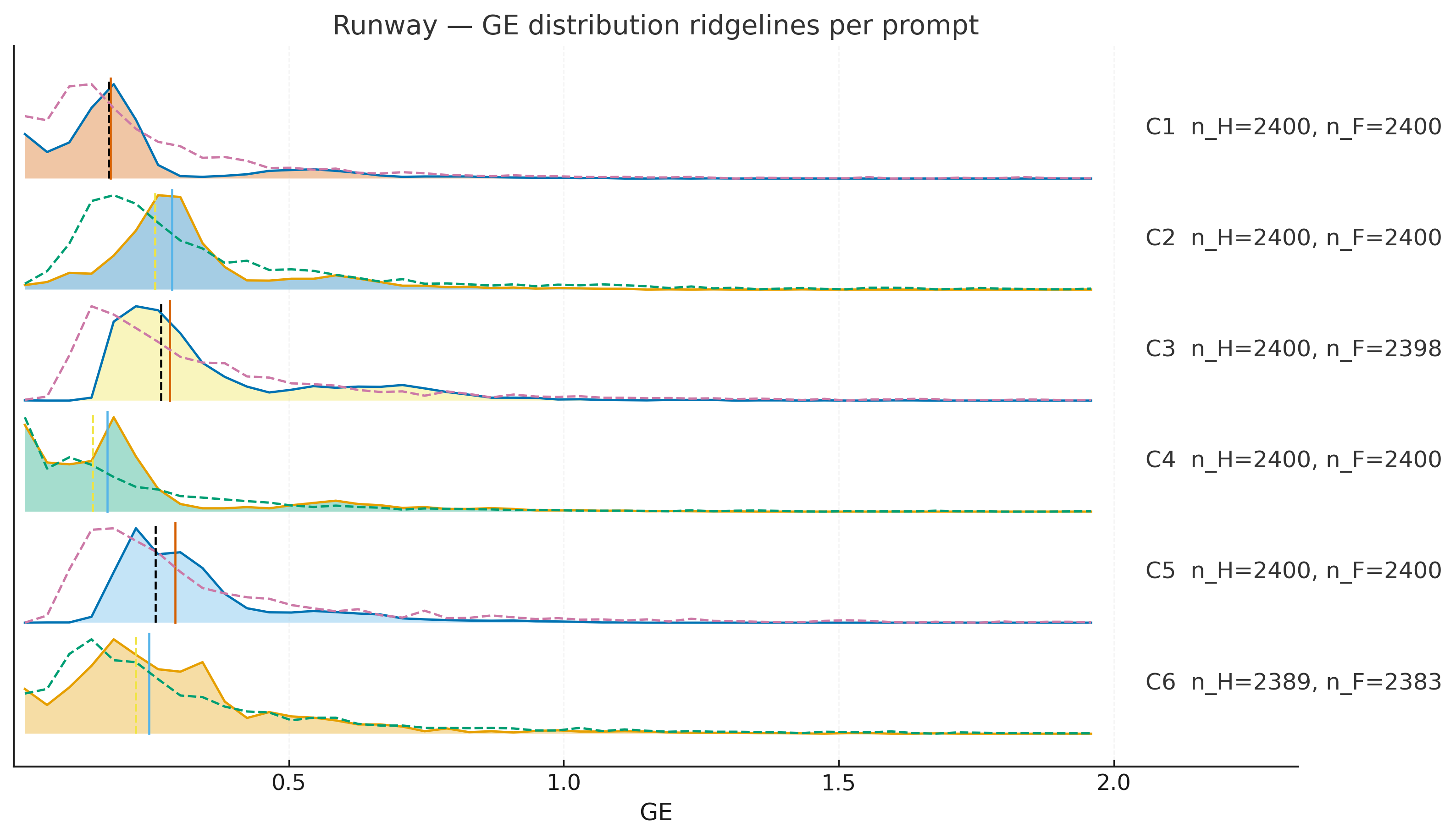}
    \caption{Runway.}
  \end{subfigure}\hfill
  \begin{subfigure}[b]{0.32\linewidth}
    \centering
    \includegraphics[width=\linewidth]{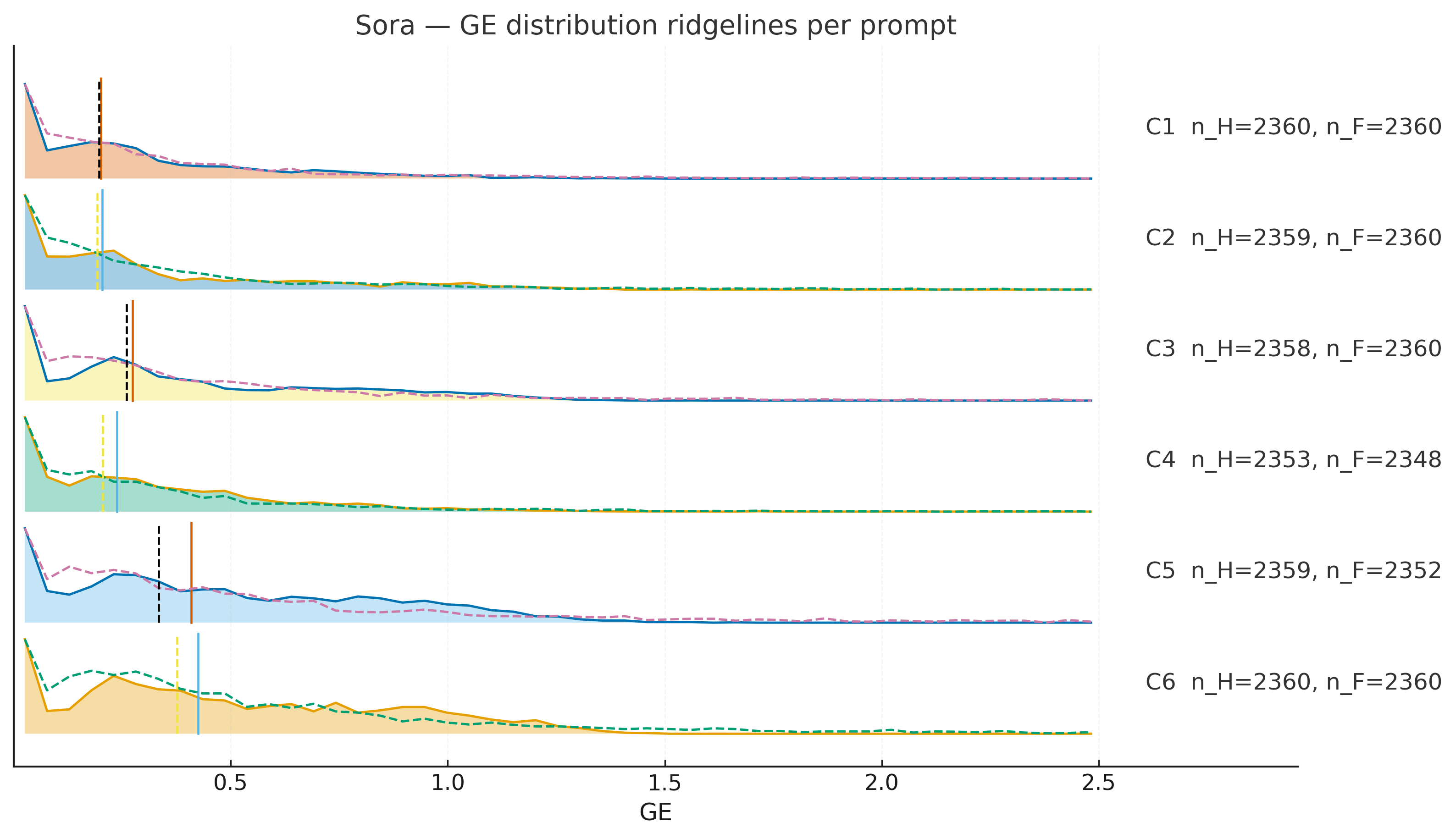}
    \caption{Sora.}
  \end{subfigure}\hfill
  \begin{subfigure}[b]{0.32\linewidth}
    \centering
    \includegraphics[width=\linewidth]{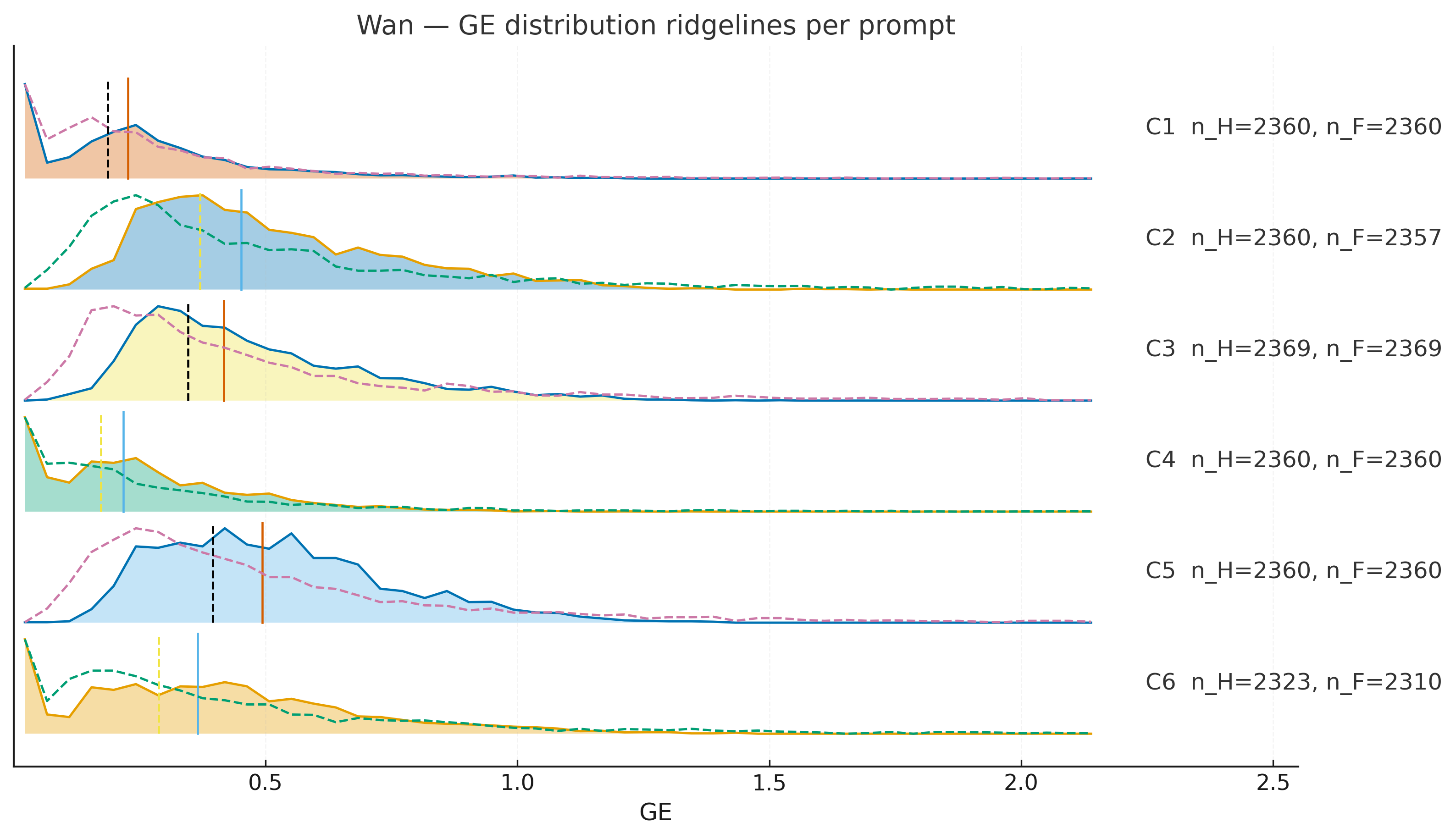}
    \caption{Wan.}
  \end{subfigure}
  \caption{GE distribution ridgelines per prompt (C1--C6).
  H shows right-shifted densities relative to F, with the largest gaps on Wan and pronounced tail mass in C4/C6.}
  \label{fig:ridge_ge}
\end{figure*}

Fig.~\ref{fig:ridge_ir} shows that \textbf{IR} concentrates in a high-rigidity regime with visibly broader right tails under the stress prompts \textbf{C4} (static camera + moving objects) and \textbf{C6} (geometric stress tests) across all three platforms.
The corresponding \textbf{GE} ridgelines in Fig.~\ref{fig:ridge_ge} reveal rightward shifts (i.e., larger errors) for \textbf{H} relative to \textbf{F}, and again heavier upper tails in \textbf{C4}/\textbf{C6}.
Together, the ridgelines indicate (i) background remains mostly rigid, but (ii) stress prompts inject a non-negligible fraction of difficult cases whose mass accumulates in the upper tails (higher $\mathrm{GE}$ / slightly lower effective $\mathrm{IR}$).

\begin{figure*}[t]
  \centering
  \includegraphics[width=\linewidth]{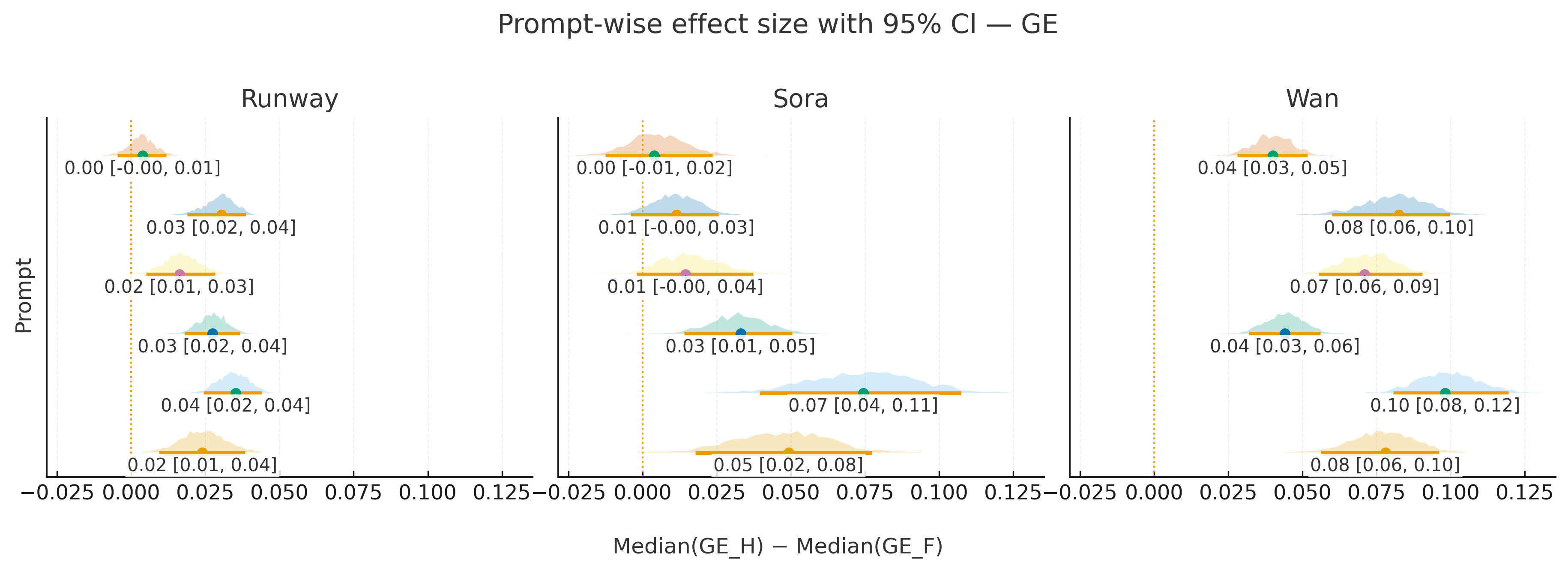}
  \caption{Prompt-wise effect size with 95\% CI --- GE.
  Median difference $\mathrm{Median}(\mathrm{GE}_H)-\mathrm{Median}(\mathrm{GE}_F)$ per prompt and platform.
  Effects are positive across the board (F$<$H), smallest on Runway, medium on Sora, largest on Wan, and amplified for C4/C6.}
  \label{fig:panel_effect_ge}
\end{figure*}

\begin{figure*}[t]
  \centering
  \includegraphics[width=\linewidth]{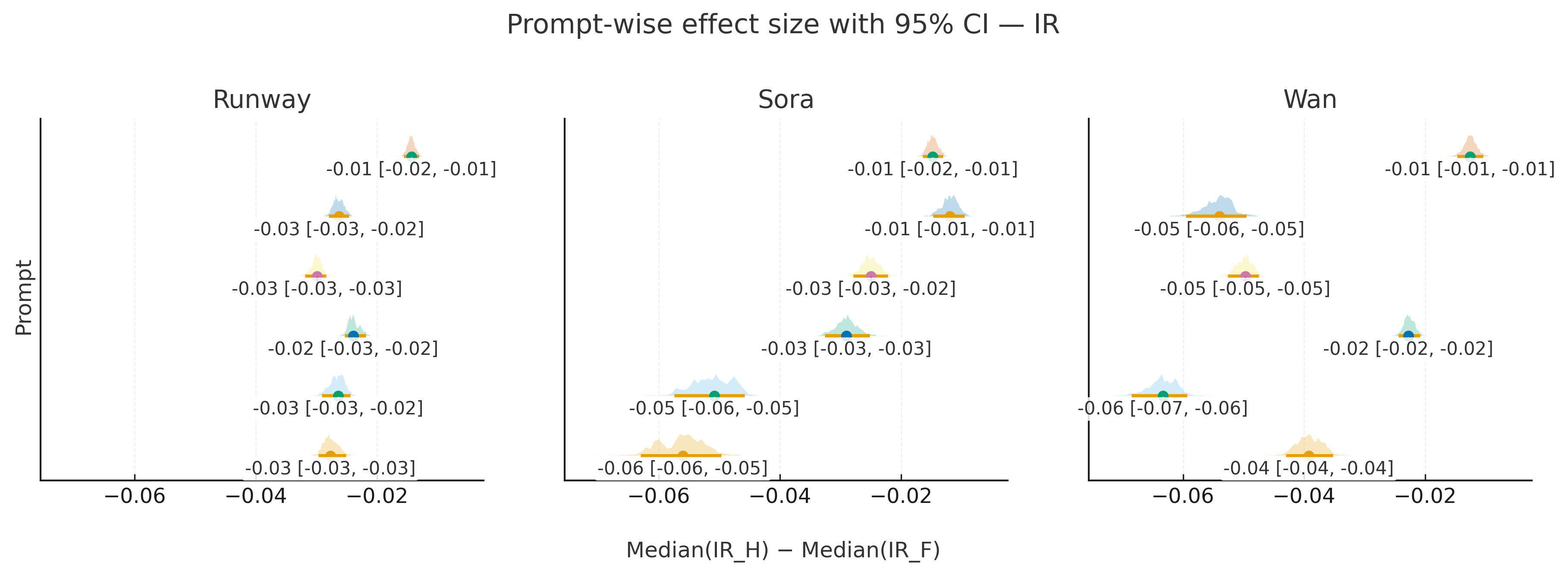}
  \caption{Prompt-wise effect size with 95\% CI --- IR.
  Median difference $\mathrm{Median}(\mathrm{IR}_H)-\mathrm{Median}(\mathrm{IR}_F)$ per prompt and platform.
  Effects are negative (F$>$H) with the strongest gaps on Sora/Wan and the mildest on Runway; the ordering by prompt mirrors C4/C6 stress.}
  \label{fig:panel_effect_ir}
\end{figure*}

To make platform-wise comparisons commensurate, we use median differences with bootstrap uncertainty.
Fig.~\ref{fig:panel_effect_ge} reports $\mathrm{Median}(\mathrm{GE}_H)-\mathrm{Median}(\mathrm{GE}_F)$: the effect is \emph{consistently positive} on all platforms, meaning \textbf{F} has lower background geometry error than \textbf{H}.
The magnitude is \textbf{small on Runway} (on the order of a few $10^{-2}$), \textbf{medium on Sora} (peaking in \mbox{C5/C6}), and \textbf{largest on Wan} (systematically the highest across C1--C6).
This ranking matches the qualitative right-shift in the \textbf{GE} ridgelines for \mbox{C4/C6} on Wan.
Fig.~\ref{fig:panel_effect_ir} reports $\mathrm{Median}(\mathrm{IR}_H)-\mathrm{Median}(\mathrm{IR}_F)$: the effect is \emph{consistently negative} (i.e., \textbf{F} yields slightly higher rigidity) with the strongest gaps on Sora/Wan and milder gaps on Runway.
Across platforms, the signs are stable, the CIs are tight, and the ordering by prompt mirrors the stress semantics: \textbf{C4} and \textbf{C6} show the most significant gaps, while \textbf{C1--C3/C5} remain smaller and often overlap.

The \textbf{F} family, which enforces epipolar-consistent background structure, systematically reduces $\mathrm{GE}$ and \emph{increases} $\mathrm{IR}$ (higher rigidity), with effects most pronounced under stress prompts (C4/C6)---see the rightward density mass in Fig.~\ref{fig:ridge_ge} and the positive GE gaps in Fig.~\ref{fig:panel_effect_ge}, together with the negative IR gaps in Fig.~\ref{fig:panel_effect_ir}.
Wan exhibits the largest \textbf{GE} gaps (H worse than F) across nearly all prompts; Sora shows medium gaps concentrated in C5/C6; Runway shows the smallest gaps.
The same ranking is echoed in \textbf{IR}, supporting a consistent story across both metrics.
(3) \emph{Stress prompts magnify tails rather than medians.}
Ridgelines show that medians stay near the high-rigidity regime, while the upper tail mass expands in C4/C6 (Fig.~\ref{fig:ridge_ir}, Fig.~\ref{fig:ridge_ge}).
Consequently, the effect-size panels capture meaningful, prompt-aligned differences even when central tendency moves only slightly.

The Gardner–Altman representation in Fig.~\ref{fig:panel_effect_ge}--\ref{fig:panel_effect_ir} (half-violin density + CI + point estimate) provides (i) robust ranking immune to outliers from stress cases; (ii) cross-platform comparability via a common horizontal scale; and (iii) visually verifiable uncertainty.
These properties are crucial because stress prompts primarily affect the tails, which can mislead mean-based summaries but are faithfully captured by medians and their sampling distribution.

Across three datasets and six prompts, the evidence is consistent and convergent:
\emph{the F family dominates the H family for background rigidity}, reducing $\mathrm{GE}$ and improving $\mathrm{IR}$, with the largest wins under C4/C6 and the strongest overall gaps on Wan, followed by Sora, then Runway.
This agrees with the qualitative mass shifts in the ridgelines and the quantitative, uncertainty-aware effect sizes.


\subsection{Dynamic coherence: flow-guided plausibility and occlusion ordering}
\label{subsec:dynamic_coherence}

We assess dynamic plausibility with the per-frame AdaDFRC-W2 score, designed to increase when occlusion ordering or flow consistency becomes implausible.
For each object track within a clip we summarize its temporal severity by the robust upper-tail statistic
\begin{equation}
    \mathrm{AdaPninety} \;\;=\;\; \mathrm{quantile}_{0.9}\left(\{\mathrm{AdaDFRC}\text{-}\mathrm{W2}_t\}_{t=1}^{T}\right),
\end{equation}
which captures the magnitude of the worst $10\%$ frames while being insensitive to isolated spikes.

\begin{figure*}[t]
  \centering
  \subcaptionbox{\label{fig:dyn_runway}}{\includegraphics[width=\linewidth]{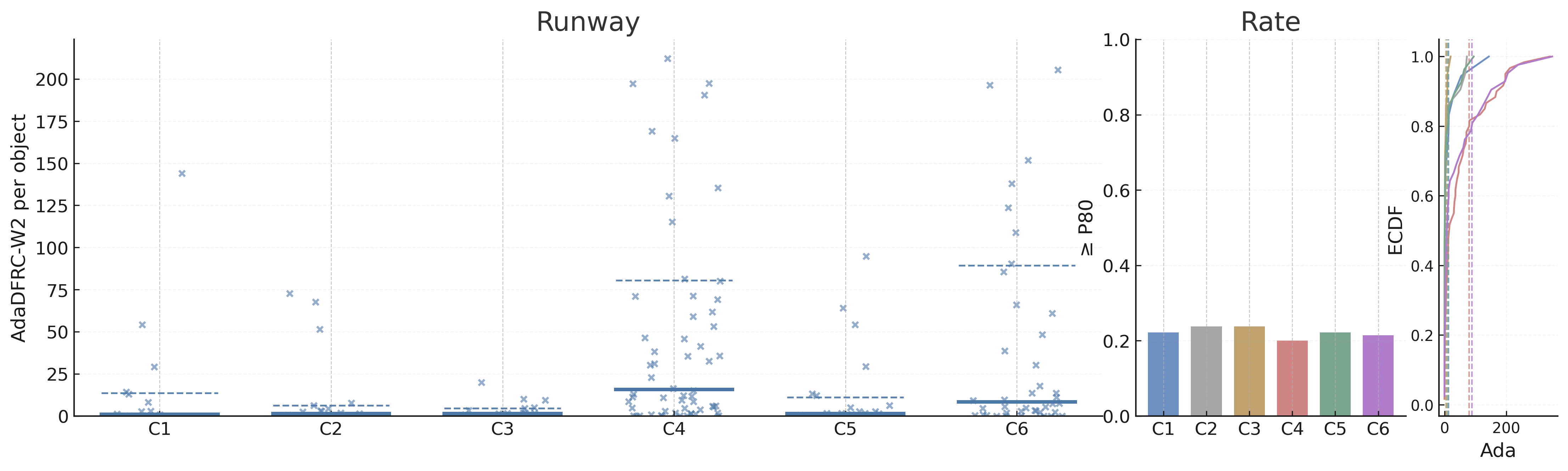}}
  \vspace{0.6em}
  \subcaptionbox{\label{fig:dyn_sora}}{\includegraphics[width=\linewidth]{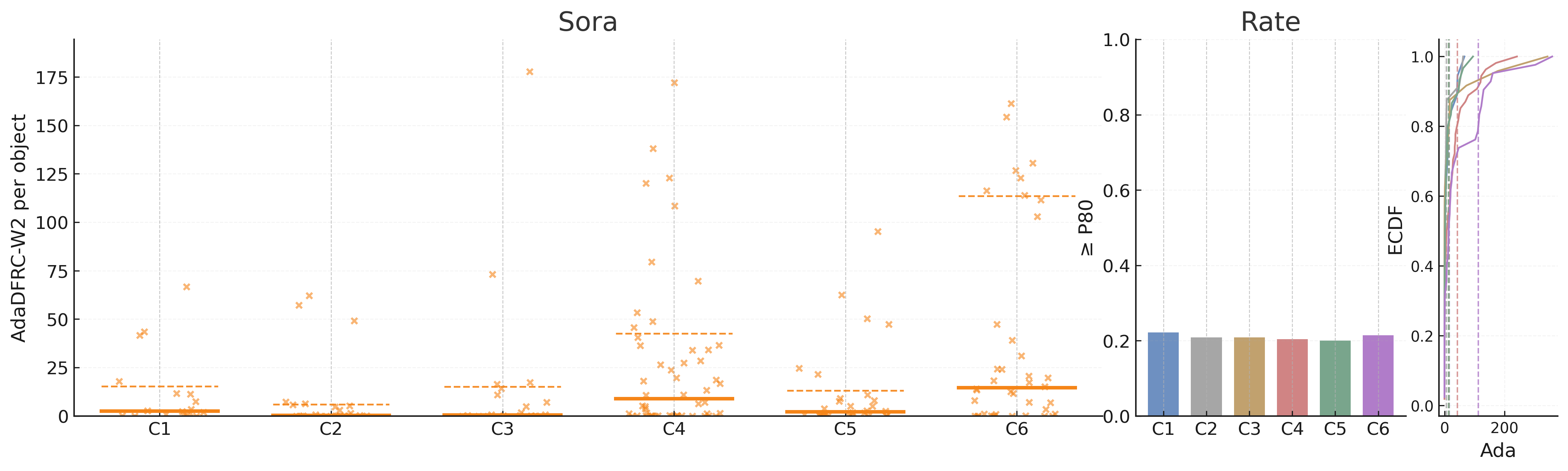}}
  \vspace{0.6em}
  \subcaptionbox{\label{fig:dyn_wan}}{\includegraphics[width=\linewidth]{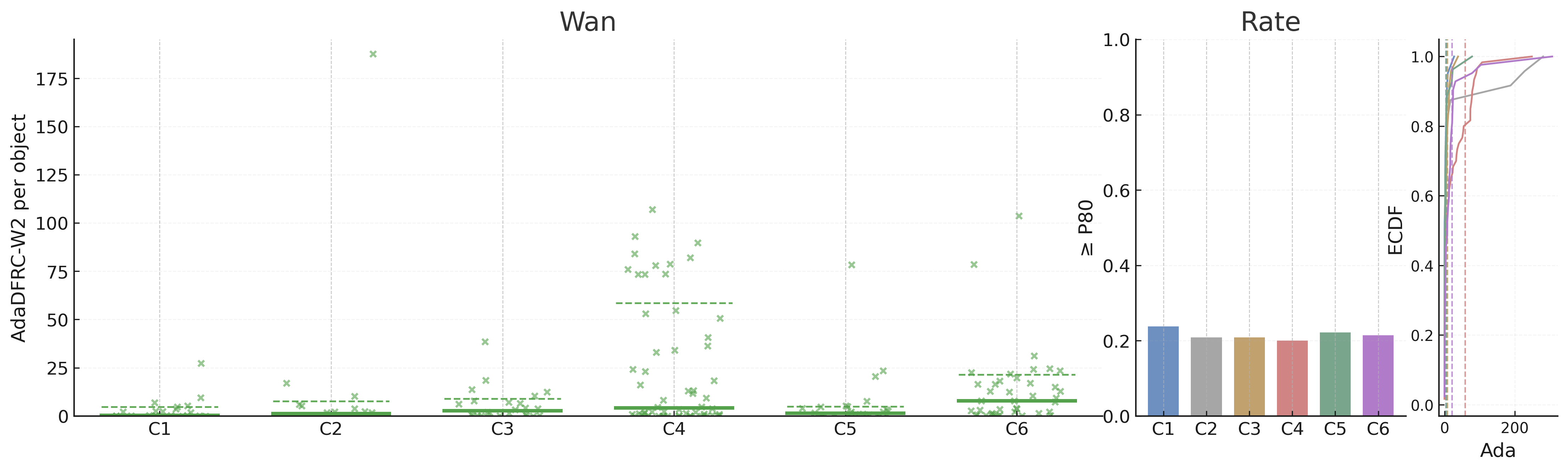}}
  \caption{Foreground rigidity via AdaDFRC-W2 across prompt families and platforms. Each row shows one platform—\textbf{(a)} Runway, \textbf{(b)} Sora, \textbf{(c)} Wan. Left: per–object AdaDFRC-W2 scores grouped by prompt family (C1–C6; lower is better). Each marker corresponds to one tracked object; points are horizontally jittered for visibility. Grey horizontal guidelines indicate the platform’s \emph{R80} robustness threshold, used to summarize the heavy-tail portion of the distribution. Right: \emph{Rate} bars report, per family, the fraction of objects whose score exceeds R80 (higher bars imply more non-rigid/unstable motion). The ECDF inset shows the empirical cumulative distribution of AdaDFRC-W2 for that platform, with the vertical dashed line marking R80.}

  \label{fig:dyn_platform}
\end{figure*}

\begin{figure*}[t]
  \centering
  \subcaptionbox{Runway\label{fig:sparks_runway}}{\includegraphics[width=0.49\linewidth]{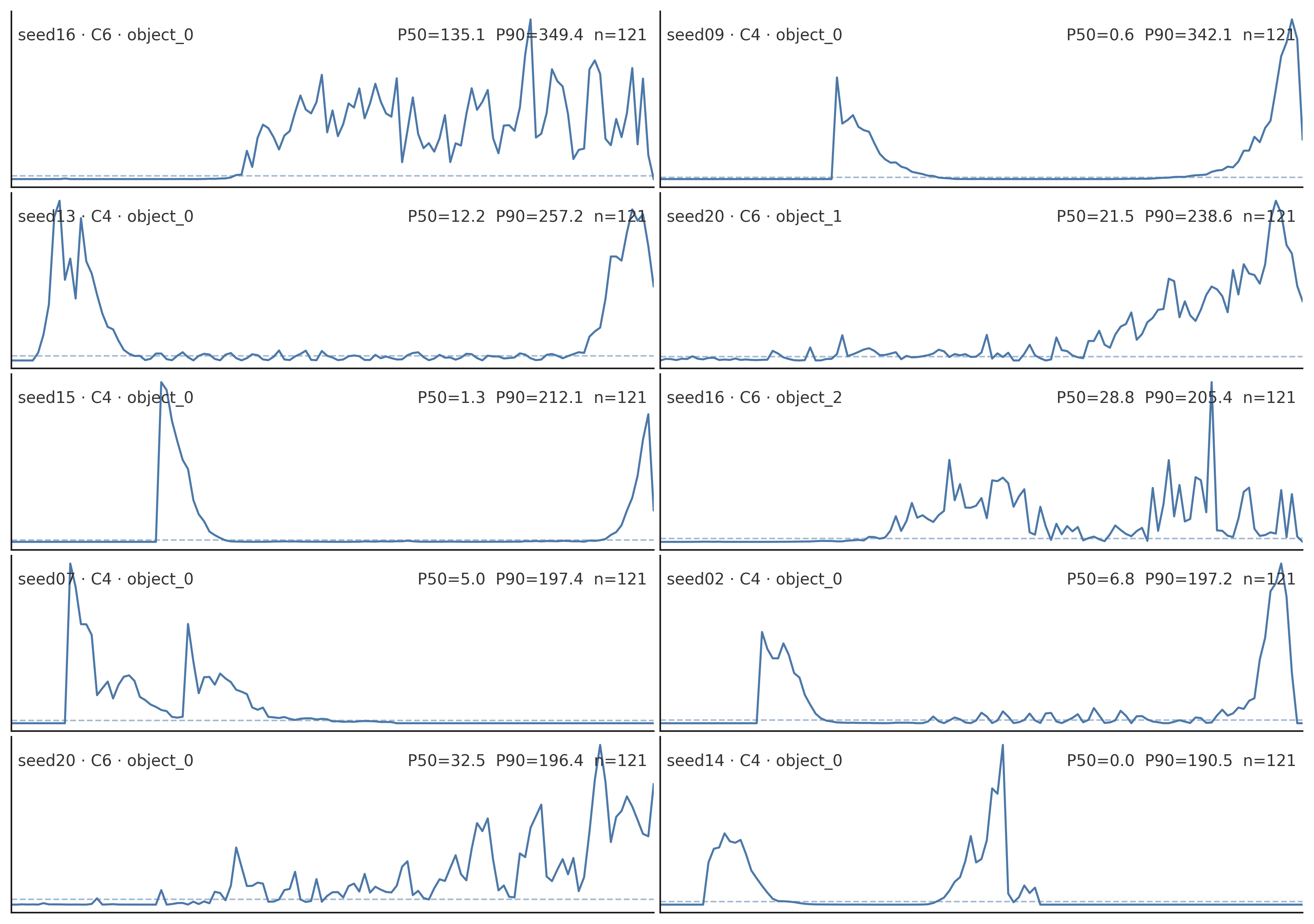}}
  \subcaptionbox{Sora\label{fig:sparks_sora}}{\includegraphics[width=0.49\linewidth]{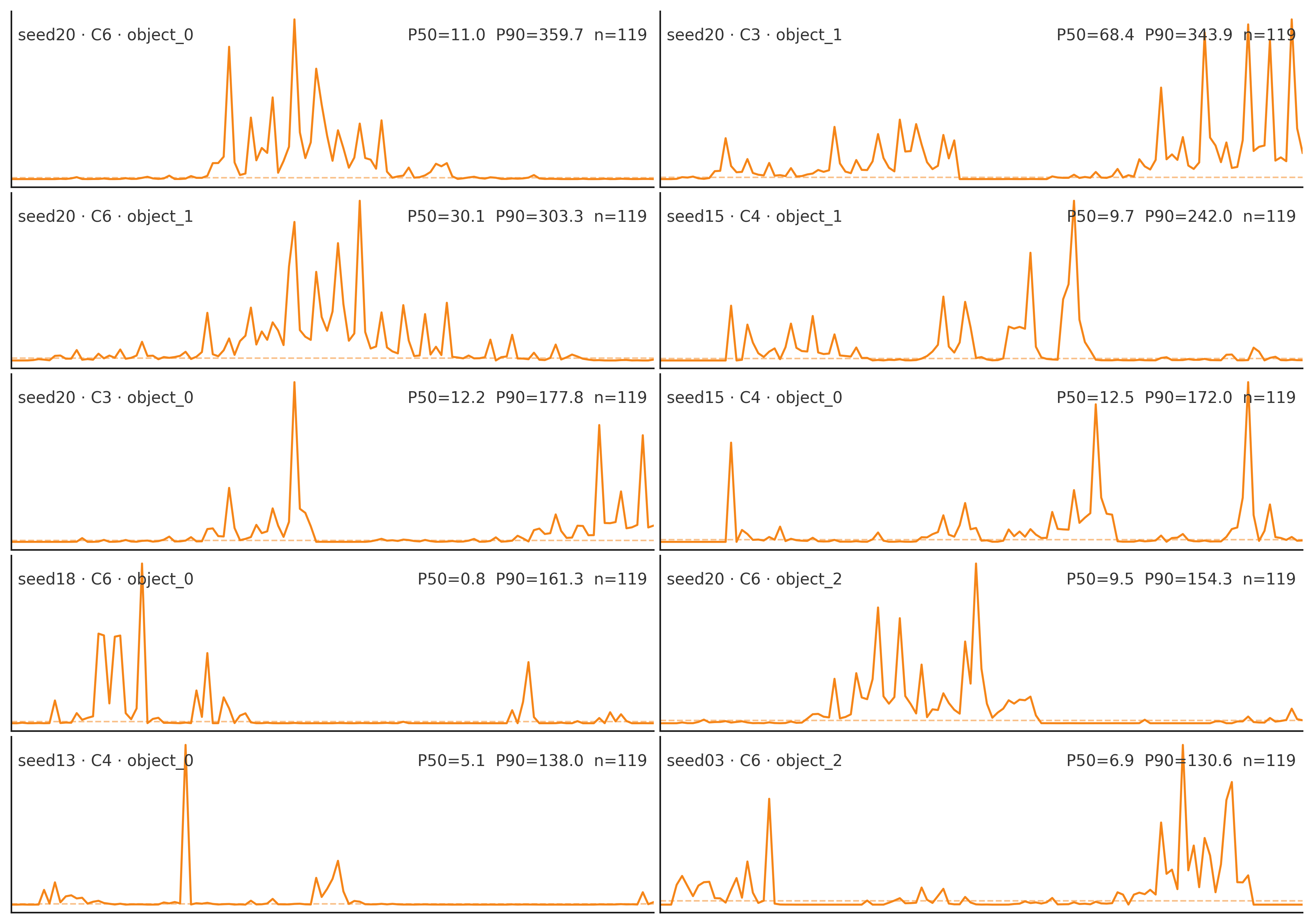}}\\
  \subcaptionbox{Wan\label{fig:sparks_wan}}{\includegraphics[width=0.49\linewidth]{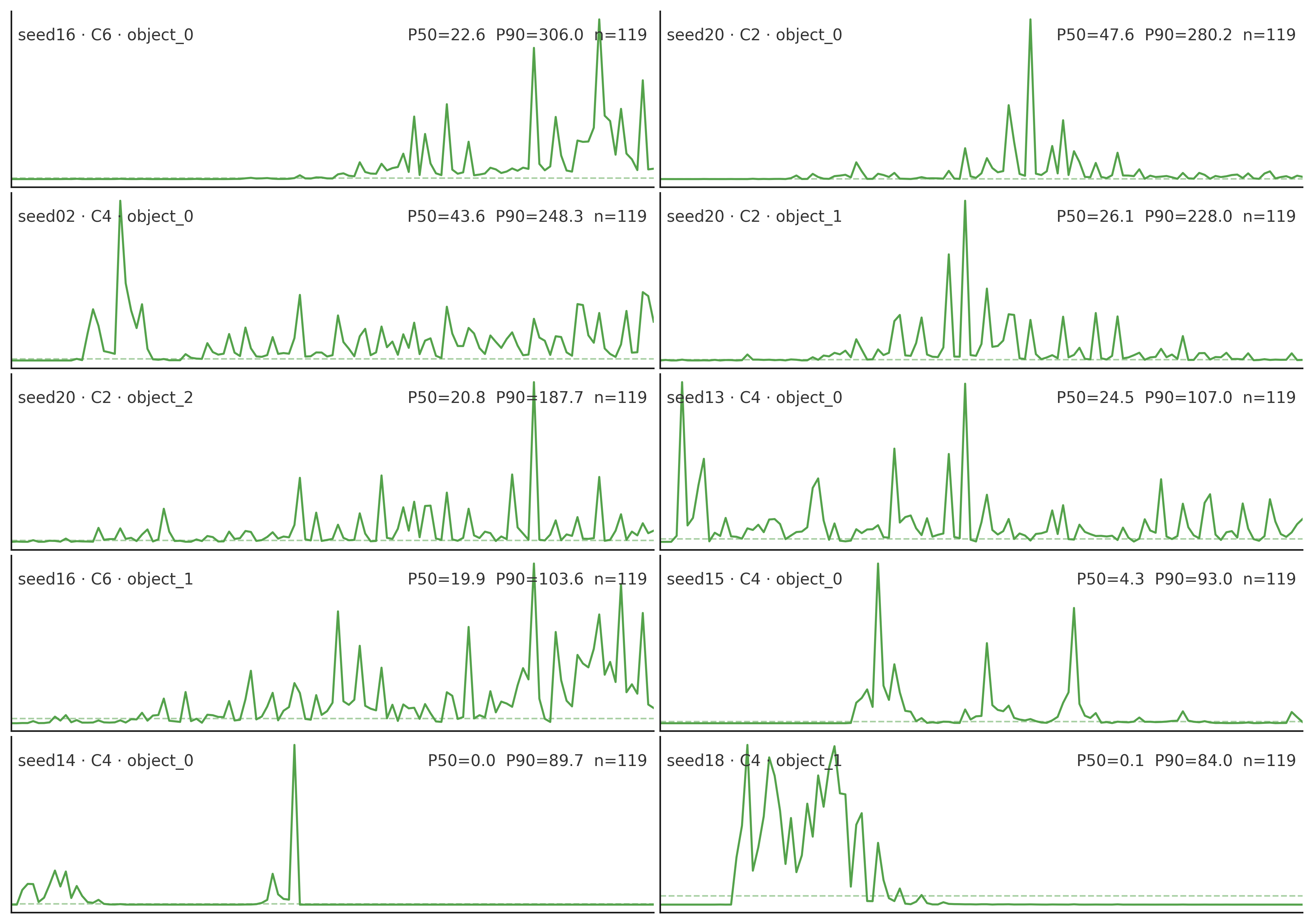}}
  \caption{Top-$N$ hardest objects per model. Each sparkline shows per-frame AdaDFRC-W2 for one object; the page-level x/y labels indicate frame index (temporal order) and AdaDFRC-W2, respectively. Dashed lines mark the model-wise median of per-prompt P80. Right-side annotations report P50, P90, and the number of frames (n).}
\end{figure*}

We rely on four complementary views.
(i) \emph{Beeswarm panels} group every object by prompt (C1--C6) and model, and overlay the per-prompt, per-model P80 thresholds (dashed ticks) and median markers (short bars), see Figs.~\ref{fig:dyn_runway}--\ref{fig:dyn_wan}.
(ii) \emph{Empirical Cumulative Distribution Function (ECDF) panels} show full distributions with the same P80 thresholds, exposing tail heaviness (right-shifts).
(iii) \emph{Anomaly-rate bars} report the fraction of objects above the \emph{per-prompt} P80; these rates are primarily descriptive (they are near $20\%$ by construction and not used for cross-prompt ranking).
(iv) \emph{Top-$N$ sparkline sheets} plot the per-frame trajectories for the hardest objects in each model with summary statistics (P50/P90/$n$), revealing whether failures are bursty or persistent (Figs.~\ref{fig:sparks_runway}--\ref{fig:sparks_wan}).

Dynamic coherence is only defined when detectable objects exist.
Across our data, the object counts per model are comparable, providing sufficient support for model- and prompt-conditioned analyses.
Prompts C4 (static camera + moving objects) and C6 (geometric stress tests) contain the richest dynamic evidence, consistent with our protocol design.

We report consistent, model-agnostic evidence that dynamic coherence is most fragile under prompts C4 (static camera + moving objects) and C6 (geometric stress tests). Across all three models, the beeswarm panels reveal visibly taller stacks for C4/C6 while the medians remain comparatively stable, and the ECDFs shift to the right with fatter upper tails. This pattern indicates that failures concentrate in the worst part of the distribution rather than uniformly raising central tendency—exactly the behavior one would expect when motion- or geometry-induced occlusion ordering becomes implausible.

Although all models exhibit heavier tails for C4/C6, the magnitude of the extremes differs across models. The largest $\mathrm{AdaPninety}$ objects—and thus the most severe dynamic inconsistencies—are most pronounced on Runway and Sora, with Wan displaying comparatively lower maxima. This cross-model ordering is independently corroborated by the Top-$N$ sparkline sheets: the largest-amplitude trajectories cluster in (Runway,~C6) and (Sora,~C6/C4), whereas Wan’s top traces remain lower in amplitude, albeit still clearly abnormal.

The temporal shapes of these failures are mechanistically distinct. The Top-$N$ trajectories expose (i) brief burst spikes aligned with object enter/exit events or edge flicker; (ii) quasi-periodic rise–fall patterns consistent with recurrent depth/flow flips; and (iii) long segments of persistent elevation indicative of sustained layer mis-ordering or non-rigid flow leakage. Notably, these dynamic signatures often arise without commensurate increases in background photogrammetric residuals, underscoring that they are genuinely dynamic (flow/ordering) failures rather than artifacts of the static fit.

We also clarify how prevalence is interpreted in our diagnostics. Because thresholds are defined \emph{per prompt and per model} at the P80 level, the within-prompt anomaly fractions are expected to be near $20\%$ by construction and are therefore not used to rank prompts. Instead, our conclusions rest on \emph{severity} (how far the upper tail extends) and \emph{tail shape} (ECDF right-shift), which together yield a clear and stable ordering in upper-tail mass: C4 $>$ C6 $>$ (C1, C2, C3, C5). Joint checks with gate–support agreement (Sec.~\ref{subsec:gate_agreement}) further show a non-trivial subset of clips in which dynamic anomalies coexist with strong background fits (''dynamic-anomaly \& background-good''), amounting to roughly a tenth of cases. This complementarity motivates including an explicit, object-aware dynamic score such as AdaDFRC-W2 alongside background rigidity fits in any holistic geo-consistency audit.

Two robustness choices make the conclusions stable:
(i) summarizing per-object with $\mathrm{AdaPninety}$ instead of maxima prevents overreaction to isolated spikes, yet keeps genuine tail mass visible;
(ii) auditing with model- and prompt-specific P80 thresholds prevents cross-model scale confounding while preserving comparability of \emph{shapes} and \emph{extremes}.
Empirically, Top-$N$ membership is stable under small changes in $N$ (8--12) and under $\pm 2$ percentile changes in the per-prompt thresholds.

Dynamic coherence is the dominant failure axis under motion- and geometry-stress prompts.
It manifests as heavy-tailed $\mathrm{AdaPninety}$ distributions with clear bursty or persistent temporal signatures, and it is \emph{not} predictable from background photogrammetric residuals alone.
Therefore, any holistic geo-consistency audit must include an object-aware, flow-guided dynamic term like AdaDFRC-W2 in addition to background rigidity fits.

\subsection{Quantitative Results}
We present the scene-balanced average scores for each generative model across all categories of GeoCon-Bench in Table~\ref{tab:results}. Higher IR is better, while lower GE and AdaDFRC-W2 are better.
We report scene-balanced averages over all seeds and categories of GeoCon-Bench. For each seed--category pair we first average frame-wise scores within a clip and then average equally across scenes. Background scores use H/F-based metrics (IR and GE) on the static layer, while foreground scores report AdaDFRC-W2 on moving objects. Higher IR is better; lower GE and AdaDFRC-W2 are better.

\begin{table}[h]
\centering
\caption{Scene-balanced average results on GeoCon-Bench}
\label{tab:results}
\begin{tabular}{lccc}
\toprule
& \multicolumn{2}{c}{\textbf{Background}} & \textbf{Foreground} \\
\cmidrule(lr){2-3} \cmidrule(lr){4-4}
\textbf{Model} & \textbf{IR (\%)} $\uparrow$ & \textbf{GE (norm)} $\downarrow$ & \textbf{AdaDFRC-W2} $\downarrow$ \\
\midrule
Runway & \textbf{97.0} & \textbf{0.274} & 18.54 \\
Sora   & 96.6 & 0.398 & 21.79 \\
Wan    & 96.2 & 0.394 & \textbf{12.76} \\
\bottomrule
\end{tabular}
\end{table}

Background geometry is consistently strong across models (IR $\approx$ 96--97\%), with sub-pixel GE magnitudes on average.
Foreground coherence (AdaDFRC-W2) varies more: Wan attains the lowest deformation on average (12.76), Runway is moderate (18.54), and Sora is the highest (21.79). 
A category-wise breakdown reveals model-specific weaknesses:
Runway exhibits elevated foreground deformation in C4 (Static camera + moving object, 31.69) and C6 (Stress, 23.91);
Sora peaks in C6 (36.84) and is also high in C3/C4 (22.83/22.36);
Wan is most challenged in C2 (Rotation-dominant, 24.42) and C4 (18.94), while remaining low in C1/C5 (2.30/4.86). 
These patterns are consistent with the combined alignment/consistency map in Fig.~\ref{fig:geocon_combined}, where translation-dominant or failure-inferred cells tend to coincide with larger foreground deformation.

Across scenes, the 95\% confidence intervals (CI) for background IR are 
Runway $[96.35,\,97.55]$, Sora $[95.91,\,97.29]$, Wan $[95.41,\,97.06]$ (in \%);\ 
for GE (px): Runway $[1.90\!\times\!10^{-4},\,2.48\!\times\!10^{-4}]$, Sora $[2.70\!\times\!10^{-4},\,3.65\!\times\!10^{-4}]$, Wan $[2.78\!\times\!10^{-4},\,3.52\!\times\!10^{-4}]$; 
for AdaDFRC-W2: Runway $[12.32,\,24.76]$ (63 scenes), Sora $[12.74,\,30.84]$ (64), Wan $[6.57,\,18.95]$ (66).

The results indicate clear differences in the geometric consistency of the evaluated models. A more detailed breakdown by scenario reveals specific strengths and weaknesses. For instance, most models perform well on pure rotation scenarios, achieving high IR with the homography model, but struggle with pure translation, where maintaining epipolar constraints for parallax is more challenging. The AdaDFRC-W2 scores are particularly revealing in categories with moving objects, quantifying the degree of unnatural deformation.

\subsection{Analysis: Decoupling Geometric and Semantic Quality}
We investigate whether semantic alignment, measured by CLIP, predicts geometric coherence of the generated videos. 
For each scene (seed--category) and platform, we robustly fit a single background motion model on background-only correspondences, using either a homography $\mathbf{H}$ (planar/rotation regime) or a fundamental matrix $\mathbf{F}$ (translation/parallax regime). 
Models are estimated in normalized coordinates for numerical stability, and residuals are \emph{denormalized to pixels} before scoring on the content mask (letterboxing excluded). 
Under either family $g\in\{\mathbf{H},\mathbf{F}\}$ we report two model-agnostic indicators: 
\emph{Inlier Ratio (IR)}—the fraction of correspondences passing the family-specific inlier test—and 
\emph{Geometric Error (GE)}—the median \emph{pixel} residual among those inliers. 
Hence IR answers whether a single global model sufficiently explains the background motion, while GE quantifies how tightly the consistent matches align, both defined for either $\mathbf{H}$ or $\mathbf{F}$.

We analyze the coupling of CLIP with geometry from:
(i) an \textbf{auto-selected} view, where each scene adopts the better-fitting family, 
$\mathrm{IR}_{\text{best}}=\max(\mathrm{IR}_H,\mathrm{IR}_F)$ and 
$\mathrm{GE}_{\text{best}}=\min(\mathrm{GE}_H,\mathrm{GE}_F)$; and
(ii) a \textbf{per-family} view that keeps $\mathbf{H}$ and $\mathbf{F}$ separate to avoid bias toward either residual family.

\begin{figure*}[t]
  \centering
  \begin{subfigure}[t]{.49\linewidth}
    \centering
    \includegraphics[width=\linewidth]{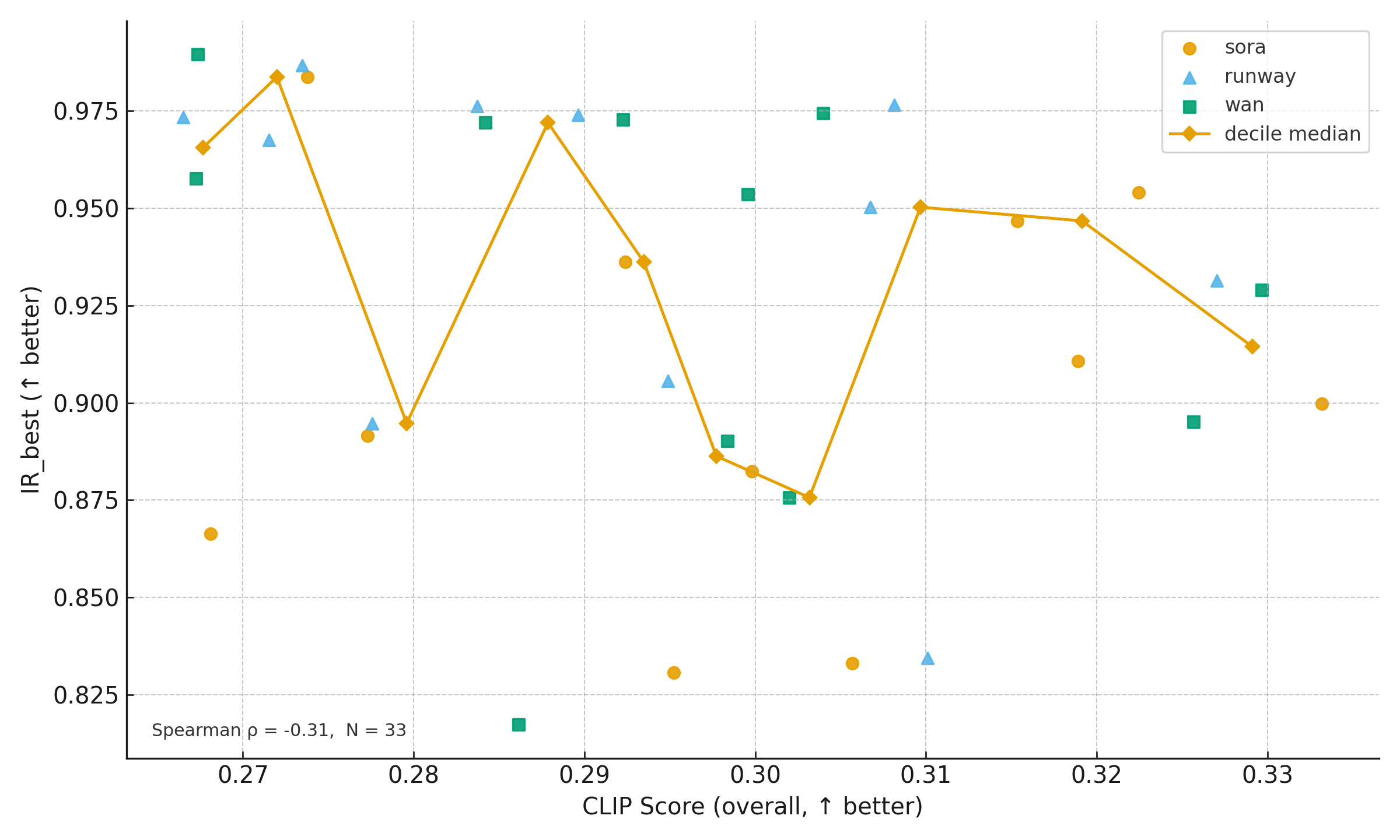}
    \caption{CLIP vs.\ $\mathrm{IR}_{\text{best}}$ (auto-selected; higher is better). Decile medians are overlaid.}
  \end{subfigure}\hfill
  \begin{subfigure}[t]{.49\linewidth}
    \centering
    \includegraphics[width=\linewidth]{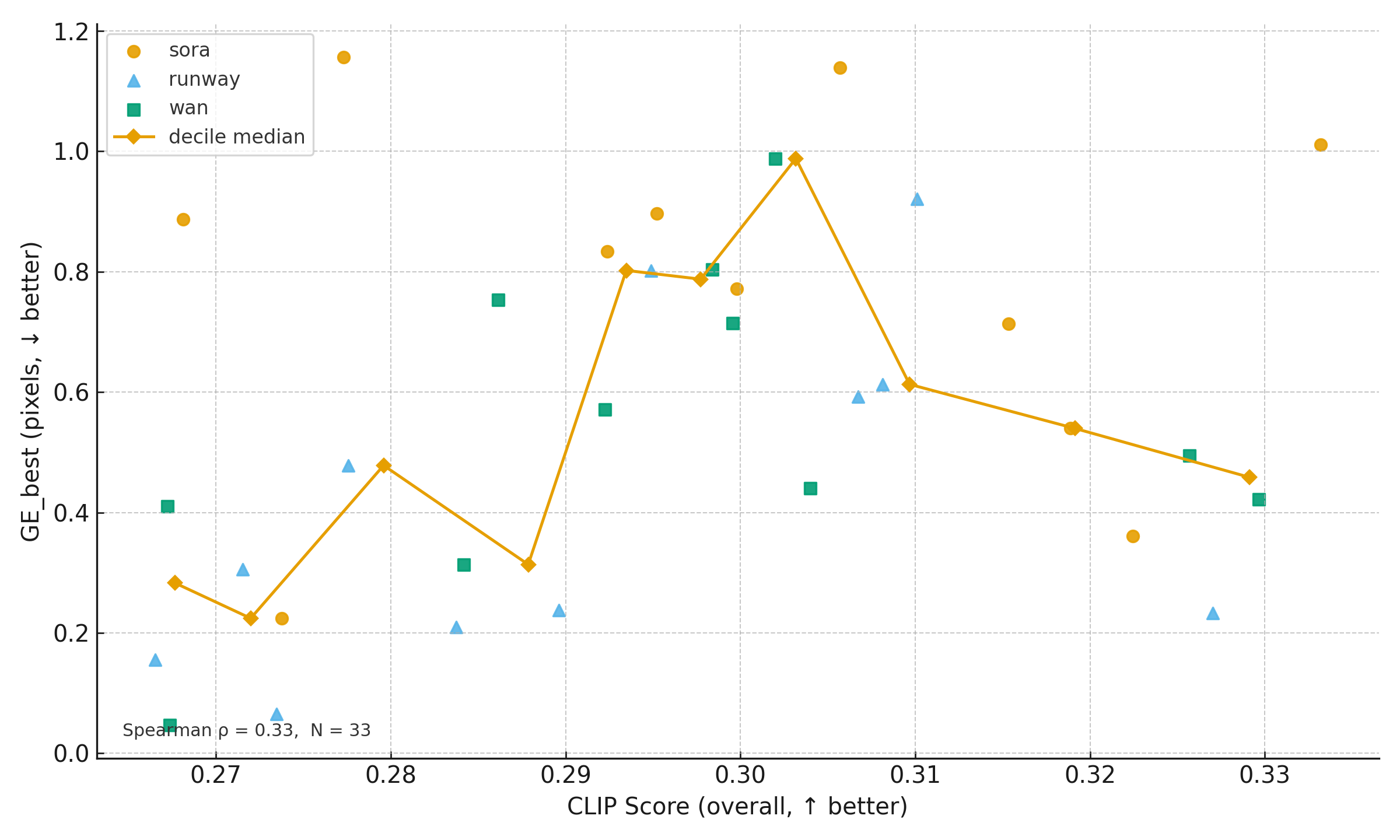}
    \caption{CLIP vs.\ $\mathrm{GE}_{\text{best}}$ in pixels (auto-selected; lower is better). Decile medians are overlaid.}
  \end{subfigure}\\[4pt]
  \begin{subfigure}[t]{.49\linewidth}
    \centering
    \includegraphics[width=\linewidth]{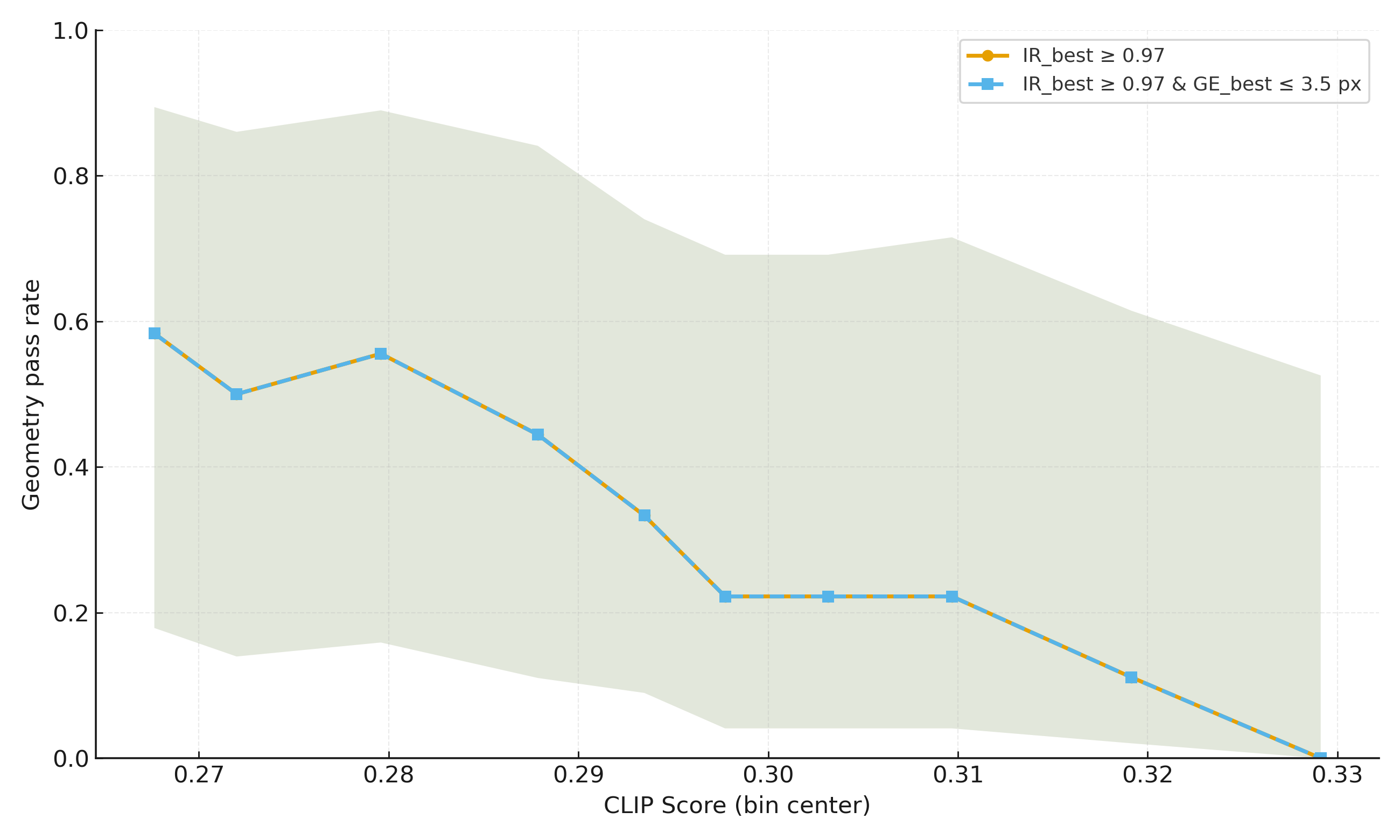}
    \caption{Reliability by CLIP deciles (auto-selected). Criteria: $\mathrm{IR}_{\text{best}}\!\ge 0.97$ (solid) and $\mathrm{IR}_{\text{best}}\!\ge 0.97$ \& $\mathrm{GE}_{\text{best}}\!\le \mathbf{3.5}$\,px (dashed), with 95\% Wilson bands.}
  \end{subfigure}\hfill
  \begin{subfigure}[t]{.49\linewidth}
    \centering
    \includegraphics[width=\linewidth]{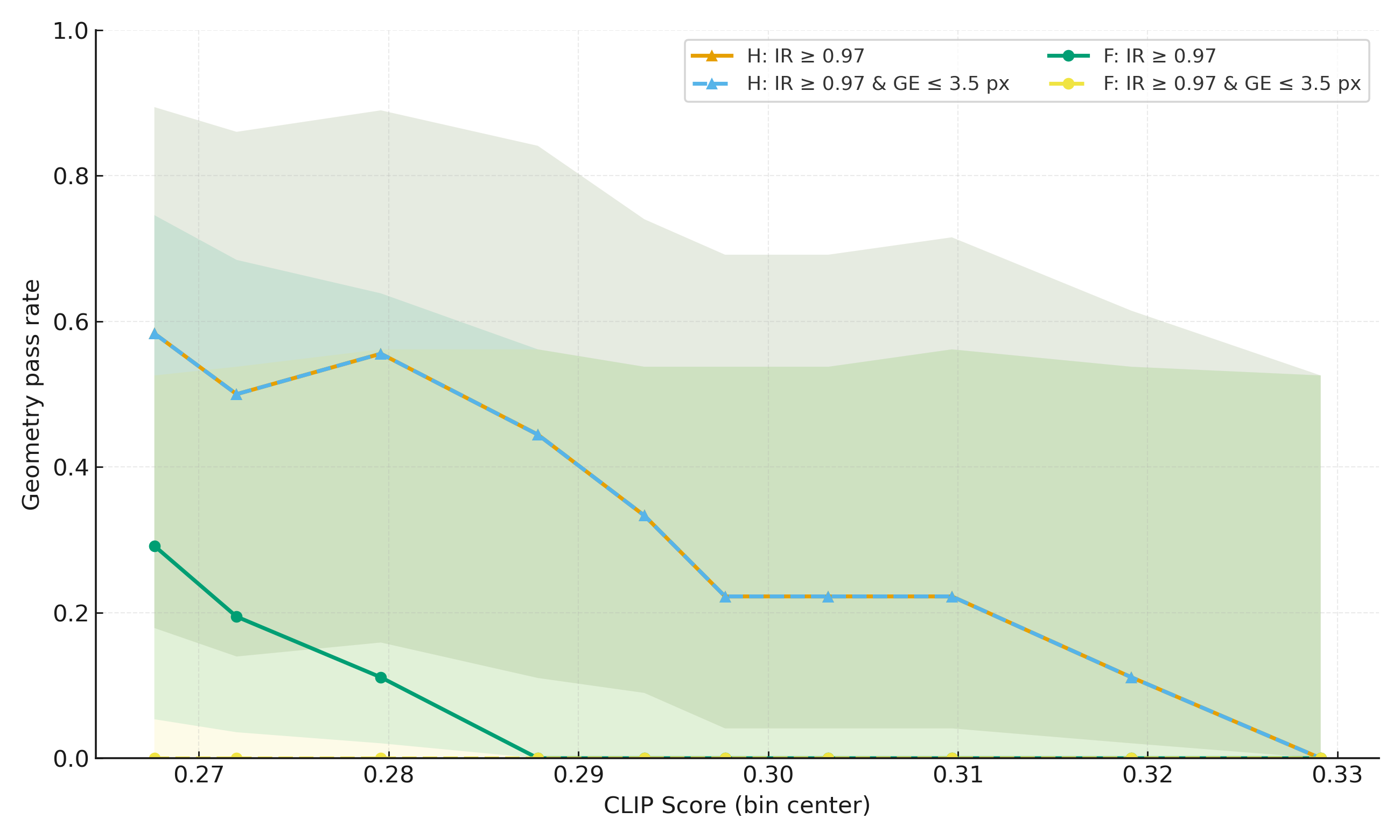}
    \caption{Per-family reliability overlays (H/F). Solid: IR-only; dashed: IR+GE with $\delta=\mathbf{3.5}$\,px. Curves remain flat or weakly non-monotonic with CLIP.}
  \end{subfigure}
  \caption{Semantic vs.\ geometric decoupling.
  All residuals are evaluated in the pixel domain on the content mask (letterboxing excluded). 
  Points aggregate Sora, Runway, and Wan over all seed--category scenes.}
  \label{fig:decouple_main}
\end{figure*}

Fig.~\ref{fig:decouple_main}-(a,b) plot overall CLIP against $\mathrm{IR}_{\text{best}}$ (higher is better) and $\mathrm{GE}_{\text{best}}$ in pixels (lower is better), respectively, pooling Sora, Runway, and Wan. 
Across platforms, the scatter shows no strong monotonic trend; the decile medians overlaid in each panel remain flat or erratic. 
Practically, we observe numerous clips with high CLIP but low $\mathrm{IR}_{\text{best}}$ or elevated $\mathrm{GE}_{\text{best}}$ (px), as well as the converse. 
This directly illustrates that semantic fidelity is not a reliable proxy for geometric consistency.

Fig.~\ref{fig:decouple_main}-(c) reports \emph{reliability by CLIP deciles}: for each CLIP bin we plot the proportion of scenes meeting a pixel-domain geometry criterion, with 95\% Wilson confidence bands. 
We consider two thresholds: (1) $\mathrm{IR}_{\text{best}}\!\ge 0.97$; and (2) $\mathrm{IR}_{\text{best}}\!\ge 0.97$ \emph{and} $\mathrm{GE}_{\text{best}}\!\le 3.5$\,px.
If CLIP were predictive of geometry, the curves should increase monotonically with CLIP. 
Instead, both curves remain flat or mildly non-monotonic, indicating that higher CLIP bins do not consistently translate into better geometry.
Fig.~\ref{fig:decouple_main}-(c) reports \emph{reliability by CLIP deciles}: for each CLIP bin we plot the proportion of scenes meeting a pixel-domain geometry criterion, with 95\% Wilson confidence bands.
Fig.~\ref{fig:decouple_main}-(d) overlays the same decile analysis for the two residual families (solid: IR-only; dashed: IR+GE). 
The shapes of the $\mathbf{H}$ and $\mathbf{F}$ curves are similar and remain weakly dependent on CLIP. 
Thus the decoupling phenomenon is not an artifact of favoring one family: whether a scene is better explained by a homography or a fundamental matrix, CLIP still fails to predict geometric coherence.

The findings, which are consistent across various platforms and resilient to the selection of residual family, indicate a significant decoupling between semantic and geometric quality.
Therefore, evaluating video generators as ''world simulators'' requires reporting geometry-oriented metrics (IR/GE in the pixel domain) in addition to semantic scores. 
In our supplementary material, we provide counterexample lists (high-CLIP with poor geometry and vice versa) and per-family rank correlations, further corroborating this conclusion.

Across three generators and all six categories, CLIP is a poor predictor of geometry: both scatter views (CLIP vs.\ IR$_\text{best}$, CLIP vs.\ GE$_\text{best}$ in pixels) and reliability-by-CLIP deciles remain flat or mildly non-monotonic.
All residuals are evaluated in the pixel domain on the content mask (letterboxing excluded) with a single gate fixed at $\delta=\mathbf{3.5}$\,px and a stringent rigidity criterion $\mathrm{IR}_{\text{best}}\!\ge 0.97$.
Per-family overlays (H/F) behave similarly, ruling out residual-family bias; higher CLIP does not reliably imply better geometry.

\subsection{Perceptual Evaluation}
\begin{figure*}[t]
\centering
\includegraphics[width=0.8\textwidth]{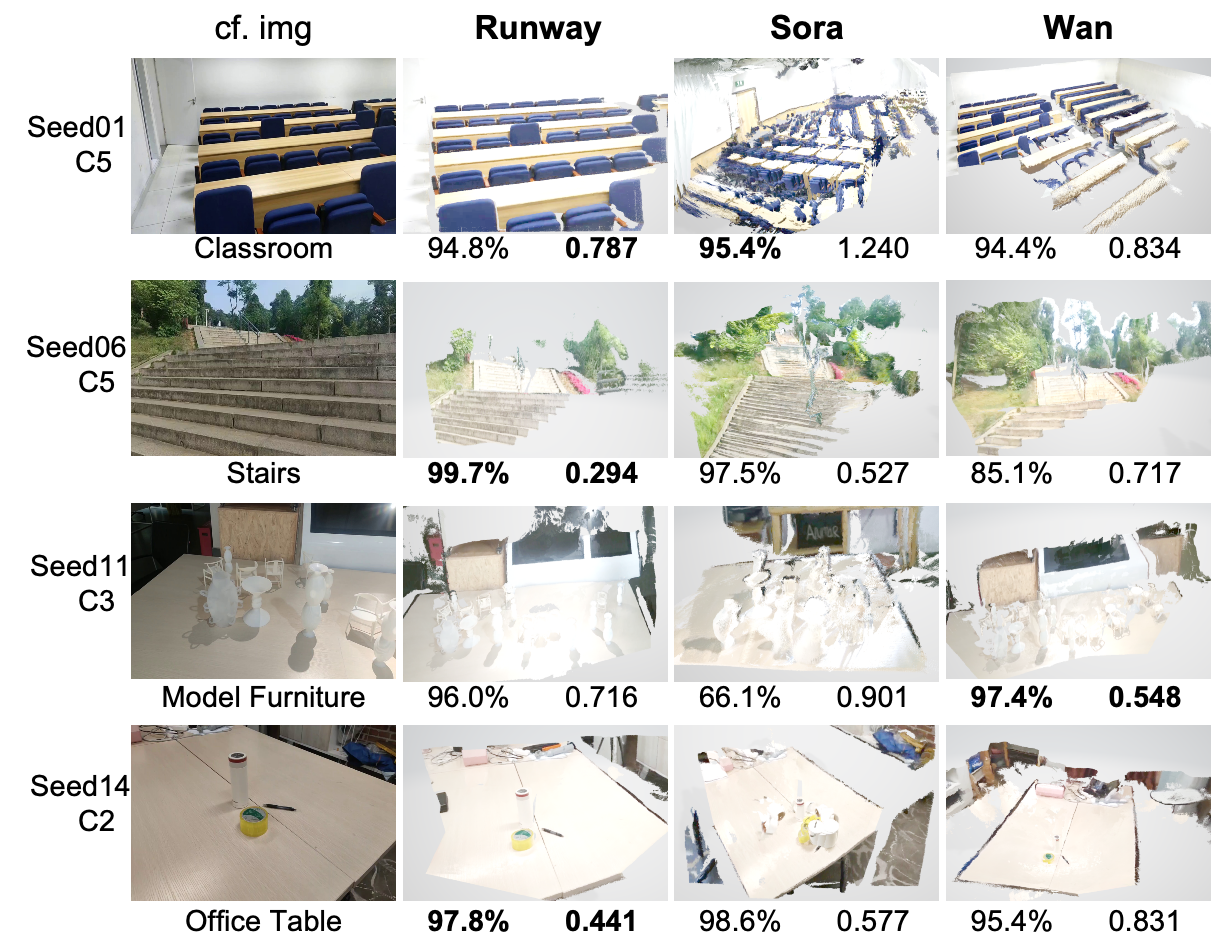}
\caption{Background geometry in practice. For each row, the left cell shows the seed image; the three cells on the right visualize background reconstructions and report IR/GE per platform. Missing or extremely sparse reconstructions are treated as $F$ under our DUSt3R gate and also exhibit poor visual stability.}
\label{fig:qual_ir_ge}
\end{figure*}

Fig.~\ref{fig:qual_ir_ge} assembles row-wise exemplars.
The leftmost cell of each row shows the seed image for context; the three cells to its right visualize the background geometry obtained from adjacent-frame matches on the content mask (letterboxing ignored) for three platforms. For each cell, we report the \emph{Inlier Ratio} (IR) and the \emph{Geometric Error} (GE, median residual in pixels after de-normalization). Two visual regularities are consistent across rows.
Scenes that are visually rigid to a human—e.g., static rooms or static landscapes viewed with small camera motion—systematically present \emph{high} IR (most background correspondences agree with one global model) and \emph{low} GE (aligned inliers are tight). The reconstructed point clouds look compact with clean planes and straight edges; tiny reprojection jitter is perceptually negligible.
When humans perceive background instability—e.g., subtle "breathing" of buildings, bending railings, or tearing near depth discontinuities—the geometry mirrors that: IR drops (more outliers against a single model) and/or GE rises (inlier alignment loosens). In rows where DUSt3R fails to return a pose for the first–last pair (because the displacement is too large or the texture is degenerate), the reconstruction is sparse or absent. By our evaluation policy, those cases are labeled $F$, and they correlate with visible large parallax or scene changes. Overall, the qualitative impressions match the quantitative trend: visually cleaner reconstructions coincide with $IR \geq 0.97$ and $GE \leq 3.5$, whereas visually degraded ones violate one or both.

\begin{figure*}[t]
\centering
\includegraphics[width=0.9\textwidth]{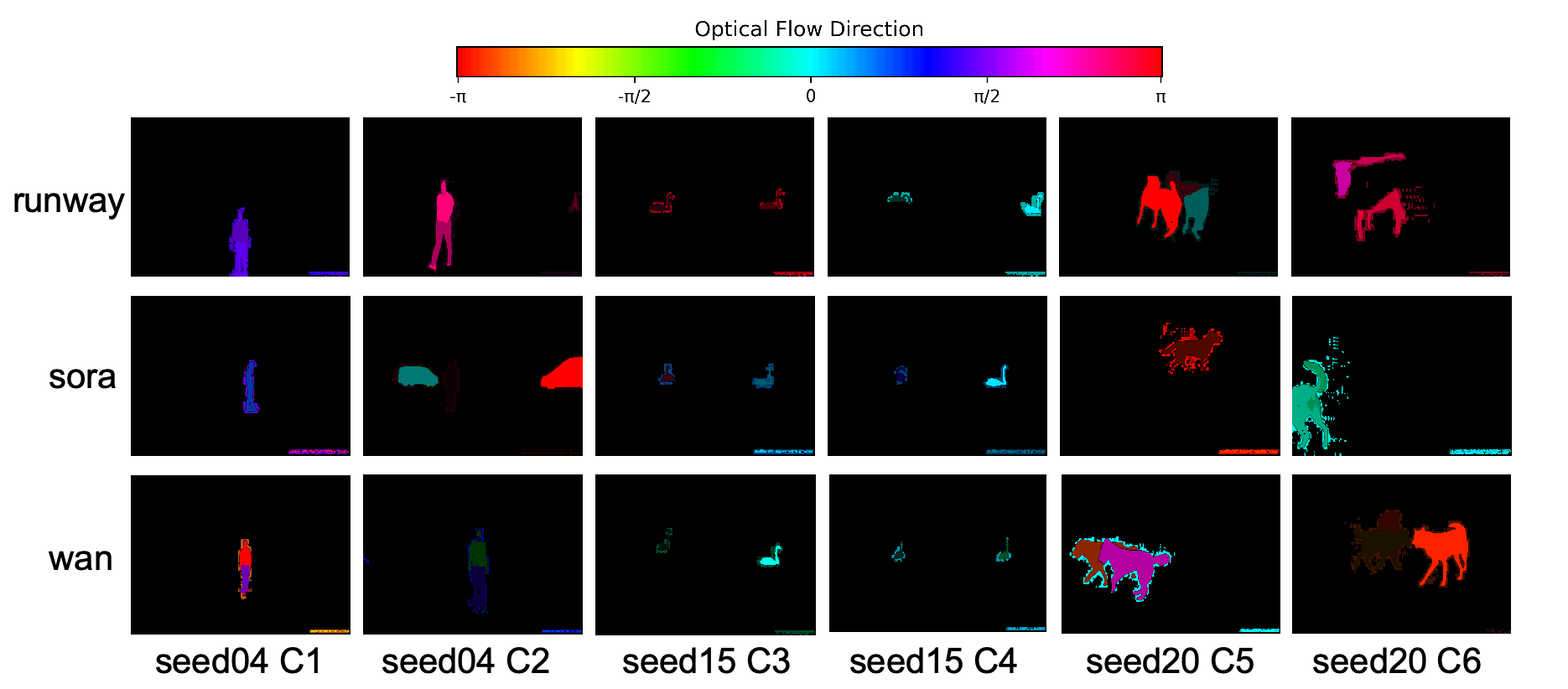}
\caption{Foreground motion slices used by AdaDFRC-W2. Columns pick representative seeds; rows juxtapose platforms. Within each object mask and time step, color encodes the direction of the per-cell \emph{median} optical flow. Directional coherence across the mask indicates near-rigid motion and leads to low AdaDFRC-W2; intra-mask direction changes or flips expose non-rigid/unstable dynamics and increase AdaDFRC-W2.}
\label{fig:qual_optical}
\end{figure*}

Fig.~\ref{fig:qual_optical} presents object-centric optical-flow slices used by our foreground metric. Horizontally, we sample representative seeds; vertically, we juxtapose platforms. For each object mask and frame pair, hue encodes the \emph{direction} of the per-cell \emph{median} flow (magnitude is normalized for visualization only). 
When an object undergoes near-rigid motion (e.g., a pedestrian translating without limb swing emphasis, or a car with minimal rolling artifacts), the flow direction is coherent within the mask and stable over time. Our AdaDFRC-W2—computed by fitting an affine flow on the object support and measuring a Wasserstein-2 transport cost from that rigid prediction to the observed per-cell medians with adaptive reliability weights—stays low.
Non-rigid behavior (articulated limbs, fluttering texture, rolling shutter–like distortions) and occlusion events create sharp direction changes within the mask. The hue then varies within a single frame or flips between frames; these spatio-temporal inconsistencies raise the transport needed from the rigid hypothesis to the observed distribution, yielding a higher AdaDFRC-W2. Cross-platform differences in these hue patterns align with our quantitative ranking of foreground consistency.

The two visual panels ground our metrics in human perception: high IR/low GE tracks "background feels rigid," while low/high signals "background looks wrong"; low AdaDFRC-W2 matches "object moves as a solid," while high values expose non-rigid or unstable motion. These qualitative correspondences validate that our thresholds produce behavior that is not merely numerical but perceptually meaningful.

\section{Discussion}

\subsection{What GeoCon-Bench Measures—and What It Does Not}
GeoCon-Bench operationalizes \emph{geometric consistency} in AIGC video along two complementary, model-agnostic axes. For the \emph{background}, we test whether a single global transformation explains adjacent-frame motion: either a homography ($H$) or a fundamental matrix ($F$), with the model routed by a DUSt3R translation gate on the first/last frames. For \emph{both} models, we report (i) the inlier ratio (IR), defined as the fraction of background correspondences that fit the chosen model under a family-specific inlier gate, and (ii) the geometric error (GE), defined as the median residual among those inliers. Inliers are defined with \emph{family-specific} gates, while \emph{all} GEs are evaluated and reported \emph{in the pixel domain after denormalization on the content mask}. At the clip level, a background is called geometrically consistent when IR${\ge}0.97$ and GE${\le}3.5$\,px. For the \emph{foreground}, AdaDFRC-W2 regresses the best affine motion inside each object mask and summarizes non-rigid residuals, exposing elastic artifacts and layer slippage that a rigid-background test cannot see. This decomposition yields interpretable diagnostics: low IR or high GE indicates camera–scene inconsistency (violations of projective/epipolar geometry), whereas high AdaDFRC-W2 flags dynamic incoherence of movers. Crucially, GeoCon-Bench \emph{does not} measure semantic fidelity; it is designed to complement, not replace, semantic metrics. The dissociation we observe between semantics and geometry underscores why a dedicated geometric axis is necessary.

\subsection{Cross-Platform Trends and Where Models Diverge}
Because prompts are categorized into six families (C1–C6), the benchmark supports per-regime comparisons rather than a single pooled score. Three regularities emerge. \emph{First}, families intended to be H-majority—\textbf{C1} (static), \textbf{C2} (rotation-dominant), and \textbf{C4} (static camera + moving objects)—concentrate mass at high IR and low GE under the homography model; when discrepancies arise, they typically reflect (i) small end-to-end translational drift that pushes a minority of seeds into the F branch, or (ii) segmentation leakage admitting moving edges into the background set, both surfaced as mild but systematic penalties. \emph{Second}, families expected to be F-majority—\textbf{C3} (translation-dominant), \textbf{C5} (complex camera motion), and \textbf{C6} (geometric stress)—exhibit broader dispersion and a higher rate of F decisions; platforms diverge most in \emph{how} parallax is realized (amount of depth-induced motion, stabilization priors), which manifests as different balances of IR versus GE. \emph{Third}, failure handling matters: present-only results isolate pure geometry where DUSt3R succeeds, whereas a conservative fail$\rightarrow$F view additionally counts first/last configurations whose displacement prevents pose recovery—an informative signal under stress conditions.
At the \emph{seed-aligned} level (same seed across platforms), two divergence types are actionable. One is \emph{routing disagreement}: the same seed–family pair routes to H on one platform but to F on another, indicating distinct camera priors or depth realism; such cases dominate off-diagonal cells in cross-platform alignment maps and co-occur with GE shifts. The other is \emph{agreement with different margins}: all platforms agree on H (or F), yet one sits closer to the IR/GE boundary—these “near-fail” seeds are highly sensitive to training or decoding changes and thus serve as high-yield regression tests. Finally, the anomaly distribution separates families where dispersion is intrinsic (\textbf{C6}) from those where instability is a warning sign (\textbf{C2}, \textbf{C4}). Together, these trends motivate reporting \emph{per-family, seed-aligned} summaries in addition to pooled scores: they are diagnostic, reproducible, and immediately actionable for model development.

\subsection{Semantic–Geometric Dissociation and Why It Matters}
High semantic alignment (e.g., CLIP) does not guarantee geometric correctness. We substantiate this with two independent lenses. \emph{(i) Thresholded reliability.} Under a fixed, pixel-domain reporting gate (IR${\ge}0.97$, GE${\le}3.5$\,px), a substantial fraction of clips that score highly on CLIP fail the background rigidity check or exhibit elevated AdaDFRC-W2. This persists whether the background model is auto-selected by DUSt3R or forced per-family (H for C1/2/4, F for C3/5/6), indicating that the dissociation is not a routing artifact (see Fig.~\ref{fig:decouple_main}). \emph{(ii) Distributional stability.} Across CLIP deciles, both the pass rate for background consistency and the distribution of foreground residuals change weakly, if at all, despite large semantic swings—again pointing to orthogonality. Mechanistically, text–vision alignment systems reward object/texture plausibility and global style, whereas our geometry axis constrains camera–scene relations and rigidity. Video generators inherit stabilization or motion-style priors that visually improve coherence yet can suppress or misshape parallax, and they often synthesize movers with elastic shortcuts that evade semantic penalties but are penalized by AdaDFRC-W2. The implication is practical: leaderboards and ablations should report semantics and geometry as \emph{two orthogonal axes}. GeoCon-Bench provides a drop-in geometric axis—background IR/GE in the pixel domain, plus object-level AdaDFRC-W2—so that model builders can diagnose failure modes that semantic metrics systematically miss. Releasing per-clip pose logs, masks, and fixed gates is essential for \emph{reproducibility} and \emph{actionable debugging}.

\subsection{Limitations and Future Work}
Our study is deliberately scoped to geometric behaviors that are most consequential for perception-centric use. We evaluate static or near-rigid backgrounds with a single global model and summarize foreground motion with an affine proxy and distributional residuals, routed by DUSt3R into either planar/rotation or translation/parallax regimes. This design prioritizes interpretability and reproducibility, and within that operating regime, our metrics exhibit stable behavior and align with human judgments. Several methodological boundaries remain. First, the routing currently relies on a two-frame statistic; clip length, baseline, or near-degenerate configurations can bias this decision. Second, residuals are computed in the pixel domain and are mildly sensitive to sensor- or codec-induced artifacts such as rolling shutter and heavy compression. Third, background–foreground separation inherits any errors in segmentation, so boundary perturbations may affect marginal cases even when coverage is reported and failure cases are handled consistently. Finally, our prompt families emphasize common scene regimes rather than exhaustive coverage. These caveats point to concrete extensions: multi-window or temporally smoothed routing that is invariant to clip length; degeneracy-aware estimation with explicit parallax and conditioning checks; residual models that account for sensor-domain effects; uncertainty- and occlusion-aware masks with consistency checks between pose and segmentation; and layered or depth-harmonized background models that retain the current interpretability while broadening applicability. Expanding the dataset with calibrated trajectories and controlled synthetic scenes will further enable stress testing and fine-grained calibration, moving the framework toward a more comprehensive standard for assessing generative video as a geometric process.

\begin{acks}
To Robert, for the bagels and explaining CMYK and color spaces.
\end{acks}

\bibliographystyle{ACM-Reference-Format}
\bibliography{main}

\appendix

\section{Prompt Component Lexicon}
\label{apx:lexicon}

This appendix provides an expanded vocabulary for the prompt components defined in our framework.
\begin{footnotesize}
\begin{longtable}{@{}L{0.22\linewidth} L{0.20\linewidth} L{0.58\linewidth}@{}}
\caption{Expanded Prompt Component Lexicon}\\
\toprule
\textbf{Component} & \textbf{Category} & \textbf{Vocabulary Examples} \\
\midrule
\endfirsthead
\toprule
\textbf{Component} & \textbf{Category} & \textbf{Vocabulary Examples} \\
\midrule
\endhead

\texttt{Motion Command} & Foundational &
``This is a single continuous shot'', ``The camera is locked down'' \\
\cmidrule(l){2-3}
& Camera Rotation &
``pans left'', ``tilts up'', ``rolls clockwise'', ``orbits around'' \\
\cmidrule(l){2-3}
& Camera Translation &
``dollies in'', ``trucks left'', ``cranes up'', ``moves in a straight line'' \\
\cmidrule(l){2-3}
& Figurative \& Phenomenological &
``as if on a cinematic dolly track''; ``creating true motion parallax'';
``as if mounted on a heavy concrete pillar''; ``shadows stretch realistically'';
``reflections move accurately across the surface'' \\
\midrule

\texttt{Qualitative Style} & Realism/Fidelity &
``hyper-realistic'', ``photorealistic'', ``8K resolution'', ``sharp focus'',
``deep depth of field'', ``shot on IMAX film'', ``documentary style'' \\
\cmidrule(l){2-3}
& Lighting &
``cinematic lighting'', ``golden hour'', ``dramatic lighting'',
``studio lighting'', ``neon glow'', ``volumetric lighting'' \\
\cmidrule(l){2-3}
& Artistic Medium &
``in the style of a vintage film'', ``black and white film noir'',
``impressionist painting'', ``stop-motion animation'' \\
\midrule

\texttt{Scene Composition} & Environment &
``a bustling city street'', ``a serene forest path'', ``a minimalist interior'',
``a futuristic laboratory'', ``a Victorian-era room'' \\
\cmidrule(l){2-3}
& Geometric Structure &
``a long corridor with repeating arches'', ``a skyscraper with a grid of windows'',
``a spiral staircase'', ``a checkerboard floor'' \\
\midrule

\texttt{Object Specification} & Material/Texture &
``reflective chrome'', ``matte black finish'', ``rough stone texture'', ``smooth glass'' \\
\cmidrule(l){2-3}
& Object State &
``a pristine new car'', ``a rusted, abandoned vehicle'', ``a blooming flower'' \\
\bottomrule
\end{longtable}
\end{footnotesize}

\section{Additional Evidence for DUSt3R Translation-Gate Calibration}
\label{app:tau_t_more}

We reassess the DUSt3R translation gate using only pairs with available poses (present-only), across all platforms and prompt families.
This analysis is geometric and independent of semantic scores; background inlier gates remain fixed throughout the paper ($\TauH{=}2$\,px, $\TauF{=}0.1$, with $\mathrm{IR}\!\ge\!0.97$ and $\mathrm{GE}\!\le\!3.5$\,px when background consistency is evaluated elsewhere).

\begin{figure}[t]
  \centering
  \includegraphics[width=0.8\linewidth]{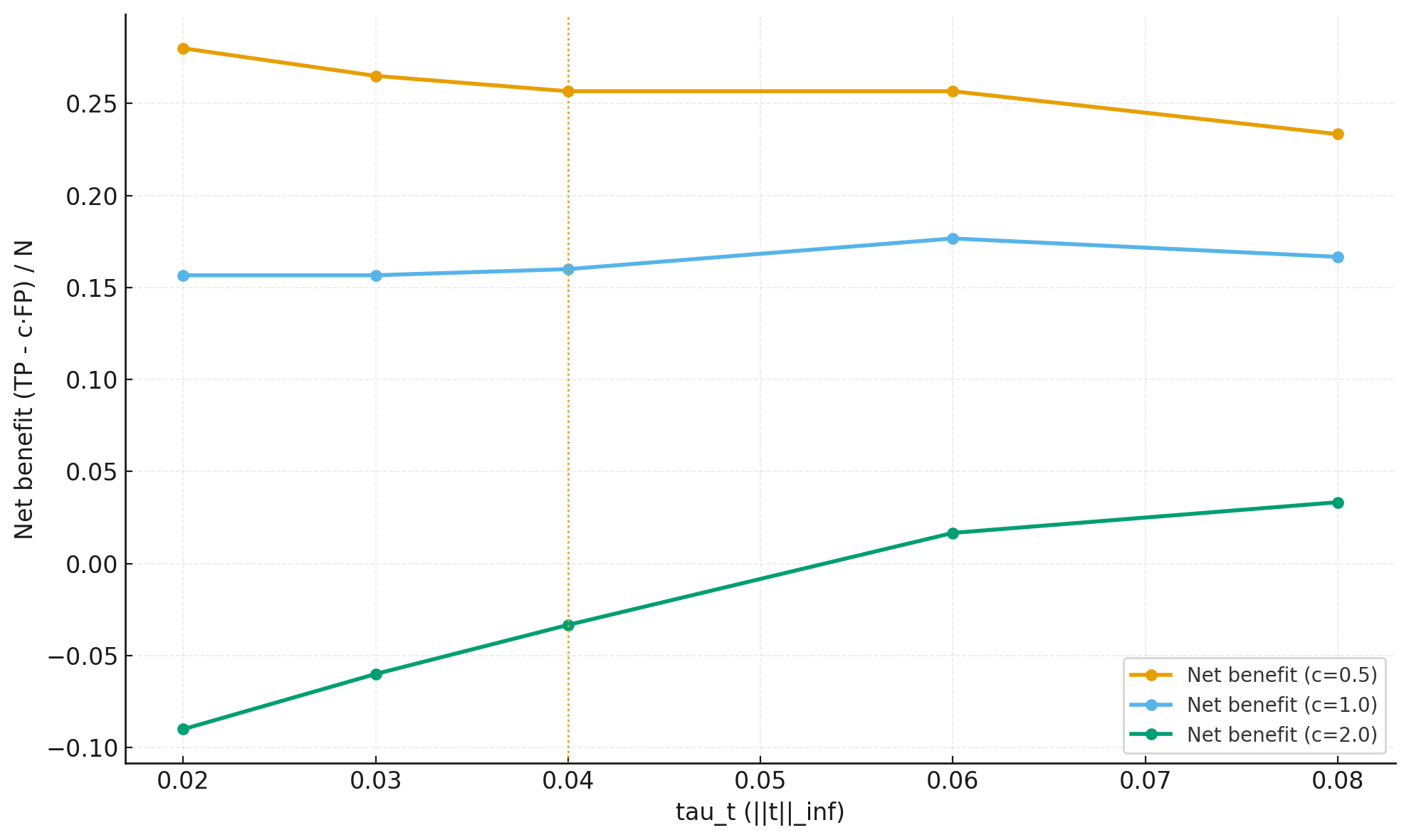}
  \caption{Decision-curve analysis (present-only): net benefit $(\mathrm{TP}-c\cdot\mathrm{FP})/N$ across $c\!\in\!\{0.5,1,2\}$.
  The choice $\tau_t{=}0.04$ lies near the maxima and remains robust to $c$.}
  \label{fig:tau_t_netbenefit}
\end{figure}

We quantify the operating trade-off between recall on F-expected scenes (C3/5/6) and spurious F on H-expected ones (C1/2/4) via the net benefit
\begin{equation}
    \mathrm{NB}(\tau_t)=\frac{\mathrm{TP}-c\cdot\mathrm{FP}}{N},
\end{equation}
for false-positive cost ratios $c\!\in\!\{0.5,1,2\}$. 
Fig.~\ref{fig:tau_t_netbenefit} shows that $\tau_t{=}0.04$ sits close to the maxima simultaneously for all $c$ and is thus robust to reasonable changes in cost.

\begin{figure*}[t]
  \centering
  \begin{subfigure}[t]{.48\linewidth}
    \centering
    \includegraphics[width=\linewidth]{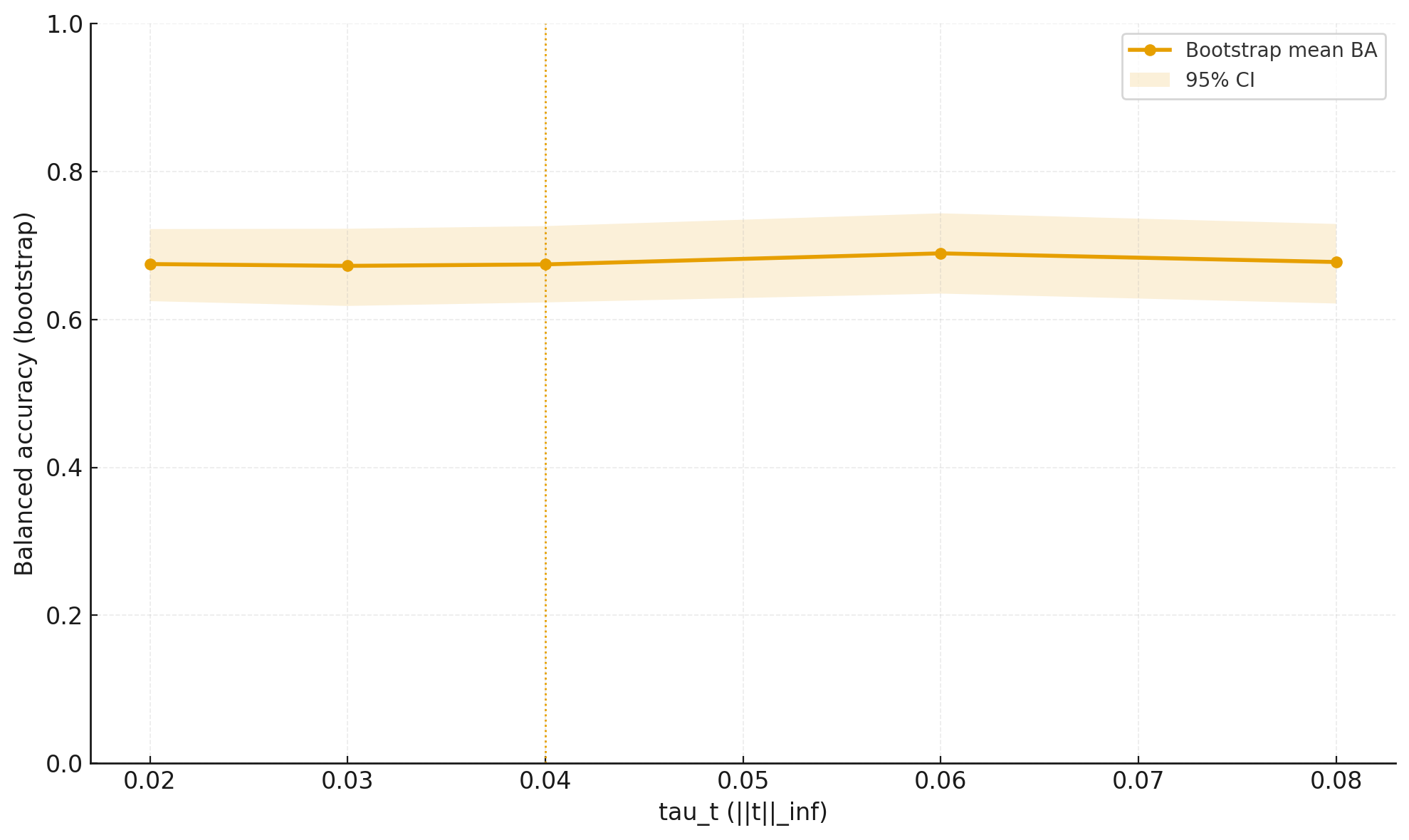}
    \caption{Balanced accuracy vs.\ $\tau_t$ (bootstrap mean $\pm$95\% CI, $B{=}400$). A broad plateau spans $[0.04,0.06]$.}
  \end{subfigure}\hfill
  \begin{subfigure}[t]{.48\linewidth}
    \centering
    \includegraphics[width=\linewidth]{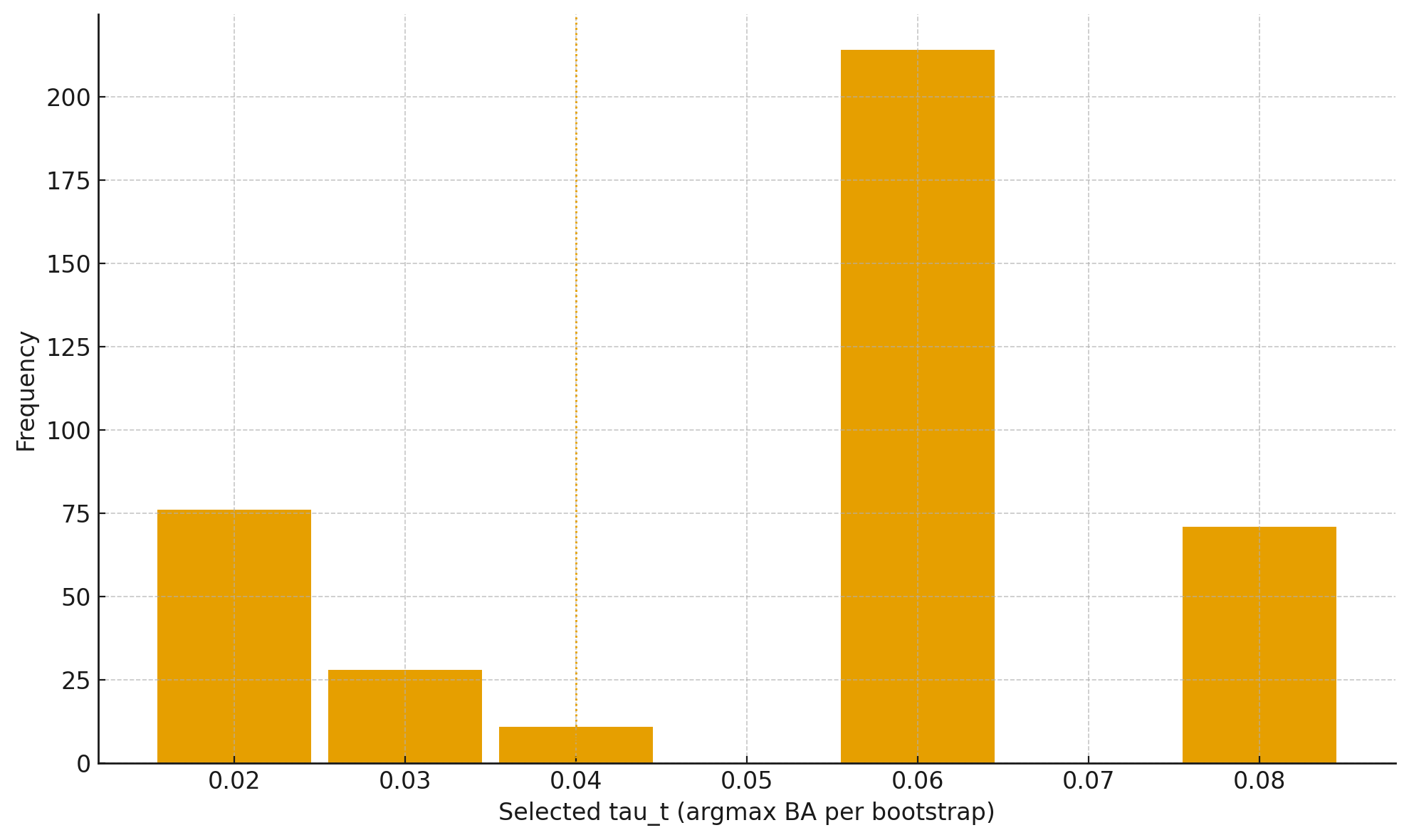}
    \caption{Distribution of $\arg\max$ $\tau_t$ over bootstraps; modes concentrate at $0.04$–$0.06$.}
  \end{subfigure}
  \caption{}
  \label{fig:tau_t_bootstrap}
\end{figure*}

We perform $B{=}400$ bootstrap resamples over the observation set (sampling units are \textit{(seed, platform, family)}), compute balanced accuracy for each $\tau_t\!\in\!\{0.02,0.03,0.04,0.06,0.08\}$, and record both the mean$\pm$95\% CI and the best threshold per resample.
Fig.~\ref{fig:tau_t_bootstrap} (left) reveals a broad performance plateau over $[0.04,0.06]$ where confidence bands strongly overlap; 
the histogram (right) concentrates the best-threshold selections in the same interval, with a visible mode at $0.04$–$0.06$.
Hence $\tau_t{=}0.04$ is either at or statistically indistinguishable from the maxima while being slightly more conservative on false positives.

\end{document}